\def\year{2020}\relax
\documentclass[letterpaper]{article} 
\usepackage{aaai20}  
\usepackage{times}  
\usepackage{helvet} 
\usepackage{courier}  
\usepackage[hyphens]{url}  
\usepackage{graphicx} 
\usepackage{graphicx}  
\usepackage{booktabs}
\usepackage{multirow}

\usepackage{adjustbox}
\usepackage{rotating}
\usepackage{tikz}
\usepackage{pgfplots}
\usepackage{chngcntr}
\usepackage{refcount}
\pgfplotsset{compat=1.5.1}
\usepackage{footnote}
\makesavenoteenv{tabular}
\makesavenoteenv{table}

\newcommand{\lab}[1]{`#1'}
\newcommand{\dataset}[1]{\textsc{#1}}

\title{Can Embeddings Adequately Represent Medical Terminology?\\ New Large-Scale Medical Term Similarity Datasets Have the Answer!\thanks{Please refer to this version for up-to-date experimental results.}}
\author{Claudia Schulz \and Damir Juric\\
Babylon Health\\
London, SW3 3DD, UK\\
\{firstname.lastname\}@babylonhealth.com}

\begin{document}

\maketitle

\begin{abstract}
A large number of embeddings trained on medical data have emerged, but it remains unclear how well they represent medical terminology, in particular whether the close relationship of semantically similar medical terms is encoded in these embeddings.
To date, only small datasets for testing medical term similarity are available, not allowing to draw conclusions about the generalisability of embeddings to the enormous amount of medical terms used by doctors. 
We present multiple automatically created large-scale medical term similarity datasets and confirm their high quality in an annotation study with doctors.
We evaluate state-of-the-art word and contextual embeddings on our new datasets, comparing multiple vector similarity metrics and word vector aggregation techniques.
Our results show that current embeddings are limited in their ability to adequately encode medical terms.
The novel datasets thus form a challenging new benchmark for the development of medical embeddings able to accurately represent the whole medical terminology.
\end{abstract}

\section{Introduction}

AI has recently enabled major breakthroughs in health-care \cite{ardila2019end,liu2018deep}, but it often requires to develop and adapt AI algorithms specifically to the domain \cite{Neumann2019ScispaCyFA}.
Especially \emph{medical terminology} differs largely from commonly used language, so
a crucial step towards the successful use of AI in health-care is to ensure that medical terminology is adequately encoded.
Doctors know a vast amount of medical terms, including which of them are similar (e.g. synonyms of a disease), but it is so far unclear whether 
embeddings
share this deep understanding of medical terminology. 

To investigate this, small datasets of a few hundred medical concept pairs
with a similarity score
have been created
\cite{PedersenPPC2007,PakhomovEtAl2010,ChiuPVK2018}.
However, testing medical language representation models on such restricted datasets does not allow to draw any reliable conclusions about the generalisability of these models to the whole medical terminology.

In this paper, we aim to overcome the generalisation problem by creating \emph{large-scale} medical term similarity datasets, the largest consisting of more than 600,000 term pairs.
Semantically similar medical terms are extracted from
 the SNOMED ontology \cite{donnelly2006snomed}
and we propose a novel strategy for creating pairs of dissimilar terms, resulting in datasets that are highly challenging for embeddings.
To ensure the correctness and reliability of the completely \emph{automatically created} datasets, we perform a manual evaluation with doctors, confirming the datasets' \emph{high quality} and correctness in representing medical term similarity.
We make our code for dataset construction freely available\footnote{\label{repo}\url{https://github.com/babylonhealth/medisim}}, allowing the easy recreation for future research.\footnote{IHTSDO prohibits to publish data derived from SNOMED CT.}

We evaluate publicly available medical word and contextual embeddings on both our new
and existing datasets to compare what conclusions can be drawn from either.
We also compare and analyse the effects of using different similarity metrics, including the commonly used cosine similarity as well as recently suggested rank-based measures \cite{ZhelezniakEtAl2019-correlation}.
We find that existing datasets are too small to realistically reflect the complexity of medical terminology and that they do not reveal significant performance differences between embeddings. In contrast, our new benchmark datasets highlight significant differences between embeddings as well as their inability to adequately represent medical terminology.

As a second evaluation of embeddings' ability to represent medical terminology, we propose a category separation task and a new error metric.
A good medical terminology representation should identify terms in similar categories as being closer than terms in dissimilar categories.

Importantly, our large-scale datasets are not only of interest for testing embeddings on the term similarity task, but also less obvious tasks, such as reducing the time to manually create and verify medical ontologies.

Our contributions are:
1) we introduce highly challenging large-scale medical term similarity benchmarks,
2) we reveal that existing datasets are too small to discover significant performance differences between embeddings, whereas our datasets do, and
3) we find that current embeddings cannot adequately represent medical terminology.

\section{Related Work}

Many benchmark datasets are available to evaluate semantic textual similarity (STS) methods, both on word and sentence level \cite{ZhelezniakEtAl2019-correlation}, but 
most of them are concerned with everyday words and sentences.
However, a method with good performance on these datasets is likely to utterly fail when applied to medical terminology.
For the medical domain, only a handful of similarity datasets exist, as summarised in Table~\ref{tab:existing_data}, all of them manually curated and comprising only commonly used medical concepts.
Furthermore, half contain only single-word terms, although medical terms are frequently made of multiple words.

Note that we here focus on medical \emph{terms} rather than \emph{concepts}. Concepts are abstract entities, represented as codes in ontologies such as SNOMED, which are described by some (potentially more than one) term.

\begin{table}[ht]
    \centering
    \small
    \begin{tabular}{l r c c c}
    \toprule
         \textbf{Dataset} & \textbf{Size} & \textbf{Scores} & \textbf{Source} & \textbf{MW}\\ \midrule
         \textbf{Hliaoutakis} & 36 & 0-1 & MeSH & 47\%\\
          \cite{Hliaoutakis2005} \\
         \textbf{MiniMayoSRS} & 29 & 1-4 & UMLS & 47\%\\
         \cite{PedersenPPC2007} \\
         \textbf{MayoSRS} & 101 & 1-4 & UMLS & 44\%\\
          \cite{PakhomovEtAl2011}\\
         \textbf{UMNSRS-Sim} & 566 & 0-1600 & UMLS & 2\%\\
          \cite{PakhomovEtAl2010}\\
         \textbf{UMNSRS-Sim-mod} & 449 & 0-1600 & UMLS & 0\%\\
          \cite{PakhomovEtAl2016}\\
         \textbf{UMNSRS-Rel} & 587 & 0-1600 & UMLS & 2\% \\
          \cite{PakhomovEtAl2010}\\
         \textbf{UMNSRS-Rel-mod} & 458 & 0-1600 & UMLS & 0\%\\
          \cite{PakhomovEtAl2016}\\
         \textbf{Bio-SimLex} & 988 & 0-10 & PubMed & 0\%\\
          \cite{ChiuPVK2018}\\
         \textbf{Bio-SimVerb} & 1000 & 0-10 & PubMed & 0\%\\
          \cite{ChiuPVK2018}\\
         \bottomrule
    \end{tabular}
    \caption{Existing datasets and \% of multi-word (MW) terms.}
    \label{tab:existing_data}
\end{table}

Regarding the automatic creation of medical terminology datasets, \citeauthor{BeamEtAl2018} \shortcite{BeamEtAl2018} extract pairs of related medical concepts using a bootstrapping approach, resulting in various datasets of related medical UMLS codes extracted from different sources.
In contrast, our dataset focuses on \emph{similar} medical terms.
\citeauthor{AgarwalEtAl2019} \shortcite{AgarwalEtAl2019} use the same approach as \citeauthor{BeamEtAl2018} to create a dataset from SNOMED's `is-a' and other relationships between disorders and drugs.
\citeauthor{WangCZ2015} \shortcite{WangCZ2015} extract 8000 synonym concepts from relationships in UMLS and then randomly create 1.6M negative pairs, whereas 
we also apply a more sophisticated negative sampling strategy.
Neither of these automatically created datasets has been evaluated regarding its quality nor are these datasets publicly available or easy to recreate.

Like us, \citeauthor{HenryCM2018} \shortcite{HenryCM2018} compare different methods for aggregating embeddings of words to measure similarity between multi-word medical concepts. They train their own medical word embeddings 
and compare summing and averaging these vectors to the performance of concept embeddings.
Instead of training yet another embedding model, we use existing embeddings and experiment with a larger variety of word vector aggregation techniques. Furthermore, we use not only cosine similarity to measure vector similarity but also apply rank-based metrics.

\section{New Large-Scale Datasets}
We choose SNOMED Clinical Terms (CT) as the basis for our medical term similarity datasets as it is the ``most comprehensive, multilingual clinical healthcare terminology in the world''\footnote{\url{https://www.snomed.org/snomed-ct/five-step-briefing}}. 
As of the January 2019 release, SNOMED CT comprises 349,548 medical concepts. 
SNOMED CT is thus ideal for our purpose of creating datasets that adequately represent the whole medical terminology used by doctors.
We create \emph{binary classification} datasets, consisting of pairs of medical terms classified as semantically similar (1) or dissimilar (0).
Note that the dataset creation is fully automatic, not requiring any costly manual annotation. 

\subsection{Extracting Positive Instances}
In the first step of the dataset creation, pairs of semantically similar terms are extracted from SNOMED CT.

\noindent\textbf{SNOMED CT Synonyms.}
Each SNOMED CT concept is associated with a unique \emph{fully specified name} (FSN) and may have one or more \emph{synonyms}, e.g.~the FSN \lab{Sprain of ankle} has a synonym \lab{Ankle sprain}.
Clearly, synonyms are semantically very similar to the FSN, so we construct a dataset consisting of all \dataset{FSN-synonym} term pairs as positive instances.

We first filter out 
concepts from the model component module, which provides metadata and organisational concepts such as \lab{Fully specified name} and \lab{Entire term case sensitive}.
For each remaining active concept, we obtain its current FSN and delete parentheses indicating the concept's category, e.g.~\lab{Malaria (disorder)}. We pair the modified FSN with all its active synonyms that are not equivalent to the modified FSN, resulting in the positive instances of our \dataset{FSN-synonym} medical term similarity dataset.
Since each synonym of an FSN is similar to the FSN, we expect that synonyms are also similar to each other. Based on this assumption, we obtain a second dataset \dataset{synonym-synonym}, by adding synonym-synonym term pairs to the \dataset{FSN-synonym} dataset.

\begin{table}[th]
    \centering
    \footnotesize
    \begin{tabular}{l r r r r}
     \toprule
        \textbf{Dataset} & \textbf{Size}  & \textbf{Pos} & \textbf{Neg-R} & \textbf{Neg-L}\\
        \midrule
        \dataset{FSN-syn.} & 451,256 & 16.53 & 37.40 & 7.97\\
        -- easy & 78,466 & 2.08 & 36.90 & 8.37\\
        -- hard & 372,790 & 19.57 & 37.51 & 7.89\\
        \midrule
        \dataset{syn.-syn.} & 726,158 & 16.57 & 35.10 & 8.00 \\
        -- easy & 122,864 & 2.21 & 35.33 & 8.66\\
        -- hard & 603,294 & 19.50 & 35.05 & 7.86\\
        \midrule
        \dataset{poss.-equiv.-to} & 57,528 & 33.92 & 49.33 & 17.95\\
        -- easy & 1,474 & 3.96 & 30.10 & 12.33\\
        -- hard & 56,054 & 34.71 & 49.84 & 18.10\\
        \midrule
        \dataset{replaced-by} & 7,082 & 20.00 & 33.93 & 11.49\\
        -- easy & 654 & 2.94 & 30.59 & 11.29\\
        -- hard & 6,428 & 21.74 & 34.27 & 11.51\\
        \midrule
        \dataset{same-as} & 20,324 & 22.71 & 33.30 & 10.71\\
        -- easy & 2,570 & 2.62 & 27.17 & 10.81\\
        -- hard & 17,754 & 25.62 & 34.19 & 10.70\\
        \bottomrule
    \end{tabular}
    \caption{Our new datasets, respective number of (positive \& negative) term pairs (Size), average Levenshtein distance of the Pos(itive) and  Neg(ative) instances with the R(andom) and L(evenshtein) strategies.}
    \label{tab:datasets}
\end{table}

\noindent \textbf{SNOMED CT Deactivated Concepts.}
Synonyms are the most obvious similar terms, but
 we can leverage another type of information about similar terms in SNOMED CT: in every release, some concepts are deactivated and replaced by a different active concept. An \emph{association} between the concepts gives the reason for replacement: 1) \dataset{possibly-equivalent-to} indicates that the deactivated concept is ambiguous and that the active concept represents one of its possible meanings, 
2) \dataset{replaced-by} applies to erroneous or obsolete deactivated concepts and their suitable replacement, and
3) \dataset{same-as} refers to (semantically) duplicate concepts.
Clearly these associations describe pairs of similar concepts, which we transform into similar term pairs.

Again, we first disregard pairs containing concepts from the model component module.
For each concept we then use the most recent FSN as the term and again drop parentheses specifying medical categories. In addition, we drop any ``[D]'' at the start or end of a FSN, which SNOMED CT uses to indicate deprecated names.
We collect the three types of term pairs in three separate datasets to investigate if any of them are easier or more difficult to identify as similar. 

\noindent \textbf{Easy vs. Hard Datasets.}
The extracted positive instances are expected to all be semantically similar terms. However,  \emph{lexically} the terms can be very similar, e.g.~\lab{Sacrum sprain} and \lab{Sacral sprain}, or completely different, e.g. \lab{Malaria} and \lab{Paludism}.
The latter requires a much deeper understanding of medical terminology, whereas the former can be guessed from the surface similarity.
To investigate how deep the understanding of term representation models is, we split the positive instances of each dataset into \emph{easy} and \emph{hard} ones.
The difficulty is measured in terms of \emph{Levenshtein distance} between the two terms.
We experimentally choose a threshold of 5, so that the hard splits mainly contain term pairs with fundamentally different words.
Table~\ref{tab:datasets} illustrates the average Levenshtein distance of term pairs in the easy and hard datasets.

\subsection{Creating Negative Instances}
SNOMED CT explicitly specifies similar terms (e.g.~synonyms), but not dissimilar ones.
Na\"{i}vely, we can thus consider all term pairs not explicitly specified as similar to be dissimilar.
For each dataset, our \emph{random} negative sampling strategy therefore matches the first term of each positive instance to a randomly selected term from another instance.
As can be seen in Table~\ref{tab:datasets}, this leads to negative term pairs with very high average Levenshtein distance, i.e.~they are mostly made of completely different words with no lexical overlap. This may make it easy for models to correctly identify these term pairs as dissimilar.

To test if models in fact have a deep understanding of medical terminology, we apply a second negative sampling strategy to create more difficult negative instances:
the first term of each positive instance is matched to the term with closest Levenshtein distance that is not (directly or indirectly) specified to be similar.
Table~\ref{tab:datasets} illustrates that this leads to a much lower Levenshtein distance between negative term pairs than using the random strategy. In the hard datasets, the Levenshtein distance of negative instances is even lower than that of positive ones. Thus, for the hard datasets with Levenshtein negative sampling, lexical similarity between terms will not help at all to distinguish similar and dissimilar pairs.

For both negative sampling strategies, we construct the same number of negative instances as there are positive ones to obtain balanced datasets.
The split into easy and hard combined with our two negative sampling strategies results in 20 different datasets.
In contrast to existing datasets, where most medical terms are single words (see Table~\ref{tab:existing_data}), SNOMED CT terms are mostly made of multiple words, resulting in 92\% multi-word terms in our datasets.
This makes the datasets both more realistic, as multi-word terms are more complex and more fine-grained, and more challenging for medical terminology representation models.

\begin{table*}[t]
    \centering
    \small
    \begin{tabular}{l ccccc}
    \toprule
        & \dataset{FSN-syn.} & \dataset{syn.-syn.} & \dataset{poss.-equiv.-to} & \dataset{replaced-by} & \dataset{same-as}\\
        \cmidrule(lr){2-2} \cmidrule(lr){3-3} \cmidrule(lr){4-4} \cmidrule(lr){5-5} \cmidrule(lr){6-6}
         & e-R \hspace{0.2cm}  h-R \hspace{0.2cm} e-L \hspace{0.2cm}  h-L & 
         e-R \hspace{0.2cm}  h-R \hspace{0.2cm} e-L \hspace{0.2cm}  h-L &
         e-R \hspace{0.2cm}  h-R \hspace{0.2cm} e-L \hspace{0.2cm}  h-L & 
         e-R \hspace{0.2cm}  h-R \hspace{0.2cm} e-L \hspace{0.2cm}  h-L&
         e-R \hspace{0.2cm}  h-R \hspace{0.2cm} e-L \hspace{0.2cm}  h-L\\ \midrule
         \textbf{NaN} & 3\% \hfill 13\% \hfill 17\% \hfill 10\% &
          13\% \hfill 10\% \hfill 20\% \hfill  20\% &  
          10\% \hfill 10\% \hfill 13\% \hfill 10\% & 
          17\% \hfill 30\% \hfill 13\% \hfill 37\% & 
          6\% \hfill 6\% \hfill 17\% \hfill  3\% \\
         \textbf{IAA} & 0.88 \hfill 0.87 \hfill 0.85 \hfill 0.93 &
         0.93 \hfill 0.79 \hfill 0.85 \hfill 0.81 &
         0.91 \hfill 0.65 \hfill 0.95 \hfill 0.74 & 
         0.90 \hfill 0.70 \hfill 0.95 \hfill 0.73 & 
         0.82 \hfill 0.83 \hfill 0.88 \hfill 0.86 \\ \midrule
         \textbf{acc} & 0.98 \hfill 0.96 \hfill 0.98 \hfill 0.98 &
         0.96 \hfill 0.96 \hfill 0.93 \hfill 0.94 & 
         1.00 \hfill 0.86 \hfill 0.98 \hfill 0.91& 
         0.96 \hfill 0.86 \hfill 0.95 \hfill 0.86& 
         0.98 \hfill 0.95 \hfill 0.98 \hfill 0.97\\
         \textbf{rec} & 1.00 \hfill 1.00 \hfill 0.96\hfill 1.00& 
         1.00 \hfill 1.00 \hfill 0.96 \hfill 1.00 & 
         1.00 \hfill 1.00 \hfill 1.00 \hfill 0.96 & 
         1.00 \hfill 1.00 \hfill 0.96 \hfill 0.93 & 
         1.00 \hfill 1.00 \hfill 1.00 \hfill 1.00 \\
         \textbf{prec} & 0.97 \hfill 0.92 \hfill 1.00 \hfill 0.97& 
         0.93 \hfill 0.93 \hfill 0.88 \hfill 0.88 & 
         1.00 \hfill 0.70 \hfill 0.96 \hfill 0.86 & 
         0.93 \hfill 0.70 \hfill 0.93 \hfill 0.70 & 
         0.96 \hfill 0.89 \hfill 0.96 \hfill 0.93 \\
         \bottomrule
    \end{tabular}
    \caption{Datasets evaluation: term pairs without ground truth score (NaN), Krippendorff's $\alpha$ IAA,
    acc(uracy), rec(all), and prec(ision) between ground truth and dataset scores for  e(asy)/h(ard) datasets with R(andom)/L(evenshtein) negative sampling.}
    \label{tab:datasets_evaluation}
\end{table*}

\subsection{Quality Evaluation}

To verify the quality of our automatically created datasets, we perform a manual evaluation with three doctors.
For each dataset we randomly select 30 positive and 30 negative instances.
Each doctor thus evaluates (the same) 1200 term pairs.
The doctors are presented with term pairs without knowing which dataset they belong to and have to decide if the terms are similar in the sense that they could be used interchangeably in consultation notes.
They are allowed to look up terms of which they do not remember the meaning and can choose ``don't know'' instead of ``same''/``not same'' for a pair of terms.
To compare the automatically created similarity scores in our datasets to the doctors' assessment, we first combine the three doctors' decisions into a \emph{ground truth score} using majority voting. If there is no majority, we assign no ground truth score (NaN).

The \emph{difficulty} (regarding human judgement) of each dataset is measured in terms of the doctors' inter-annotator agreement (IAA) and the amount of disagreement (NaNs).
The overall IAA is Krippendorff's $\alpha = 0.85$,
with the lowest agreement on a dataset being $\alpha = 0.65$ and the highest $\alpha = 0.95$ (see Table~\ref{tab:datasets_evaluation}). The doctors' decisions can thus be considered reliable.
The mostly higher IAA for easy datasets compared to hard ones confirms the intended difficulty difference.
Importantly, there is no notable difficulty difference between the two strategies for creating negative instances, so even negative instances with lexically very similar terms can be easily identified as negative by the doctors due to their semantic dissimilarity.
\dataset{Replaced-by} datasets are the most difficult as doctors frequently disagree on the similarity of two terms. 

The \emph{quality} of datasets is given by the accuracy of the automatically created dataset scores with regards to the ground truth scores.
Table~\ref{tab:datasets_evaluation} illustrates that \dataset{FSN-synonym} and \dataset{same-as} datasets are of very high quality.
The accuracy of
\dataset{Replaced-by} datasets is lower than for the other datasets, but even the lowest accuracy of 0.86 indicates that they are good-quality datasets.
We observe that all datasets using the random strategy exhibit a \emph{recall} of 1, meaning that negative instances in these datasets are indeed dissimilar terms. 
In contrast, negative instances created using the Levenshtein strategy are sometimes so close that doctors indicate them as in fact being similar terms.
The \emph{precision} furthermore shows that some positive term pairs are in fact dissimilar according to the doctors, which
 occurs more frequently for the hard datasets.
The lower precision in the \dataset{Possibly-equivalent-to} and \dataset{replaced-by} datasets indicates that, as is to be expected, positive instances in these datasets do not always denote \emph{exactly} the same. An example is the pair of terms  \lab{Abortion in first trimester} and \lab{Induced termination of pregnancy}, which are related but according to the doctors not the same.

In summary, our manual evaluation shows that all datasets have reliable similarity scores.

\section{Methods for Measuring Term Similarity}
Many embeddings specifically trained for the use in medical applications have been suggested in recent years and tested on different subsets of the existing concept similarity datasets. Instead of presenting a new embedding model to test on our dataset, we evaluate publicly available existing embeddings. Note that unfortunately many of the embeddings performing best on existing datasets are not available \cite{LingEtAl2017,HenryCM2018}.
We also do not test concept embeddings as our datasets are based on \emph{terms}, so a pair of terms may belong to the same concept.

\subsection{Word and Contextual Embeddings}

We evaluate the following types of embeddings (see Tables~\ref{tab:embeddings} and \ref{tab:embedding_coverage} in the Appendix for more detail).\\
 \textbf{1)  word2vec skip-gram \cite{MikolovEtAl2013}:} \\
 \indent - 4 \emph{Bio} embeddings \cite{PyysaloEtAl2013} trained on PubMed Central (PMC), PubMed (PM), both (PP), and both plus Wikipedia (PPW);\\
 \indent - 1 embedding trained on the \emph{BioASQ} challenge dataset \cite{KosmopoulosAP2015};\\
 \indent - 2 embeddings with window sizes 2 and 30 by the Language Technology Lab (\emph{LTL}) \cite{ChiuEtAl2016};\\
 \indent - 2 embeddings by the 
 Athens University of Economics and Business (\emph{AUEB}) with vector dimensionalities 200 and 400 \cite{McdonaldEtAl2018}.\\
 \textbf{2) Fasttext \cite{BojanowskiGJM2017}:}\\
 \indent - 2 embeddings using the \emph{MeSH} thesaurus in addition to PM for training with window size 2 for \emph{intrinsic} tasks and size 20 for \emph{extrinsic} ones \cite{ZhangEtAl2019};\\
 \indent - 1 embedding (and its model (M)) based on the previous plus the \emph{MIMIC}-III dataset \cite{ChenPL2019}. \\
 \textbf{3) Non-medical:} As a comparison, we also include \\
 \indent - the \emph{GloVe} word embedding \cite{PenningtonEtAl2014}; \\
 \indent - 2 \emph{Fasttext} embeddings trained on Wikipedia and Common Crawl (plus its model (M)) \cite{MikolovEtAl2018}.
 
The MeSH and MIMIC embeddings have least out-of-vocabulary terms (OOV) regarding our new datasets, but some of the other embeddings (esp. non-medical) can represent less than 50\% of terms (see Table~\ref{tab:embeddings} in the Appendix).

\noindent \textbf{4) Contextual embeddings:} since the majority of terms in our new datasets are made of multiple words, we also experiment with 
\emph{ELMo} \cite{peters-etal-2018-deep} and its biomedical version \emph{ELMoPubMed}, \emph{Flair} \cite{akbik2019naacl} trained on PubMed, \emph{BERT}  \cite{DBLP:journals/corr/abs-1810-04805} and its biomedical version \emph{SciBERT} \cite{Beltagy2019SciBERT}, and \emph{GPT} \cite{Radford2018ImprovingLU}.

\begin{table*}[th]
    \centering
    \small
    \begin{tabular}{l l l l l l l l l l}
         Dataset & \textbf{Hlia.} & \textbf{MM-av} & \textbf{Mayo} & \textbf{Sim} & \textbf{Sim-m} & \textbf{Rel} & \textbf{Rel-m} & \textbf{SimLex} & \textbf{SimVerb}\\ 
         Subset Size & 36/36 & 29/29 & 81/101 & 352/566 & 340/449 & 347/587 & 339/458 & 964/988 & 909/1000\\
         \midrule
Bio PMC & $0.53^{2}$ & $0.80^{3}$ & $0.44^{3}$ & $0.48^{5/-12}$ & $0.46^{5/-14}$ & $0.36^{4/-16}$ & $0.36^{4/-16}$ & $0.71^{5/-3}$ & $0.45^{4/-4}$ \\
Bio PM & $0.59^{6}$ & $0.83^{3}$ & $0.54^{3}$ & $0.58^{7/-2}$ & $0.56^{7/-3}$ & $0.47^{7/-4}$ & $0.48^{7/-5}$ & $0.69^{4/-5}$ & $0.44^{4/-5}$ \\
Bio PP & $\mathbf{0.59^{7}}$ & $0.78^{4}$ & $0.50^{3}$ & $0.54^{7/-9}$ & $0.53^{7/-9}$ & $0.44^{6/-7}$ & $0.45^{7/-7}$ & $0.71^{5/-2}$ & $0.45^{4/-4}$ \\
Bio PPW & $0.57^{2}$ & $\mathbf{0.85^{8}}$ & $0.50^{3}$ & $0.54^{7/-8}$ & $0.53^{6/-8}$ & $0.45^{7/-7}$ & $0.45^{6/-7}$ & $0.72^{8/-1}$ & $0.47^{5/-4}$ \\
BioASQ & $0.48^{1}$ & $0.80^{3}$ & $0.55^{3}$ & $0.60^{9/-1}$ & $0.59^{9/-2}$ & $0.48^{7/-4}$ & $0.49^{7/-4}$ & $0.69^{4/-5}$ & $0.42^{3/-12}$ \\
LTL win2 & $0.52^{2}$ & $0.76^{2}$ & $0.47^{3/-1}$ & $0.60^{8/-2}$ & $0.59^{8/-2}$ & $0.50^{7/-2}$ & $0.51^{7/-2}$ & $0.72^{5/-2}$ & $0.46^{5/-4}$ \\
LTL win30 & $\mathbf{0.59^{7}}$ & $0.81^{4}$ & $\mathbf{0.57^{5}}$ & $\mathbf{0.66^{14}}$ & $\mathbf{0.66^{13}}$ & $0.58^{13}$ & $0.59^{13}$ & $0.69^{4/-4}$ & $0.44^{4/-6}$ \\
AUEB200 & $0.42^{-1}$ & $0.78^{3}$ & $0.51^{3}$ & $0.62^{9}$ & $0.62^{9}$ & $0.53^{9/-2}$ & $0.54^{9/-2}$ & $0.71^{6/-2}$ & $0.46^{5/-4}$ \\
AUEB400 & $0.48$ & $0.77^{2}$ & $0.51^{3}$ & $0.64^{9}$ & $0.63^{9}$ & $0.54^{9/-1}$ & $0.55^{10}$ & $0.72^{6/-2}$ & $0.47^{5/-4}$ \\
MeSH extr & $0.46^{1}$ & $0.82^{7}$ & $0.50^{3}$ & $0.63^{9/-1}$ & $0.62^{9/-1}$ & $0.54^{9/-1}$ & $0.55^{9/-1}$ & $0.70^{5/-4}$ & $0.47^{5/-4}$ \\
MeSH intr & $0.42$ & $0.82^{4}$ & $0.55^{3}$ & $\mathbf{0.66^{14}}$ & $\mathbf{0.65^{14}}$ & $\mathbf{0.59^{15}}$ & $\mathbf{0.59^{14}}$ & $0.66^{4/-14}$ & $0.44^{4/-4}$ \\
MIMIC & $0.51^{1}$ & $0.81^{4}$ & $0.53^{3}$ & $0.64^{9}$ & $0.63^{9}$ & $0.56^{11}$ & $0.57^{11}$ & $0.71^{5/-2}$ & $0.48^{6/-3}$ \\
MIMIC M & $0.52^{1}$ & $0.81^{4}$ & $0.53^{3}$ & $0.64^{9}$ & $0.63^{10}$ & $0.56^{11}$ & $0.57^{11}$ & $0.71^{6/-2}$ & $0.48^{6/-2}$ \\
\midrule
GloVe & $0.37$ & $0.53^{-2}$ & $0.37^{1}$ & $0.55^{5/-2}$ & $0.54^{6/-2}$ & $0.49^{7}$ & $0.49^{7}$ & $0.75^{10}$ & $0.56^{16}$ \\
Fastt Wiki & $0.29^{-3}$ & $0.57^{-1}$ & $0.38$ & $0.53^{5/-2}$ & $0.55^{6}$ & $0.49^{7}$ & $0.52^{7}$ & $0.75^{11}$ & $0.56^{15}$ \\
Fastt Cr & $0.32^{-3}$ & $0.57^{-2}$ & $0.40^{1}$ & $0.59^{6}$ & $0.59^{7}$ & $0.54^{7}$ & $0.55^{7}$ & $0.77^{18}$ & $\mathbf{0.58^{18}}$ \\
Fastt Cr M & $0.30^{-3}$ & $0.57^{-2}$ & $0.41^{1}$ & $0.58^{6}$ & $0.59^{7}$ & $0.54^{7}$ & $0.55^{7}$ & $\mathbf{0.77^{19}}$ & $\mathbf{0.58^{18}}$ \\
\midrule
ELMoPM & $0.42^{2}$ & $0.66^{1}$ & $0.38^{1}$ & $0.44^{5/-14}$ & $0.42^{5/-15}$ & $0.33^{4/-15}$ & $0.34^{4/-15}$ & $0.72^{6/-2}$ & $0.51^{6}$ \\
Flair & $-0.07^{-11}$ & $0.06^{-10}$ & $0.18^{-1}$ & $0.19^{-18}$ & $0.19^{-18}$ & $0.08^{-18}$ & $0.08^{-18}$ & $0.38^{-19}$ & $0.18^{-19}$ \\
SciBERT & $0.40$ & $0.59$ & $0.26$ & $0.19^{-18}$ & $0.19^{-18}$ & $0.21^{-16}$ & $0.21^{-16}$ & $0.35^{-19}$ & $0.30^{-18}$ \\
BERT & $-0.01^{-3}$ & $0.21^{-9}$ & $-0.01^{-13}$ & $0.12^{-18}$ & $0.11^{-18}$ & $0.08^{-18}$ & $0.04^{-18}$ & $0.39^{-19}$ & $0.18^{-19}$ \\
ELMo & $0.00^{-7}$ & $0.11^{-13}$ & $0.08^{-17}$ & $0.20^{-18}$ & $0.21^{-18}$ & $0.13^{-18}$ & $0.14^{-18}$ & $0.63^{4/-9}$ & $0.5^{4/-2}$ \\
GPT & $0.00^{-1}$ & $-0.17^{-13}$ & $0.01^{-13}$ & $0.10^{-18}$ & $0.09^{-18}$ & $0.08^{-18}$ & $0.06^{-18}$ & $0.32^{-19}$ & $0.26^{-19}$ \\
    \end{tabular}
    \caption{Spearman's correlation of each embedding ($fJ$ for word embeddings, $avg\_cos$ for GloVe, $\tau$ for contextual embeddings).
    An embedding has significantly better/worse correlation than the number of embeddings given by the positive/negative superscripts ($\alpha = 0.0002$, i.e. $\alpha = 0.05$ with Bonferroni correction). MM-av: average scores of coders and physicians.}
    \label{tab:significance_existingDatasets} 
\end{table*}

\subsection{Similarity Metrics}
In contrast to contextual embeddings, which compute a single term vector for any multi-word input string (e.g. a term), word embeddings can only represent single words so that the different word vectors of a multi-word term need to be aggregated to form a \emph{term vector}. To compare the similarity of embedding vectors, the most commonly applied metric is $cos$(ine) similarity.
\citeauthor{HenryCM2018} \shortcite{HenryCM2018}
experimented with averaging and summing word vectors to obtain a term vector and using $cos$ as a similarity measure, but found no significant difference.

We experiment with applying similarity measures to averaged ($avg$) word vectors as well as computing pairwise ($pair$) word similarities and averaging these. 
In addition to $cos$ as a similarity measure, we apply the rank correlation coefficients (Pearson's) $r$, (Spearman's) $\rho$ and (Kendall's) $\tau$, as recently proposed by \citeauthor{ZhelezniakEtAl2019-correlation} \shortcite{ZhelezniakEtAl2019-correlation}.
For word embeddings, we furthermore experiment with fuzzy Jaccard ($fJ$) 
and max Jaccard ($mJ$) similarity, which can handle multi-word strings \cite{ZhelezniakEtAl2019-fuzzyJaccard}.

\section{Evaluation}
To compare what conclusions can be drawn from existing versus our new datasets, we evaluate embeddings on both, investigating 1) which similarity metric works best for the various embeddings and whether the differences are significant and 2) which embedding performs best on each dataset and whether the differences are significant.
For fair comparison, all analyses are performed on a subset of each dataset containing no OOV instances for any embedding.
For the interested reader, detailed results are in the Appendix.

\subsection{Small Existing Datasets}
As in previous work, we measure the performance of embeddings in terms of Spearman's correlation. Since the similarity scores of different embeddings are not independent and we cannot assume that they are normally distributed, bias-corrected and accelerated (BCa) bootstrap confidence intervals \cite{ZhelezniakEtAl2019-fuzzyJaccard}
are applied to assess if there are significant differences between the predictions of different embeddings with different similarity metrics.

\noindent\textbf{Effect of Similarity Metrics.}
Comparing the Spearman's correlations of a word embedding obtained with the different similarity metrics, no metric consistently performs best (see Table~\ref{tab:significance_existingDatasets_full} in the Appendix).
We find nearly \emph{no significant differences} between applying different similarity metrics to an embedding on the Hliaoutakis and \mbox{MiniMayoSRS} datasets (see Tables~\ref{tab:significance_existingDatasets_methods1}-\ref{tab:significance_existingDatasets_methods4} in the Appendix). This illustrates that these datasets are simply \emph{too small} to draw any meaningful conclusions about performance differences of different embeddings and similarity metrics.
For the larger datasets, $mJ$ has significantly lower correlation than most other similarity metrics for various word embeddings. This is interesting as \citeauthor{ZhelezniakEtAl2019-fuzzyJaccard} \shortcite{ZhelezniakEtAl2019-fuzzyJaccard} find that for sentence similarity tasks $mJ$ outperforms $avg\_cos$.
None of the other similarity metrics performs significantly better than all others for any dataset and word embedding. 
We therefore use the standard $avg\_cos$ 
to compare the performance of word embeddings in the next section, except for GloVe where $avg\_r$ is applied as it significantly outperforms most other metrics (on the larger datasets).
For contextual embeddings, $\rho$ and $\tau$ are often
significantly better than the other metrics, with the latter slightly outperforming the former. We therefore use $\tau$ for the
comparison of embeddings. 

\begin{table*}[ht]
    \centering
\scriptsize	
     \begin{tabular}{l l l l l l l l l l l}
\multirow{2}{*}{Dataset}& \textbf{\dataset{FSN-syn.}} & \textbf{\dataset{FSN-syn.}} & \textbf{\dataset{syn-syn.}} & \textbf{\dataset{syn-syn.}} & \textbf{\dataset{poss.-equ.}} & \textbf{\dataset{poss.-equ.}} & \textbf{\dataset{repl.-by}} & \textbf{\dataset{repl.-by}} & \textbf{\dataset{same-as}} & \textbf{\dataset{same-as}} \\
& \textbf{easy} & \textbf{hard} & \textbf{easy} & \textbf{hard} & \textbf{easy} & \textbf{hard} & \textbf{easy} & \textbf{hard} & \textbf{easy} & \textbf{hard} \\ 
Subset Size & 65.2\% & 58.6\% & 63.0\% & 56.5\% & 72.3\% & 69.0\% & 49.4\% & 62.9\% & 69.7\% & 71.0\% \\
\midrule
BioNLP PMC & $74.5^{9/-13}$ & $54.9^{13/-6}$ & $73.8^{9/-13}$ & $52.4^{6/-10}$ & $77.1^{4/-4}$ & $52.7^{10/-8}$ & $70.6^{3}$ & $56.7^{9/-5}$ & $79.0^{5/-3}$ & $57.7^{11/-5}$ \\
BioNLP PM & $77.0^{16/-4}$ & $55.5^{17/-3}$ & $76.1^{16/-4}$ & $53.2^{15/-6}$ & $79.1^{4}$ & $53.3^{16/-6}$ & $74.3^{4}$ & $59.7^{18}$ & $79.2^{5/-3}$ & $58.9^{16/-4}$ \\
BioNLP PP & $76.4^{11/-5}$ & $54.5^{12/-10}$ & $75.6^{11/-7}$ & $52.4^{7/-10}$ & $78.3^{4/-3}$ & $53.0^{13/-7}$ & $73.1^{3}$ & $57.7^{10/-4}$ & $79.4^{6/-3}$ & $58.0^{11/-5}$ \\
BioNLP PPW & $76.3^{11/-5}$ & $53.9^{10/-11}$ & $75.4^{11/-7}$ & $52.4^{6/-10}$ & $78.3^{4/-1}$ & $52.8^{12/-7}$ & $72.1^{3}$ & $57.6^{9/-4}$ & $79.1^{5/-3}$ & $57.7^{11/-5}$ \\
BioASQ & $76.6^{11/-5}$ & $55.2^{14/-4}$ & $75.4^{11/-7}$ & $52.6^{14/-8}$ & $79.6^{4}$ & $59.4^{20/-1}$ & $74.3^{4}$ & $59.9^{18}$ & $78.9^{5/-4}$ & $60.9^{19/-1}$ \\
LTL win2 & $76.3^{11/-5}$ & $52.3^{7/-15}$ & $75.7^{12/-7}$ & $52.4^{6/-11}$ & $78.7^{4}$ & $52.6^{10/-7}$ & $73.1^{3}$ & $56.6^{8/-6}$ & $79.5^{7/-2}$ & $55.2^{7/-12}$ \\
LTL win30 & $75.1^{10/-12}$ & $57.4^{21/-1}$ & $74.5^{10/-12}$ & $54.8^{21/-1}$ & $\mathbf{81.1^{9}}$ & $54.3^{17/-5}$ & $72.8^{3}$ & $60.6^{18}$ & $81.1^{8}$ & $60.9^{19/-1}$ \\
AUEB200 & $78.8^{21/-1}$ & $55.1^{13/-5}$ & $77.4^{19/-1}$ & $52.5^{13/-9}$ & $80.8^{7}$ & $57.8^{18/-3}$ & $73.4^{3}$ & $58.1^{11/-4}$ & $81.6^{9}$ & $57.9^{11/-4}$ \\
AUEB400 & $78.4^{19/-2}$ & $55.6^{18/-3}$ & $77.3^{19/-1}$ & $53.3^{15/-5}$ & $80.1^{5}$ & $57.6^{18/-3}$ & $73.1^{3}$ & $58.0^{10/-5}$ & $82.0^{13}$ & $58.0^{11/-4}$ \\
MeSH extr & $\mathbf{79.2^{22}}$ & $56.7^{20/-2}$ & $\mathbf{77.7^{22}}$ & $53.9^{17/-2}$ & $81.1^{8}$ & $59.5^{20/-1}$ & $74.6^{4}$ & $59.3^{13/-1}$ & $82.3^{15}$ & $60.4^{19/-1}$ \\
MeSH intr & $78.3^{19/-2}$ & $\mathbf{58.5^{22}}$ & $77.1^{19/-1}$ & $\mathbf{55.6^{22}}$ & $\mathbf{81.6^{9}}$ & $\mathbf{61.4^{22}}$ & $74.0^{4}$ & $\mathbf{61.2^{19}}$ & $\mathbf{82.9^{18}}$ & $\mathbf{64.2^{22}}$ \\
MIMIC & $76.7^{12/-4}$ & $55.0^{13/-5}$ & $76.0^{16/-4}$ & $52.4^{6/-10}$ & $79.0^{4}$ & $52.5^{10/-9}$ & $74.3^{3}$ & $57.4^{9/-4}$ & $80.2^{7/-1}$ & $57.4^{11/-5}$ \\
MIMIC M & $76.7^{12/-4}$ & $55.0^{13/-5}$ & $76.0^{16/-4}$ & $52.4^{6/-10}$ & $79.0^{4}$ & $52.5^{10/-9}$ & $74.3^{3}$ & $57.4^{9/-4}$ & $80.0^{7/-1}$ & $57.4^{11/-5}$ \\
\midrule
GloVe & $72.2^{8/-14}$ & $51.6^{6/-16}$ & $71.7^{8/-14}$ & $52.4^{6/-10}$ & $78.1^{4}$ & $52.0^{9/-13}$ & $70.3^{3}$ & $54.9^{6/-12}$ & $77.6^{5/-5}$ & $55.1^{7/-12}$ \\
Fastt Wiki & $61.0^{-22}$ & $53.6^{10/-11}$ & $61.6^{-22}$ & $53.6^{16/-4}$ & $60.0^{-20}$ & $50.9^{-16}$ & $54.2^{-20}$ & $51.0^{-17}$ & $58.9^{-21}$ & $50.9^{-18}$ \\
Fastt Crawl & $66.2^{3/-18}$ & $53.4^{9/-13}$ & $66.4^{3/-18}$ & $54.2^{19/-2}$ & $62.4^{-20}$ & $50.9^{-16}$ & $60.4^{-18}$ & $51.1^{-17}$ & $62.1^{2/-20}$ & $50.9^{-18}$ \\
Fastt Crawl M & $63.3^{1/-20}$ & $53.2^{8/-14}$ & $63.6^{1/-20}$ & $53.9^{18/-3}$ & $60.7^{-20}$ & $50.9^{-16}$ & $54.5^{-20}$ & $51.1^{-17}$ & $60.3^{-21}$ & $50.8^{-19}$ \\
\midrule
ELMoPubMed & $76.1^{11/-7}$ & $50.4^{-17}$ & $75.3^{11/-8}$ & $52.1^{-17}$ & $78.8^{4}$ & $51.3^{6/-14}$ & $74.0^{4}$ & $55.9^{8/-8}$ & $79.6^{7/-1}$ & $54.7^{7/-12}$ \\
Flair & $70.6^{6/-15}$ & $50.4^{-17}$ & $70.0^{6/-15}$ & $52.1^{-17}$ & $70.3^{3/-19}$ & $50.9^{-16}$ & $67.5^{2/-5}$ & $51.1^{-17}$ & $73.2^{3/-16}$ & $51.0^{-18}$ \\
SciBERT & $63.4^{1/-20}$ & $50.4^{-17}$ & $64.2^{1/-20}$ & $52.1^{-17}$ & $75.9^{4/-4}$ & $51.1^{6/-14}$ & $69.0^{3}$ & $53.8^{6/-14}$ & $72.2^{3/-17}$ & $54.3^{7/-12}$ \\
BERT & $67.4^{5/-17}$ & $50.4^{-17}$ & $67.2^{5/-17}$ & $52.1^{-17}$ & $78.5^{4}$ & $50.9^{-16}$ & $70.3^{3}$ & $51.3^{-17}$ & $75.3^{3/-10}$ & $51.7^{1/-17}$ \\
ELMo & $70.2^{6/-15}$ & $50.4^{-17}$ & $70.1^{6/-15}$ & $52.1^{-17}$ & $76.0^{4/-5}$ & $51.1^{-14}$ & $70.3^{3}$ & $53.4^{6/-14}$ & $76.0^{4/-9}$ & $53.0^{5/-16}$ \\
GPT & $65.9^{3/-18}$ & $50.4^{-17}$ & $65.8^{3/-18}$ & $52.1^{-17}$ & $77.0^{4/-1}$ & $50.9^{-16}$ & $68.7^{2}$ & $51.1^{-17}$ & $78.5^{5/-2}$ & $52.2^{4/-16}$ \\
    \end{tabular}     
    \caption{Accuracy of each embedding ($fJ$ for word embeddings, $\tau$ for contextual embeddings) on datasets created with Levenshtein negative sampling. 
    An embedding has significantly better/worse accuracy than the number of embeddings given by the positive/negative superscripts ($\alpha = 0.0002$, i.e. $\alpha = 0.05$ with Bonferroni correction).}
    \label{tab:significance_newDatasets_subset_negL} 
\end{table*}

\noindent\textbf{Embedding Comparison.} 
Table~\ref{tab:significance_existingDatasets} reports the Spearman's correlation for each embedding
 and indicates how many other embeddings it significantly outperforms and falls behind. 
Note that higher correlations have been reported for the UMNSRS-Sim/Rel datasets (e.g. \cite{LingEtAl2017,AbdeddaimVS2018}), but none of these embeddings are publicly available and thus not included here. 
Overall, the correlations of word embeddings are moderate to strong, \emph{suggesting} that embeddings are able to decently encode medical terms and their similarity.
For the Hliaoutakis and MiniMayo datasets,
\emph{no significant differences} between
biomedical and, in most cases, even the non-medical word embeddings are observed, despite correlation differences as large as $0.2$. 
This is due to the small size of these datasets and highlights the \emph{need for larger datasets} to obtain more meaningful embedding comparisons. 
On the UMNSRS-Sim/Rel datasets, the BioNLP embeddings perform significantly worse than most other biomedical embeddings, even though they achieve the highest correlations on the very small datasets.
This demonstrates that existing datasets \emph{do not allow any judgements} about the generalisability of embeddings to unseen similarity instances.
The other biomedical word embeddings do not exhibit significant differences and even the non-medical word embeddings do not perform significantly worse.
This raises the \emph{question if existing datasets are representative} of the highly difficult medical terminology.
All word embeddings significantly outperform
all contextual embeddings except ELMoPubMed, which performs significantly better than the other contextual embeddings. BERT models usually require fine tuning, so their lower performance is expected. Flair's lower performance likely stems from it not having an explicit notion of words, whereas the remaining contextual embeddings lack medical knowledge.

The results on Bio-SimLex and Bio-SimVerb are surprising: the non-medical Fasttext significantly outperforms all biomedical word embeddings, achieving much higher correlations  than previously reported \cite{ChiuPVK2018}.

\subsection{New Large-Scale Datasets}
Since our new datasets frame a binary classification task, we evaluate the embeddings' separability of similar versus dissimilar term pairs using the area under the ROC curve (AUC) and accuracy based on a classification threshold optimising the accuracy (different threshold for each embedding and similarity metric).
Significance between classifications of the different embeddings using the optimised thresholds is measured by McNemar's test.
Since the accuracy scores follow the AUC trends (see Tables~\ref{tab:newDatasets_auc_negL} and \ref{tab:newDatasets_auc_negR} in the Appendix), we present accuracy scores and their significant differences in Table~\ref{tab:significance_newDatasets_subset_negL}.

\noindent\textbf{Effects of Similarity Measures.}
In contrast to the existing datasets, $fJ$ \emph{significantly} outperforms most other similarity metrics for nearly all word embeddings (see Tables~\ref{tab:significance_snomedDatasets_methods1}-\ref{tab:significance_snomedDatasets_methods5} in the Appendix).
Furthermore, $pair$ metrics perform significantly worse than other metrics -- a difference not observable on the existing datasets.
For contextual embeddings, $\tau$ and and $\rho$ again significantly outperform the other metrics on some datasets.
For the following comparison of embeddings, we thus use $fJ$ for all word embeddings and $\tau$ for all contextual embeddings.

\noindent\textbf{Embedding Comparison.}
Table~\ref{tab:significance_newDatasets_subset_negL} shows \emph{significant} performance differences between the embeddings on the new datasets created with Levenshtein negative sampling, which are not revealed by existing datasets.
MeSH intr yields the best overall separation of similar and dissimilar term pairs, \emph{significantly} outperforming the non-medical word embeddings and the contextual embeddings as well as most of the medical word embeddings -- especially on the hard datasets.
In contrast, the performances of all medical word embeddings on the datasets with random negative sampling are very similar and high (see details in Tables~\ref{tab:significance_newDatasets_subset_negR} and \ref{tab:newDatasets_auc_negR} in the Appendix). 
This shows that, as is to be expected, random negative sampling creates term pairs that are easily identifiable as dissimilar. 
In the following, we thus focus on the datasets with Levenshtein negative sampling.

\noindent\textbf{Easy vs. Hard Datasets.}
For the hard datasets, accuracy is much lower than for the easy datasets, sometimes barely over 50\% indicating \emph{no separation} between similar and dissimilar term pairs.
Recall that in these hard datasets, similar term pairs have a larger Levenshtein distance than dissimilar ones (see Table~\ref{tab:datasets}), making them highly challenging. 
In fact, contextual embeddings predict dissimilar terms to be more similar than the actual similar terms (AUC lower than 0.5, see Table~\ref{tab:newDatasets_auc_negL} in the Appendix).
In contrast, for the easy datasets, where similar terms have a lower Levenshtein distance than dissimilar terms, the performance of ELMoPubMed is en par with the performance of some of the medical word embeddings.
This behaviour can be attributed to the fact that contextual embeddings are based on n-grams/characters, so that lexically similar medical terms are represented by similar vectors.

\noindent\textbf{Conclusion of Analysis.}
The performance analysis of embeddings on both existing and new datasets shows:
1) Our new datasets reveal \emph{significant} performance differences between embeddings and similarity metrics, not observable on the (too small) existing datasets.
2) Existing datasets suggest decent performance of current embeddings, whereas our datasets prove that embeddings are in fact \emph{unable} to correctly identify difficult term pairs as (dis)similar.
3) Our datasets thus provide a challenging novel \emph{benchmark} for future research, representing the whole medical terminology.

\subsection{Category Separation}
Both our new and existing datasets encode only very closely related terms as similar.
An adequate representation of medical terminology, mirroring a doctor's understanding, should however go further: medical terms are also similar on a broader level, forming distinct categories.
We thus propose to also use \emph{category separation} to test medical term representations and perform a first small evaluation to motivate this type of evaluation for future research.

Again, we make use of SNOMED CT and choose the two semantically close categories
\emph{Diagnostic Procedure} (DP) and \emph{Therapeutic Procedure} (TP) as well as the category \emph{Organism} (Org), which is semantically distant from the other two. 
Intuitively, we expect that terms (of concepts) in DP and TP are more similar than terms (of concepts) in DP and Org. 
To quantify to what extent an embedding satisfies this intuition, we introduce a category \emph{overlap} error metric
\[\#O = \sum_{t_i \in DP, t_j \in TP, t_k \in Org} 1 \mid sim(t_i, t_j) \leq sim(t_i,t_k)\]
counting the number of term pairs of semantically close categories that have lower $sim$(ilarity scores) than term pairs of distant categories. 
Since there may be OOV terms for some word embeddings, we report the \emph{relative overlap}, i.e.~the overlap error count compared to the maximum possible number of overlap errors,
$O = \#O  / (|DP| \times |TP| \times |Org|)$, where $|DP|$ (resp. $|TP|$, $|Org|$) denotes the number of terms in $DP$ that can be encoded by the respective embedding.
\citeauthor{DBLP:journals/corr/abs-1803-04488} \shortcite{DBLP:journals/corr/abs-1803-04488} use
a similar evaluation for non-medical terms, but apply a different metric. 

Table~\ref{tab:categorisation_evaluation} shows that, although contextual embeddings performed poorly on the term similarity task, ELMoPubMed achieves the best separation between categories (see more details in Table~\ref{tab:categorisation_evaluation_withSimScores} in the Appendix). 
Interestingly, the best performance of ELMoPubMed is achieved using $r$, whereas $\tau$ -- which performed best on the term similarity task -- produces the worst results. 
Furthermore, the MeSH embeddings, performing best on the term similarity task, exhibit comparably bad performance here.
These observations provide interesting first insights for future work.

\begin{table}[th]
    \centering
    \small
    \begin{tabular}{l l r}
         & metric (best/worst) & $O$ (best/worst)\\ 
         \midrule
LTL win2  & $fJ / pair\_\tau$ & $8.6\% / 20.1\%$ \\
AUEB200 & $fJ / mJ$ & $8.6\%/15.4\%$ \\
MeSH intr & $avg\_cos / pair\_\rho$ & $13.9\% / 17.1\%$ \\
MeSH extr & $avg\_cos / mJ$ & $12.2\% / 19.0\%$ \\
\midrule
ELMoPubMed & $r / \tau$  &$5.6\% /13.4\% $ \\
Flair & $r / \tau$ &$17.5\%/21.9\%$\\
SciBERT & $\rho / r$ &$21.1\% /24.4\%$\\
BERT & $\rho / cos$ &$10.9\%/11.9\%$ \\
ELMo & $r / \tau$ &$14.4\%/18.8\%$ \\
GPT & $\rho / r$ &$33.2\%/35.4\%$ \\
    \end{tabular}
    \caption{Relative overlap with best/worst similarity metric of 2 best word embeddings and 2 best from similarity task.}
    \label{tab:categorisation_evaluation} 
\end{table}

\section{Conclusion}
We have shown that existing datasets for medical term similarity are too small to detect significant performance differences between embeddings and similarity metrics applied to an embedding.
In contrast, using our new large-scale datasets, \emph{significant} differences are revealed.
Furthermore, the new datasets expose the enormous difficulty of current embeddings in predicting the similarity of non-obvious term pairs, i.e.~semantically similar terms that are lexically dissimilar and vice versa.
The datasets thus constitute a challenging \emph{new benchmark} for medical term similarity.
Our analysis also showed that the recently introduced Fuzzy Jaccard similarity measure for multi-word strings \cite{ZhelezniakEtAl2019-fuzzyJaccard} yields better results for most medical word embeddings than the standard cosine similarity and should thus receive attention in future work.
Overall, we conclude that available embeddings are \emph{unable} to adequately represent medical terminology at scale. 
In contrast to doctors' explicit knowledge of term (dis)similarity, as captured in ontologies such as SNOMED, embeddings are based on terms' occurrence in context, thus making similarity much more implicit.
We saw that embeddings making use of explicit knowledge (MeSH thesaurus) yield the best representations, which is thus a promising direction for future research.


\bibliographystyle{aaai}
\bibliography{conceptSimilarity}

\onecolumn
\newpage

\twocolumn
\appendix
\section*{Appendix}

\counterwithin{table}{section}
\counterwithin{figure}{section}

\noindent For the interested reader, this appendix contains more detailed results of our evaluations and analyses. \\

\begin{itemize}
    \item[A] \textbf{Word Embeddings}
    \begin{itemize}
        \item[] \textit{Table A.1}: Details about word embeddings used
        \item[] \textit{Table A.2}: Vocabulary overlap of word embeddings
    \end{itemize}
    \item[B] \textbf{Evaluation on Existing Datasets}
    \begin{itemize}
        \item[] \textit{Figure B.1}: Variance of correlation with different similarity metrics
        \item[] \textit{Table B.1}: Correlation with best similarity metrics for each embedding and dataset
        \item[] \textit{Tables B.2-5}: Significance of correlation differences for an embedding with different similarity metrics
    \end{itemize}
    \item[C] \textbf{Evaluation on New Large-Scale Datasets}
    \begin{itemize}
        \item[] \textit{Table C.1}: Accuracy results and significance of differences for datasets with random negative sampling
        \item[] \textit{Figure C.1}: Variance of accuracy with different similarity metrics
        \item[] \textit{Tables C.2-3}: AUC results
        \item[] \textit{Tables C.4-8}: Significance of accuracy differences for an embedding with different similarity metrics
    \end{itemize}
    \item[D] \textbf{Evaluation on Category Separation}
    \begin{itemize}
        \item[] \textit{Table D.1}: Relative overlap error and average in- and cross-category similarity scores
    \end{itemize}
\end{itemize}

\newpage

\section*{Acknowledgements}
We would like to thank Alex Szolnoki, Rebecca Sells, and Yun-Hsuan Chang for participating in the manual dataset evaluation, and Vitalii Zhelezniak for many useful discussions.
Our thanks also go to Nils Hammerla and Dane Sherburn for useful comments on the paper draft, and Adam Bozson for his support with running experiments.

\onecolumn

\section{Word Embeddings}
Table~\ref{tab:embeddings} compares characteristics of the different word embeddings used in our experiments.
It also summarises how well the vocabulary of the different embeddings covers the existing as well as our new datasets.
This is measured as the percentage of concept pairs without any out-of-vocabulary (OOV) words.
The coverage on existing datasets is much higher than for our new ones, illustrating that our new datasets encode much more of the enormous medical terminology, including less frequently used concepts, making these new datasets much more challenging than existing ones.
Note that all embeddings have 100\% coverage on the MiniMayo and Hliaoutakis dataset, which is thus not included in the statistics given in Table~\ref{tab:embeddings}.
Furthermore, the Fasttext model (M) embeddings have 100\% coverage on all datasets as they create embeddings for any word on-the-fly.

\begin{table}[h]
    \centering
    \small
    \begin{tabular}{l c r c r c r}
    \toprule
         Embedding & Date & Size & Corpus & Tokens & Method & Coverage (\%) \\
         \midrule
         BioNLP PMC\footnote{\url{http://bio.nlplab.org/}} & \multirow{4}{*}{2013} & 2,515,686 x 200 & PMC & 2.6B & 
         \multirow{4}{*}{w2v SG, win 5} & 75--99 /  51--86\\
         BioNLP PM & & 2,351,706 x 200 & PM & 2.9B & & 84--99 / 53--90\\ 
         BioNLP PP & & 4,087,446 x 200 & PMC, PM & 5.5B & & 87--100 / 56--91\\ 
         BioNLP PPW & & 5,443,656 x 200 & PMC, PM, Wiki & N/A & & 88--100 / 58--93\\ \midrule[.01em]
         BioASQ\footnote{ \url{http://bioasq.org/news/bioasq-releases-continuous-space-word-vectors-obtained-applying-word2vec-pubmed-abstracts}} & 2015 & 1,701,632 x 200 & BioASQ (PM) & N/A & w2v SG, win5 & 90--100 / 66--94 \\ \midrule[.01em]
         LTL win2\footnote{\url{https://github.com/cambridgeltl/BioNLP-2016}} &  \multirow{2}{*}{2016} & \multirow{2}{*}{2,231,686 x 200} &  \multirow{2}{*}{PM} &  \multirow{2}{*}{2.7B}
         & w2v SG, win 2 & \multirow{2}{*}{82--99 / 53--90} \\
         LTL win30 & & & & & w2v SG, win 30 \\ \midrule[.01em]
         AUEB200\footnote{\url{http://nlp.cs.aueb.gr/software.html}} & \multirow{2}{*}{2018} & \multirow{2}{*}{2,665,547 x 200} &\multirow{2}{*}{PM} &  \multirow{2}{*}{3.6B}
         & \multirow{2}{*}{w2v SG, win 5} & \multirow{2}{*}{93--100 / 69--96}\\
         AUEB400  \\ \midrule[.01em]
         MeSH extr\footnote{\url{https://github.com/ncbi-nlp/BioWordVec}} & \multirow{2}{*}{2019} & \multirow{2}{*}{2,324,849 x 200} & \multirow{2}{*}{PM + MeSH} & 3.7B
         & Fasttext, win 20 & \multirow{2}{*}{92--100 / 69--96}\\
         MeSH intr & & & & + 28M
         & Fasttext, win 5 \\
         MIMIC\footnote{\url{https://github.com/ncbi-nlp/BioSentVec}} & \multirow{2}{*}{2019} & 16,545,452 x 200 & \multirow{2}{*}{PM, MIMIC-III + MeSH} & 4.4B
         & \multirow{2}{*}{Fasttext, win 20} & 96--100 / 78--98\\
         MIMIC M & & on-the-fly x 200 & & + 0.5B & & 100 / 100\\
         \midrule
         GloVe\footnote{\url{https://nlp.stanford.edu/projects/glove/}} & 2014 & 2,196,016 x 300 & Common Crawl & 42B & GloVe & 84--99 / 45--82\\
         Fasttext Wiki\footnote{\url{https://fasttext.cc/docs/en/english-vectors.html}} & 2018 & 999,994 x 300 & Wiki, UMBC WebBase, & 16.6B & Fasttext, win 15 & 60--97 / 41--75\\
         & & & statmt.org News \\
         Fasttext Crawl& \multirow{2}{*}{2018} & 2,000,000 x 300 & \multirow{2}{*}{Common Crawl} & \multirow{2}{*}{630B}& \multirow{2}{*}{Fasttext, win 15} & 81--99 / 45--81\\
         Fasttext Crawl M &  & on-the-fly x 300 & & & & 100 / 100\\
         \bottomrule
    \end{tabular}
    \caption{Word Embeddings used here: release date, vocabulary size and vector dimensionality, training corpus and number of tokens therein, embedding method (win = window size), and coverage (non-OOV) as min--max for existing / our new datasets.}
    \label{tab:embeddings}
\end{table}

Interestingly, the AUEB embeddings have a much better coverage than BioNLP PM and LTL even though all are trained on a very similar corpus, i.e.~PubMed from different years, using the same method.
Closer inspection of the different embeddings' vocabulary overlap, illustrated in  Table~\ref{tab:embedding_coverage}, shows that BioNLP PM and LTL share over 80\% of their vocabulary. In contrast, AUEB only covers 33/35\% of the BioNLP PM/LTL vocabulary. The better dataset coverage of AUEB is thus not due to the larger vocabulary size compared to BioNLP and LTL but rather to the words included in it.

From Table~\ref{tab:embedding_coverage} we further observe that the embeddings can be split into three groups with large vocabulary overlaps:
1) The different BioNLP embeddings and LTL, 2) BioASQ, AUEB, MeSH and MIMIC, and 3) the non-medical GloVe and Fasttext embeddings.
Note that AUEB contains 90\% of the vocabulary of MeSH whereas vice versa the coverage is only 78\%, even though MeSH creates word vectors from PubMed (used for AUEB) enriched with MeSH data. Due to the large vocabulary overlap of  AUEB and MeSH, their coverage on both existing and our new datasets is nearly the same.

The embedding with the largest vocabulary is MIMIC, which also has the best dataset coverage. The eight times larger vocabulary results in a 10 percentage points higher minimum coverage than the next best embeddings (AUEB and MeSH).
As expected, the non-medical GloVe and Fasttext embeddings have low dataset coverage.
Furthermore, they only cover a very small percentage of the medical embeddings' vocabulary.

\begin{sidewaystable}[h]
    \centering
    \small
    \begin{tabular}{l | r r r r r r r r r r r r | r r}
    & PMC & PM & PP & PPW & ASQ & LTL & AUEB & MeSH & MIMIC & GloVe & Fastt W & Fastt C & Av & Av-Med \\ \midrule
PMC  & & 47.53 & 61.33 & 46.07 & 25.37 & 49.67 & 18.74 & 21.30 & 4.76 & 24.62 & 40.59 & 27.70 & 33.43 & 34.35\\
PM & 44.43 &  & 57.53 & 43.20 & 38.97 & 86.97 & 29.82 & 34.39 & 6.13 & 21.86 & 36.06 & 24.74 & 38.56 & 42.68\\
PP & 99.65 & 100.00 & & 74.42 & 42.09 & 91.39 & 34.99 & 40.06 & 8.68 & 29.99 & 47.2 & 33.16 & 54.69 & 61.41\\
PPW & 99.70 & 100.00 & 99.11 &  & 43.26 & 91.68 & 36.51 & 41.75 & 9.63 & 50.21 & 82.01 & 54.97 & \textbf{64.44} & \textbf{65.21}\\
ASQ & 17.16 & 28.19 & 17.51 & 13.52 &  & 28.94 & 62.82 & 59.64 & 8.76 & 11.60 & 17.16 & 12.94 & 25.29 & 29.57 \\
LTL & 44.06 & 82.53 & 49.90 & 37.58 & 37.97 & & 29.06 & 33.66 & 6.01 & 21.74 & 35.97 & 24.67 & 36.65 & 40.10\\
AUEB & 19.86 & 33.80 & 22.82 & 17.88 & 98.45 & 34.71 &  & 89.83 & 13.30 & 14.34 & 20.29 & 15.8 & 34.64 & 41.33 \\
MeSH & 19.68 & 33.99 & 22.79 & 17.83 & 81.52 & 35.07 & 78.34 &  & 13.44 & 13.08 & 20.28 & 15.17 & 31.93 & 37.83\\
MIMIC & 31.28 & 43.11 & 35.13 & 29.28 & 85.20 & 44.58 & 82.58 & 95.67 &  & 23.62 & 28.60 & 24.36 & 47.59 & 55.86\\
GloVe & 21.49 & 20.41 & 16.11 & 20.26 & 14.97 & 21.39 & 11.82 & 12.36 & 3.14 &  & 78.39 & 68.21 & 26.23 & 17.74\\
Fastt W & 16.13 & 15.34 & 11.55 & 15.07 & 10.09 & 16.12 & 7.61 & 8.72 & 1.73 & 35.70 &  & 40.82 & 16.26 & 12.79\\
Fastt C & 22.02 & 21.04 & 16.22 & 20.20 & 15.21 & 22.11 & 11.85 & 13.05 & 2.94 & 62.12 & 81.63 & &  26.22 & 18.08\\
    \end{tabular}
    \caption{Vocabulary overlap of word embeddings: coverage (in \%) of the row embedding vocabularies regarding the column embedding vocabularies, i.e.~\% of concepts in the column vocabulary that also occurs in the row vocabulary. The last two columns give the average coverage of a row embeddings regarding all embeddings and all medical embeddings.}
    \label{tab:embedding_coverage}
\end{sidewaystable}

\clearpage
\section{Evaluation on Existing Datasets}
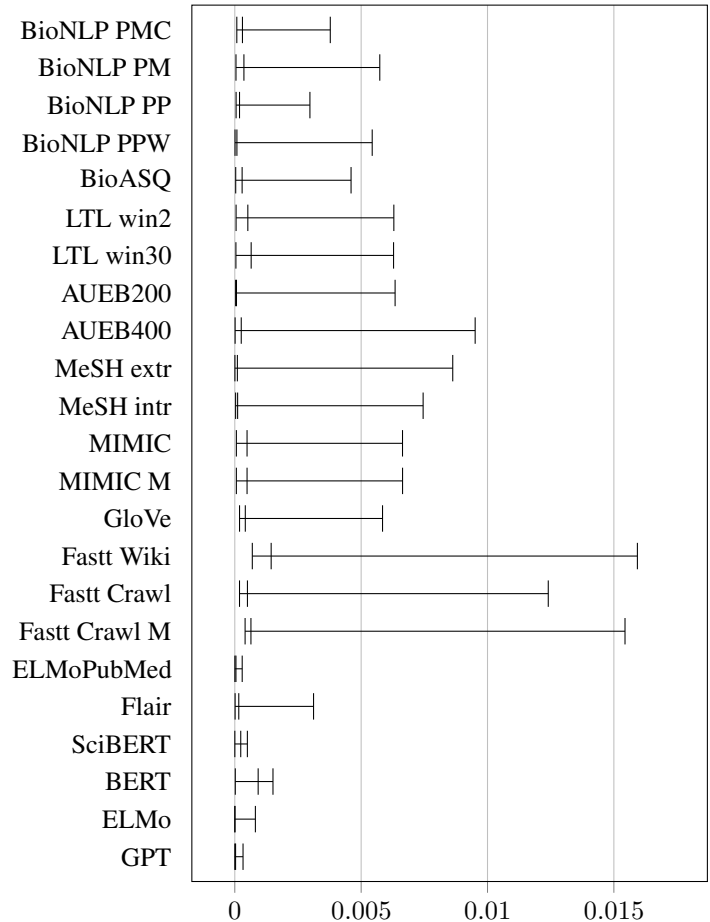
\begin{figure}[h]
    \centering
\begin{tikzpicture}
\begin{axis}[symbolic y coords={
                GPT,
                ELMo,
                BERT,
                SciBERT,
                Flair,
                ELMoPubMed,
                Fastt Crawl M,
                Fastt Crawl,
                Fastt Wiki,
                GloVe,
                MIMIC M, 
                MIMIC,
                MeSH intr,
                MeSH extr,
                AUEB400,
                AUEB200,
                LTL win30,
                LTL win2,
                BioASQ,
                BioNLP PPW,
                BioNLP PP,
                BioNLP PM,
                BioNLP PMC
                },
                ytick={
                GPT,
                ELMo,
                BERT,
                SciBERT,
                Flair,
                ELMoPubMed,
                Fastt Crawl M,
                Fastt Crawl,
                Fastt Wiki,
                GloVe,
                MIMIC M, 
                MIMIC,
                MeSH intr,
                MeSH extr,
                AUEB400,
                AUEB200,
                LTL win30,
                LTL win2,
                BioASQ,
                BioNLP PPW,
                BioNLP PP,
                BioNLP PM,
                BioNLP PMC
                },
                y=0.5cm, enlarge y limits ={true, value=0.03}, enlarge x limits=true, ytick=data, xtick align=outside, ytick align=outside, tick pos=left, ytick style = {draw=none}, xmajorgrids = true,
  x tick label style=
{/pgf/number format/fixed, 
/pgf/number format/precision=3,
scaled x ticks = false}
  ]
\addplot[white,only marks] coordinates{(0.017,SciBERT)(0.017,BioASQ)(0.017,Fastt Crawl M)(0.017,MeSH extr)(0.017,GloVe)(0.017,Fastt Wiki)(0.017,Fastt Crawl)(0.017,MeSH intr)(0.017,MIMIC)(0.017,MIMIC M)(0.017,BioNLP PMC)(0.017,BioNLP PM)(0.017,BioNLP PP)(0.017,BioNLP PPW)(0.017,LTL win2)(0.017,LTL win30)(0.017,AUEB200)(0.017,AUEB400)(0.017,BERT)(0.017,ELMo)(0.017,ELMoPubMed)(0.017,Flair)(0.017,GPT)};

\addplot[mark=|, mark size = 5pt] coordinates {(7.978486202568047e-05,BioNLP PMC) (0.0003028623532845101,BioNLP PMC) (0.003778328086302747,BioNLP PMC)};
\addplot[mark=|, mark size = 5pt] coordinates {(4.6898621936173715e-05,BioNLP PM) (0.00036097786186608713,BioNLP PM) (0.0057372119165277485,BioNLP PM)};
\addplot[mark=|, mark size = 5pt] coordinates {(5.1385355191435167e-05,BioNLP PP) (0.00018734657629580485,BioNLP PP) (0.0029727218546013183,BioNLP PP)};
\addplot[mark=|, mark size = 5pt] coordinates {(9.029637323963518e-06,BioNLP PPW) (8.380101545440594e-05,BioNLP PPW) (0.005437306978614955,BioNLP PPW)};
\addplot[mark=|, mark size = 5pt] coordinates {(3.642272850817202e-05,BioASQ) (0.00028988389758505923,BioASQ) (0.004600465222426633,BioASQ)};
\addplot[mark=|, mark size = 5pt] coordinates {(5.112910107689137e-05,LTL win2) (0.0005158030663810192,LTL win2) (0.006293608312265542,LTL win2)};
\addplot[mark=|, mark size = 5pt] coordinates {(4.4331563978753435e-05,LTL win30) (0.0006477287728881755,LTL win30) (0.006282643862686034,LTL win30)};
\addplot[mark=|, mark size = 5pt] coordinates {(3.2899998663112944e-05,AUEB200) (6.666112601880676e-05,AUEB200) (0.00634324683229031,AUEB200)};
\addplot[mark=|, mark size = 5pt] coordinates {(1.2263882180469292e-05,AUEB400) (0.00025503442589986915,AUEB400) (0.00951113445853835,AUEB400)};
\addplot[mark=|, mark size = 5pt] coordinates {(4.9101078185253315e-06,MeSH extr) (0.00010186374953120048,MeSH extr) (0.008624437864910519,MeSH extr)};
\addplot[mark=|, mark size = 5pt] coordinates {(3.283988084823803e-05,MeSH intr) (0.00011208667038742671,MeSH intr) (0.007452198890905878,MeSH intr)};
\addplot[mark=|, mark size = 5pt] coordinates {(6.225592284383071e-05,MIMIC) (0.00048504102768578793,MIMIC) (0.006636825509680465,MIMIC)};
\addplot[mark=|, mark size = 5pt] coordinates {(6.224335223118666e-05,MIMIC M) (0.000485324294933379,MIMIC M) (0.006638833657134865,MIMIC M)};
\addplot[mark=|, mark size = 5pt] coordinates {(0.00018816243195409924,GloVe) (0.00041549071536214536,GloVe) (0.0058466309836667,GloVe)};
\addplot[mark=|, mark size = 5pt] coordinates {(0.0006928915365017656,Fastt Wiki) (0.0014388505422559346,Fastt Wiki) (0.01593055474495457,Fastt Wiki)};
\addplot[mark=|, mark size = 5pt] coordinates {(0.00018682529516106275,Fastt Crawl) (0.0004999499052559881,Fastt Crawl) (0.012403746208818208,Fastt Crawl)};
\addplot[mark=|, mark size = 5pt] coordinates {(0.00040762333296780667,Fastt Crawl M) (0.0006387397916503436,Fastt Crawl M) (0.015440769628030041,Fastt Crawl M)};
\addplot[mark=|, mark size = 5pt] coordinates {(2.9377713265507807e-06,ELMoPubMed) (4.5534854562522086e-05,ELMoPubMed) (0.0002929251675049423,ELMoPubMed)};
\addplot[mark=|, mark size = 5pt] coordinates {(1.189520068873285e-05,Flair) (0.0001576290945099139,Flair) (0.003116858534292471,Flair)};
\addplot[mark=|, mark size = 5pt] coordinates {(1.3556856681453546e-06,SciBERT) (0.00023548240660637034,SciBERT) (0.0004974092571958428,SciBERT)};
\addplot[mark=|, mark size = 5pt] coordinates {(1.9607398020387293e-05,BERT) (0.0009245395475553244,BERT) (0.0015109012914048424,BERT)};
\addplot[mark=|, mark size = 5pt] coordinates {(7.896302913929762e-07,ELMo) (6.518092379938733e-06,ELMo) (0.0008144470246385535,ELMo)};
\addplot[mark=|, mark size = 5pt] coordinates {(7.387330949625146e-07,GPT) (3.1371005682677705e-05,GPT) (0.00032615382081642436,GPT)};
\end{axis}
\end{tikzpicture}
    \caption{Min/median/max variance between the Spearman's correlations of the different similarity metrics applied to an embedding observed for an existing dataset.}
    \label{fig:variance_existingDatasets}
\end{figure}

Table~\ref{tab:significance_existingDatasets_full} reports the Spearman's correlations of embeddings on existing datasets using the similarity metric resulting in the highest correlation for each embedding and dataset.
We observe that there is no single similarity metric that is consistently the best for an embedding (i.e. the best on all datasets). 
Figure~\ref{fig:variance_existingDatasets} illustrates the variance of correlation obtained with the different similarity metrics. We observe that the contextual embeddings are barely affected by the choice of similarity metric, whereas the non-medical fasttext embeddings are more influenced by it. 
Tables~\ref{tab:significance_existingDatasets_methods1}-\ref{tab:significance_existingDatasets_methods4} furthermore detail for each embedding, which similarity metrics perform significantly worse or better than others for each dataset.
All analyses are performed on subsets of each dataset that have no OOV concepts for any embedding.

Note that the MiniMayo dataset has been annotated twice: by medical \emph{coders} (c) and by \emph{physicians} (p). We report performance on both as well as regarding the average (av) of the two scores.

\begin{table}
    \centering
    \footnotesize
    \begin{adjustbox}{angle=90}
    \begin{tabular}{l l l l l l l l l l l l}
         Dataset & Hlia. & MM-av & MM-c & MM-p & Mayo & Sim & Sim-m & Rel & Rel-m & SimLex & SimVerb\\ 
Subset Size & 36/36 & 29/29 & 29/29 & 29/29 & 81/101 & 352/566 & 340/449 & 347/587 & 339/458 & 964/988 & 909/1000\\
\midrule
BioNLP PMC & $0.53^{1}$ & $0.83^{4}$ & $0.77^{2}$ & $0.82^{4}$ & $0.48^{3/-3}$ & $0.48^{5/-12}$ & $0.46^{5/-14}$ & $0.36^{4/-15}$ & $0.36^{4/-16}$ & $0.71^{5/-3}$ & $0.45^{4/-4}$ \\
 & $avg\_cos$ & $pair\_\tau$ & $pair\_\tau$ & $pair\_\tau$ & $pair\_r$ & $avg\_\rho$ & $pair\_r$ & $fJ$ & $pair\_r$ & $pair\_cos$ & $pair\_r$ \\
BioNLP PM & $0.59^{3}$ & $0.86^{5}$ & $0.84^{3}$ & $0.83^{4}$ & $0.57^{5}$ & $0.58^{7/-2}$ & $0.56^{7/-4}$ & $0.47^{7/-4}$ & $0.48^{7/-4}$ & $0.69^{4/-4}$ & $0.44^{4/-5}$ \\
 & $avg\_cos$ & $pair\_r$ & $fJ$ & $pair\_r$ & $pair\_\rho$ & $avg\_\rho$ & $pair\_cos$ & $avg\_cos$ & $pair\_cos$ & $fJ$ & $pair\_cos$ \\
BioNLP PP & $0.59^{2}$ & $0.81^{3}$ & $0.8^{2}$ & $0.79^{3}$ & $0.52^{3}$ & $0.54^{6/-9}$ & $0.53^{6/-9}$ & $0.44^{7/-7}$ & $0.45^{7/-7}$ & $0.71^{5/-3}$ & $0.45^{4/-5}$ \\
 & $avg\_cos$ & $pair\_\tau$ & $pair\_\tau$ & $pair\_\tau$ & $pair\_r$ & $avg\_cos$ & $pair\_cos$ & $avg\_cos$ & $pair\_cos$ & $fJ$ & $pair\_r$ \\
BioNLP PPW & $0.6^{2}$ & $0.86^{3}$ & $0.82^{5}$ & $0.83^{5}$ & $0.52^{3}$ & $0.55^{7/-8}$ & $0.53^{7/-8}$ & $0.45^{7/-6}$ & $0.45^{7/-7}$ & $0.72^{7/-2}$ & $0.47^{5/-4}$ \\
 & $fJ$ & $avg\_\rho$ & $pair\_\tau$ & $avg\_\rho$ & $pair\_\rho$ & $avg\_\rho$ & $pair\_r$ & $avg\_\tau$ & $pair\_\rho$ & $fJ$ & $pair\_r$ \\
BioASQ & $0.49^{1}$ & $0.83^{4}$ & $0.81^{3}$ & $0.79^{3}$ & $0.58^{4}$ & $0.6^{9/-1}$ & $0.59^{9/-1}$ & $0.48^{7/-4}$ & $0.49^{7/-4}$ & $0.69^{4/-5}$ & $0.42^{3/-11}$ \\
& $avg\_\rho$ & $fJ$ & $fJ$ & $pair\_\rho$ & $pair\_\rho$ & $fJ$ & $pair\_r$ & $avg\_\rho$ & $fJ$ & $pair\_cos$ & $pair\_\tau$ \\
LTL win2 & $0.55^{2}$ & $0.8^{3}$ & $0.78^{2}$ & $0.78^{3}$ & $0.53^{4}$ & $0.61^{8/-1}$ & $0.59^{8/-1}$ & $0.5^{7/-2}$ & $0.51^{7/-1}$ & $0.72^{6/-2}$ & $0.46^{5/-4}$ \\
& $avg\_\tau$ & $pair\_\tau$ & $pair\_\tau$ & $pair\_\tau$ & $pair\_r$ & $avg\_\tau$ & $pair\_\tau$ & $avg\_\tau$ & $pair\_\tau$ & $pair\_cos$ & $fJ$ \\
LTL win30 & $0.62^{3}$ & $0.83^{3}$ & $0.81^{2}$ & $0.81^{3}$ & $0.64^{8}$ & $0.67^{14}$ & $0.66^{13}$ & $0.58^{12}$ & $0.6^{12}$ & $0.69^{4/-4}$ & $0.44^{4/-6}$ \\
& $fJ$ & $avg\_r$ & $pair\_\rho$ & $avg\_r$ & $pair\_\rho$ & $avg\_\tau$ & $pair\_\tau$ & $avg\_\tau$ & $pair\_\tau$ & $pair\_cos$ & $pair\_cos$ \\
AUEB200 & $0.46$ & $0.81^{4}$ & $0.79^{3}$ & $0.79^{3}$ & $0.59^{4}$ & $0.63^{9}$ & $0.62^{9}$ & $0.54^{9/-1}$ & $0.55^{9}$ & $0.71^{6/-3}$ & $0.46^{4/-4}$ \\
& $fJ$ & $pair\_\tau$ & $pair\_\tau$ & $pair\_\tau$ & $pair\_\rho$ & $avg\_r$ & $pair\_\rho$ & $avg\_\tau$ & $pair\_\tau$ & $pair\_r$ & $fJ$ \\
AUEB400 & $0.5^{2}$ & $0.82^{3}$ & $0.79^{5}$ & $0.79^{2}$ & $0.6^{5}$ & $0.64^{9}$ & $0.63^{9}$ & $0.54^{8/-1}$ & $0.55^{9}$ & $0.72^{6/-2}$ & $0.47^{5/-4}$ \\
 & $pair\_\rho$ & $pair\_\rho$ & $pair\_\tau$ & $pair\_\rho$ & $pair\_\rho$ & $avg\_cos$ & $pair\_cos$ & $avg\_cos$ & $fJ$ & $pair\_cos$ & $fJ$ \\
MeSH extr & $0.5^{1}$ & $0.83^{5}$ & $0.8^{5}$ & $0.8^{4}$ & $0.61^{5}$ & $0.63^{9/-1}$ & $0.62^{9/-1}$ & $0.55^{9/-1}$ & $0.55^{9/-1}$ & $0.7^{5/-4}$ & $0.47^{5/-4}$ \\
 & $fJ$ & $fJ$ & $fJ$ & $fJ$ & $pair\_cos$ & $avg\_\rho$ & $pair\_r$ & $avg\_\rho$ & $pair\_r$ & $fJ$ & $pair\_cos$ \\
MeSH intr & $0.47^{1}$ & $0.84^{5}$ & $0.82^{5}$ & $0.81^{4}$ & $0.66^{8}$ & $0.66^{13}$ & $0.65^{12}$ & $0.59^{15}$ & $0.59^{12}$ & $0.66^{4/-14}$ & $0.45^{4/-4}$ \\
 & $pair\_\tau$ & $pair\_\tau$ & $pair\_\tau$ & $pair\_\tau$ & $pair\_cos$ & $avg\_cos$ & $pair\_cos$ & $avg\_cos$ & $pair\_cos$ & $fJ$ & $pair\_\rho$ \\
MIMIC & $0.54^{1}$ & $0.84^{4}$ & $0.8^{2}$ & $0.82^{4}$ & $0.6^{5}$ & $0.64^{9}$ & $0.63^{10}$ & $0.56^{11}$ & $0.57^{11}$ & $0.71^{5/-3}$ & $0.49^{6/-3}$ \\
 & $pair\_r$ & $pair\_\tau$ & $pair\_\tau$ & $pair\_\tau$ & $pair\_\rho$ & $avg\_cos$ & $avg\_cos$ & $avg\_cos$ & $pair\_cos$ & $fJ$ & $pair\_\tau$ \\
MIMIC M & $0.57^{2}$ & $0.84^{3}$ & $0.8^{2}$ & $0.82^{4}$ & $0.6^{5}$ & $0.64^{9}$ & $0.63^{10}$ & $0.56^{11}$ & $0.57^{11}$ & $0.71^{5/-3}$ & $0.49^{6/-3}$ \\
 & $pair\_r$ & $pair\_\tau$ &  & $pair\_cos$ & $avg\_cos$ & $pair\_cos$ & $fJ$ & $pair\_\tau$ \\
\midrule
GloVe & $0.37^{-1}$ & $0.62^{-6}$ & $0.61^{-4}$ & $0.59^{1/-6}$ & $0.43^{1/-2}$ & $0.55^{5/-2}$ & $0.54^{5/-2}$ & $0.49^{7}$ & $0.49^{7}$ & $0.75^{10/-2}$ & $0.56^{15}$ \\
& $avg\_r$ & $pair\_\tau$ & $pair\_\tau$ & $pair\_\tau$ & $pair\_\tau$ & $avg\_\tau$ & $pair\_\tau$ & $avg\_r$ & $pair\_\tau$ & $pair\_\tau$ & $pair\_\tau$ \\
Fastt Wiki & $0.37^{1}$ & $0.68^{1}$ & $0.66$ & $0.64^{-1}$ & $0.48^{2}$ & $0.54^{5/-2}$ & $0.56^{6}$ & $0.49^{6}$ & $0.52^{7}$ & $0.76^{15}$ & $0.59^{19}$ \\
 & $pair\_r$ & $pair\_\rho$ & $pair\_cos$ & $pair\_\rho$ & $pair\_\rho$ & $pair\_\rho$ & $pair\_\rho$ & $pair\_cos$ & $pair\_cos$ & $pair\_\rho$ & $pair\_\rho$ \\
Fastt Crawl & $0.38$ & $0.69^{1}$ & $0.68^{1/-5}$ & $0.66^{1}$ & $0.48^{2}$ & $0.59^{6}$ & $0.59^{7}$ & $0.54^{7}$ & $0.55^{7}$ & $0.78^{20}$ & $0.6^{19}$ \\
 & $pair\_cos$ & $pair\_\tau$ & $pair\_\tau$ & $pair\_\tau$ & $pair\_cos$ & $avg\_\rho$ & $pair\_r$ & $avg\_\rho$ & $pair\_r$ & $pair\_\tau$ & $pair\_\tau$ \\
Fastt Crawl M & $0.38^{1}$ & $0.69^{2}$ & $0.68^{1/-5}$ & $0.66^{1}$ & $0.48^{2}$ & $0.58^{6}$ & $0.59^{7}$ & $0.54^{7}$ & $0.55^{7}$ & $0.78^{20}$ & $0.6^{19}$ \\
 & $pair\_cos$ & $pair\_\tau$ & $pair\_\tau$ & $pair\_\tau$ & r & $avg\_\rho$ & $pair\_r$ & $pair\_\tau$ & $pair\_\tau$ \\
\midrule
ELMoPubMed & $0.43^{2}$ & $0.67$ & $0.65$ & $0.65$ & $0.38^{1/-6}$ & $0.44^{5/-13}$ & $0.42^{5/-14}$ & $0.33^{4/-16}$ & $0.34^{4/-16}$ & $0.72^{6/-2}$ & $0.52^{7/-3}$ \\
 & $\rho$ & $cos$ & $\tau$ & $cos$ & $\rho$ & $\rho$ & $\rho$ & $\rho$ & $\rho$ & $\rho$ & $\rho$ \\
Flair & $-0.0^{-12}$ & $0.14^{-10}$ & $0.21^{-3}$ & $0.12^{-11}$ & $0.18^{-10}$ & $0.19^{-18}$ & $0.2^{-18}$ & $0.08^{-18}$ & $0.09^{-18}$ & $0.38^{-19}$ & $0.19^{-19}$ \\
 & $cos$ & $cos$ & $cos$ & $cos$ & $\tau$ & $\rho$ & $\rho$ & $\rho$ & $\rho$ & $\rho$ & $\rho$ \\
SciBERT & $0.4$ & $0.59$ & $0.57$ & $0.59$ & $0.27^{-2}$ & $0.19^{-18}$ & $0.19^{-18}$ & $0.21^{-16}$ & $0.21^{-16}$ & $0.35^{-19}$ & $0.3^{-18}$ \\
& $\tau$ & $\tau$ & $\rho$ & $\tau$ & $cos$ & $\tau$ & $\tau$ & $\tau$ & $\rho$ & $\tau$ & $\tau$ \\
BERT & $0.05$ & $0.21^{-8}$ & $0.27$ & $0.19^{-2}$ & $0.05^{-13}$ & $0.12^{-18}$ & $0.11^{-18}$ & $0.08^{-18}$ & $0.04^{-18}$ & $0.39^{-19}$ & $0.18^{-19}$ \\
 & $cos$ & $\tau$ & $\tau$ & $\tau$ & $cos$ & $\tau$ & $\tau$ & $\tau$ & $\tau$ & $\tau$ & $\rho$ \\
ELMo & $0.0^{-11}$ & $0.15^{-15}$ & $0.16^{-15}$ & $0.15^{-15}$ & $0.08^{-18}$ & $0.2^{-18}$ & $0.21^{-18}$ & $0.13^{-18}$ & $0.14^{-18}$ & $0.64^{4/-9}$ & $0.51^{4/-3}$ \\
& $cos$ & $r$ & $r$ & $r$ & $\rho$ & $cos$ & $\rho$ & $r$ & $cos$ & $\rho$ & $cos$ \\
GPT & $0.01^{-1}$ & $-0.17^{-14}$ & $-0.12^{-11}$ & $-0.18^{-14}$ & $0.01^{-16}$ & $0.1^{-18}$ & $0.09^{-18}$ & $0.08^{-18}$ & $0.06^{-18}$ & $0.32^{-19}$ & $0.26^{-19}$ \\
 & $\rho$ & $\tau$ & $\tau$ & $\tau$ & $\tau$ & $\tau$ & $\tau$ & $\tau$ & $\tau$ & $cos$ & $\rho$ \\
    \end{tabular}
    \end{adjustbox}
    \caption{Spearman's Correlation (1st row) of the similarity metric (2nd row) achieving highest correlation for an embedding and dataset. An embedding (with chosen similarity method) has significantly better/worse correlation than the number of embeddings given by the positive/negative superscripts. Significance measured by bias-corrected and accelerated (BCa)
bootstrap confidence intervals with $\alpha = 0.0002$, i.e. $\alpha = 0.05$ with Bonferroni correction).}
    \label{tab:significance_existingDatasets_full} 
\end{table}

\begin{sidewaystable*}
    \centering
    \small
    \begin{tabular}{l l l l l l l l l l l l}
    \toprule
         Dataset & Hlia. & MM-av & MM-c & MM-p & Mayo & Sim & Sim-m & Rel & Rel-m & SimLex & SimVerb\\ 
Subset Size & 36/36 & 29/29 & 29/29 & 29/29 & 81/101 & 352/566 & 340/449 & 347/587 & 339/458 & 964/988 & 909/1000\\
\midrule
\multirow{6}{1.5cm}{BioNLP PMC} & \multirow{6}{1.5cm}{} & \multirow{6}{1.5cm}{} & \multirow{6}{1.5cm}{} & \multirow{6}{1.5cm}{} & \multirow{6}{1.5cm}{$fJ^{1}$, $mJ^{-5}$, $cA^{1}$, $r A^{1}$, $\rho A^{1}$, $\tau A^{1}$} & \multirow{6}{1.5cm}{$cP^{1}$, $r P^{1}$, $\rho P^{1}$, $\tau P^{1}$, $fJ^{1}$, $mJ^{-9}$, $cA^{1}$, $r A^{1}$, $\rho A^{1}$, $\tau A^{1}$} & \multirow{6}{1.5cm}{$cP^{1}$, $r P^{1}$, $\rho P^{1}$, $\tau P^{1}$, $fJ^{1}$, $mJ^{-9}$, $cA^{1}$, $r A^{1}$, $\rho A^{1}$, $\tau A^{1}$} & \multirow{6}{1.5cm}{} & \multirow{6}{1.5cm}{$fJ^{1}$, $mJ^{-1}$} & \multirow{6}{1.5cm}{$cP^{1}$, $r P^{1}$, $\rho P^{1}$, $\tau P^{1}$, $fJ^{1}$, $mJ^{-9}$, $cA^{1}$, $r A^{1}$, $\rho A^{1}$, $\tau A^{1}$} & \multirow{6}{1.5cm}{$mJ^{-1}$, $\rho A^{1}$} \\ \\ \\ \\ \\ \\
\midrule
\multirow{6}{1.5cm}{BioNLP PM} & \multirow{6}{1.5cm}{} & \multirow{6}{1.5cm}{} & \multirow{6}{1.5cm}{} & \multirow{6}{1.5cm}{} & \multirow{6}{1.5cm}{$\rho P^{1}$, $\tau P^{1}$, $fJ^{1}$, $mJ^{-7}$, $cA^{1}$, $r A^{1}$, $\rho A^{1}$, $\tau A^{1}$} & \multirow{6}{1.5cm}{} & \multirow{6}{1.5cm}{} & \multirow{6}{1.5cm}{} & \multirow{6}{1.5cm}{} & \multirow{6}{1.5cm}{} & \multirow{6}{1.5cm}{$cP^{1}$, $r P^{1}$, $\rho P^{1}$, $\tau P^{1}$, $fJ^{1}$, $mJ^{-9}$, $cA^{1}$, $r A^{1}$, $\rho A^{1}$, $\tau A^{1}$} \\ \\ \\ \\ \\ \\
\midrule
\multirow{6}{1.5cm}{BioNLP PP} & \multirow{6}{1.5cm}{} & \multirow{6}{1.5cm}{} & \multirow{6}{1.5cm}{} & \multirow{6}{1.5cm}{} & \multirow{6}{1.5cm}{$mJ^{-3}$, $r A^{1}$, $\rho A^{1}$, $\tau A^{1}$} & \multirow{6}{1.5cm}{$r P^{-1}$, $cA^{1}$} & \multirow{6}{1.5cm}{} & \multirow{6}{1.5cm}{} & \multirow{6}{1.5cm}{} & \multirow{6}{1.5cm}{$cP^{1}$, $r P^{1}$, $\rho P^{1}$, $\tau P^{1}$, $fJ^{1}$, $mJ^{-9}$, $cA^{1}$, $r A^{1}$, $\rho A^{1}$, $\tau A^{1}$} & \multirow{6}{1.5cm}{} \\ \\ \\ \\ \\ \\
\midrule
\multirow{6}{1.5cm}{BioNLP PPW} & \multirow{6}{1.5cm}{} & \multirow{6}{1.5cm}{} & \multirow{6}{1.5cm}{} & \multirow{6}{1.5cm}{} & \multirow{6}{1.5cm}{$cP^{1}$, $r P^{1}$, $\rho P^{1}$, $\tau P^{1}$, $fJ^{1}$, $mJ^{-9}$, $cA^{1}$, $r A^{1}$, $\rho A^{1}$, $\tau A^{1}$} & \multirow{6}{1.5cm}{} & \multirow{6}{1.5cm}{} & \multirow{6}{1.5cm}{} & \multirow{6}{1.5cm}{} & \multirow{6}{1.5cm}{} & \multirow{6}{1.5cm}{$r P^{1}$, $mJ^{-1}$} \\ \\ \\ \\ \\ \\
\midrule
\multirow{6}{1.5cm}{BioASQ} & \multirow{6}{1.5cm}{} & \multirow{6}{1.5cm}{} & \multirow{6}{1.5cm}{} & \multirow{6}{1.5cm}{} & \multirow{6}{1.5cm}{$cP^{1}$, $r P^{1}$, $\rho P^{1}$, $\tau P^{1}$, $fJ^{1}$, $mJ^{-9}$, $cA^{1}$, $r A^{1}$, $\rho A^{1}$, $\tau A^{1}$} & \multirow{6}{1.5cm}{$cP^{1}$, $fJ^{1}$, $mJ^{-4}$, $cA^{1}$, $\rho A^{1}$} & \multirow{6}{1.5cm}{$cP^{1}$, $r P^{1}$, $\rho P^{1}$, $\tau P^{1}$, $fJ^{1}$, $mJ^{-9}$, $cA^{1}$, $r A^{1}$, $\rho A^{1}$, $\tau A^{1}$} & \multirow{6}{1.5cm}{} & \multirow{6}{1.5cm}{} & \multirow{6}{1.5cm}{$cP^{1}$, $r P^{1}$, $\rho P^{1}$, $\tau P^{1}$, $fJ^{1}$, $mJ^{-9}$, $cA^{1}$, $r A^{1}$, $\rho A^{1}$, $\tau A^{1}$} & \multirow{6}{1.5cm}{} \\ \\ \\ \\ \\ \\
\midrule
\multirow{6}{1.5cm}{LTL win2} & \multirow{6}{1.5cm}{} & \multirow{6}{1.5cm}{} & \multirow{6}{1.5cm}{} & \multirow{6}{1.5cm}{} & \multirow{6}{1.5cm}{$fJ^{1}$, $mJ^{-5}$, $cA^{1}$, $r A^{1}$, $\rho A^{1}$, $\tau A^{1}$} & \multirow{6}{1.5cm}{} & \multirow{6}{1.5cm}{} & \multirow{6}{1.5cm}{} & \multirow{6}{1.5cm}{} & \multirow{6}{1.5cm}{$cP^{1}$, $r P^{1}$, $\rho P^{1}$, $\tau P^{1}$, $fJ^{1}$, $mJ^{-9}$, $cA^{1}$, $r A^{1}$, $\rho A^{1}$, $\tau A^{1}$} & \multirow{6}{1.5cm}{} \\ \\ \\ \\ \\ \\
\midrule
\multirow{6}{1.5cm}{LTL win30} & \multirow{6}{1.5cm}{} & \multirow{6}{1.5cm}{$mJ^{-2}$, $r A^{1}$, $\tau A^{1}$} & \multirow{6}{1.5cm}{} & \multirow{6}{1.5cm}{} & \multirow{6}{1.5cm}{$cP^{1}$, $r P^{1}$, $\rho P^{1}$, $\tau P^{1}$, $fJ^{1}$, $mJ^{-9}$, $cA^{1}$, $r A^{1}$, $\rho A^{1}$, $\tau A^{1}$} & \multirow{6}{1.5cm}{$cP^{1}$, $r P^{1}$, $\rho P^{1}$, $\tau P^{1}$, $mJ^{-8}$, $cA^{1}$, $r A^{1}$, $\rho A^{1}$, $\tau A^{1}$} & \multirow{6}{1.5cm}{$cP^{1}$, $r P^{1}$, $\rho P^{1}$, $\tau P^{1}$, $fJ^{1}$, $mJ^{-9}$, $cA^{1}$, $r A^{1}$, $\rho A^{1}$, $\tau A^{1}$} & \multirow{6}{1.5cm}{} & \multirow{6}{1.5cm}{$r P^{1}$, $\rho P^{1}$, $\tau P^{1}$, $mJ^{-7}$, $cA^{1}$, $r A^{1}$, $\rho A^{1}$, $\tau A^{1}$} & \multirow{6}{1.5cm}{$cP^{1}$, $r P^{1}$, $\rho P^{1}$, $\tau P^{1}$, $fJ^{1}$, $mJ^{-9}$, $cA^{1}$, $r A^{1}$, $\rho A^{1}$, $\tau A^{1}$} & \multirow{6}{1.5cm}{} \\ \\ \\ \\ \\ \\
\bottomrule
    \end{tabular}
\caption{Similarity metrics for each embedding and existing dataset that have significantly better/worse Spearman's correlation than the number of methods given by the positive/negative superscripts. Significance measured by bias-corrected and accelerated (BCa)
bootstrap confidence intervals with $\alpha = 0.001$/$0.008$ (word/sentence embeddings), i.e. $\alpha = 0.05$ with Bonferroni correction. }
    \label{tab:significance_existingDatasets_methods1} 
\end{sidewaystable*}

\begin{sidewaystable*}
    \centering
    \small
    \begin{tabular}{l l l l l l l l l l l l}
    \toprule
         Dataset & Hlia. & MM-av & MM-c & MM-p & Mayo & Sim & Sim-m & Rel & Rel-m & SimLex & SimVerb\\ 
Subset Size & 36/36 & 29/29 & 29/29 & 29/29 & 81/101 & 352/566 & 340/449 & 347/587 & 339/458 & 964/988 & 909/1000\\
\midrule
\multirow{5}{1.5cm}{AUEB200} & \multirow{5}{1.5cm}{} & \multirow{5}{1.5cm}{} & \multirow{5}{1.5cm}{} & \multirow{5}{1.5cm}{} & \multirow{5}{1.5cm}{$cP^{1}$, $\rho P^{1}$, $\tau P^{1}$, $mJ^{-7}$, $cA^{1}$, $r A^{1}$, $\rho A^{1}$, $\tau A^{1}$} & \multirow{5}{1.5cm}{} & \multirow{5}{1.5cm}{} & \multirow{5}{1.5cm}{} & \multirow{5}{1.5cm}{} & \multirow{5}{1.5cm}{} & \multirow{5}{1.5cm}{} \\ \\ \\ \\ \\
\midrule
\multirow{6}{1.5cm}{AUEB400} & \multirow{6}{1.5cm}{} & \multirow{6}{1.5cm}{} & \multirow{6}{1.5cm}{} & \multirow{6}{1.5cm}{} & \multirow{6}{1.5cm}{$cP^{1}$, $r P^{1}$, $\rho P^{1}$, $\tau P^{1}$, $fJ^{1}$, $mJ^{-9}$, $cA^{1}$, $r A^{1}$, $\rho A^{1}$, $\tau A^{1}$} & \multirow{6}{1.5cm}{} & \multirow{6}{1.5cm}{$cP^{1}$, $r P^{1}$, $\rho P^{1}$, $\tau P^{1}$, $fJ^{1}$, $mJ^{-9}$, $cA^{1}$, $r A^{1}$, $\rho A^{1}$, $\tau A^{1}$} & \multirow{6}{1.5cm}{} & \multirow{6}{1.5cm}{$cP^{1}$, $r P^{1}$, $\tau P^{1}$, $fJ^{1}$, $mJ^{-6}$, $cA^{1}$, $\rho A^{1}$} & \multirow{6}{1.5cm}{} & \multirow{6}{1.5cm}{} \\ \\ \\ \\ \\ \\
\midrule
\multirow{6}{1.5cm}{MeSH extr} & \multirow{6}{1.5cm}{} & \multirow{6}{1.5cm}{} & \multirow{6}{1.5cm}{} & \multirow{6}{1.5cm}{} & \multirow{6}{1.5cm}{$cP^{1}$, $r P^{1}$, $\rho P^{1}$, $\tau P^{1}$, $fJ^{1}$, $mJ^{-9}$, $cA^{1}$, $r A^{1}$, $\rho A^{1}$, $\tau A^{1}$} & \multirow{6}{1.5cm}{} & \multirow{6}{1.5cm}{} & \multirow{6}{1.5cm}{$fJ^{-1}$, $\rho A^{1}$} & \multirow{6}{1.5cm}{} & \multirow{6}{1.5cm}{} & \multirow{6}{1.5cm}{} \\ \\ \\ \\ \\ \\
\midrule
\multirow{6}{1.5cm}{MeSH intr} & \multirow{6}{1.5cm}{} & \multirow{6}{1.5cm}{} & \multirow{6}{1.5cm}{} & \multirow{6}{1.5cm}{} & \multirow{6}{1.5cm}{$cP^{1}$, $r P^{1}$, $\rho P^{1}$, $\tau P^{1}$, $fJ^{1}$, $mJ^{-9}$, $cA^{1}$, $r A^{1}$, $\rho A^{1}$, $\tau A^{1}$} & \multirow{6}{1.5cm}{} & \multirow{6}{1.5cm}{} & \multirow{6}{1.5cm}{} & \multirow{6}{1.5cm}{} & \multirow{6}{1.5cm}{} & \multirow{6}{1.5cm}{$\rho P^{1}$, $\tau P^{1}$, $mJ^{-4}$, $r A^{1}$, $\tau A^{1}$} \\ \\ \\ \\ \\ \\
\midrule
\multirow{8}{1.5cm}{MIMIC} & \multirow{8}{1.5cm}{} & \multirow{8}{1.5cm}{$\rho P^{1}$, $\tau P^{1}$, $fJ^{1}$, $mJ^{-6}$, $r A^{1}$, $\rho A^{1}$, $\tau A^{1}$} & \multirow{8}{1.5cm}{} & \multirow{8}{1.5cm}{$\tau P^{1}$, $mJ^{-5}$, $cA^{1}$, $r A^{1}$, $\rho A^{1}$, $\tau A^{1}$} & \multirow{8}{1.5cm}{$cP^{1}$, $r P^{1}$, $\rho P^{1}$, $\tau P^{1}$, $fJ^{1}$, $mJ^{-9}$, $cA^{1}$, $r A^{1}$, $\rho A^{1}$, $\tau A^{1}$} & \multirow{8}{1.5cm}{$cP^{1}$, $r P^{2}$, $\rho P^{-2}$, $fJ^{1}$, $mJ^{-5}$, $cA^{3}$, $r A^{-1}$, $\rho A^{1}$} & \multirow{8}{1.5cm}{$cP^{1}$, $r P^{1}$, $fJ^{1}$, $mJ^{-5}$, $cA^{1}$, $\rho A^{1}$} & \multirow{8}{1.5cm}{$cP^{1}$, $r P^{1}$, $\rho P^{-4}$, $\tau P^{-4}$, $fJ^{3}$, $mJ^{-3}$, $cA^{5}$, $r A^{-2}$, $\rho A^{4}$, $\tau A^{-1}$} & \multirow{8}{1.5cm}{$cP^{1}$, $r P^{1}$, $fJ^{1}$, $mJ^{-5}$, $cA^{1}$, $\rho A^{1}$} & \multirow{8}{1.5cm}{} & \multirow{8}{1.5cm}{} \\ \\ \\ \\ \\ \\ \\ \\
\midrule
\multirow{8}{1.5cm}{MIMIC M} & \multirow{8}{1.5cm}{} & \multirow{8}{1.5cm}{$\tau P^{1}$, $fJ^{1}$, $mJ^{-5}$, $cA^{1}$, $r A^{1}$, $\tau A^{1}$} & \multirow{8}{1.5cm}{} & \multirow{8}{1.5cm}{$\tau P^{1}$, $mJ^{-5}$, $cA^{1}$, $r A^{1}$, $\rho A^{1}$, $\tau A^{1}$} & \multirow{8}{1.5cm}{$cP^{1}$, $r P^{1}$, $\rho P^{1}$, $\tau P^{1}$, $fJ^{1}$, $mJ^{-9}$, $cA^{1}$, $r A^{1}$, $\rho A^{1}$, $\tau A^{1}$} & \multirow{8}{1.5cm}{$cP^{2}$, $r P^{1}$, $\rho P^{-2}$, $fJ^{1}$, $mJ^{-5}$, $cA^{2}$, $\rho A^{1}$} & \multirow{8}{1.5cm}{$cP^{1}$, $r P^{1}$, $fJ^{1}$, $mJ^{-5}$, $cA^{1}$, $\rho A^{1}$} & \multirow{8}{1.5cm}{$cP^{2}$, $r P^{1}$, $\rho P^{-5}$, $\tau P^{-4}$, $fJ^{3}$, $mJ^{-3}$, $cA^{5}$, $r A^{-2}$, $\rho A^{5}$, $\tau A^{-2}$} & \multirow{8}{1.5cm}{$cP^{3}$, $r P^{2}$, $\rho P^{-2}$, $\tau P^{-1}$, $fJ^{1}$, $mJ^{-5}$, $cA^{5}$, $r A^{-3}$, $\rho A^{1}$, $\tau A^{-1}$} & \multirow{8}{1.5cm}{} & \multirow{8}{1.5cm}{} \\ \\ \\ \\ \\ \\ \\ \\
\bottomrule
    \end{tabular}
\caption{Similarity metrics for each embedding and existing dataset that have significantly better/worse Spearman's correlation than the number of methods given by the positive/negative superscripts. Significance measured by bias-corrected and accelerated (BCa)
bootstrap confidence intervals with $\alpha = 0.001$/$0.008$ (word/sentence embeddings), i.e. $\alpha = 0.05$ with Bonferroni correction.}
    \label{tab:significance_existingDatasets_methods2} 
\end{sidewaystable*}

\begin{sidewaystable*}
    \centering
    \small
    \begin{tabular}{l l l l l l l l l l l l}
    \toprule
         Dataset & Hlia. & MM-av & MM-c & MM-p & Mayo & Sim & Sim-m & Rel & Rel-m & SimLex & SimVerb\\ 
Subset Size & 36/36 & 29/29 & 29/29 & 29/29 & 81/101 & 352/566 & 340/449 & 347/587 & 339/458 & 964/988 & 909/1000\\
\midrule
\multirow{8}{1.5cm}{GloVe} & \multirow{8}{1.5cm}{} & \multirow{8}{1.5cm}{} & \multirow{8}{1.5cm}{} & \multirow{8}{1.5cm}{} & \multirow{8}{1.5cm}{} & \multirow{8}{1.5cm}{$cP^{-4}$, $r P^{-4}$, $\rho P^{3}$, $\tau P^{2}$, $fJ^{-2}$, $cA^{-2}$, $r A^{5}$, $\rho A^{-3}$, $\tau A^{5}$} & \multirow{8}{1.5cm}{$cP^{-4}$, $r P^{-4}$, $\rho P^{5}$, $\tau P^{5}$, $fJ^{-4}$, $cA^{-4}$, $r A^{5}$, $\rho A^{-4}$, $\tau A^{5}$} & \multirow{8}{1.5cm}{$cP^{-4}$, $r P^{-4}$, $\rho P^{2}$, $\tau P^{3}$, $fJ^{-3}$, $cA^{-2}$, $r A^{5}$, $\rho A^{-2}$, $\tau A^{5}$} & \multirow{8}{1.5cm}{$cP^{-4}$, $r P^{-4}$, $\rho P^{5}$, $\tau P^{5}$, $fJ^{-4}$, $cA^{-4}$, $r A^{5}$, $\rho A^{-4}$, $\tau A^{5}$} & \multirow{8}{1.5cm}{$cP^{-3}$, $r P^{-4}$, $\rho P^{5}$, $\tau P^{6}$, $fJ^{-3}$, $mJ^{-4}$, $cA^{-4}$, $r A^{5}$, $\rho A^{-4}$, $\tau A^{6}$} & \multirow{8}{1.5cm}{$cP^{-3}$, $r P^{-4}$, $\rho P^{3}$, $\tau P^{2}$, $fJ^{-1}$, $cA^{-2}$, $r A^{3}$, $\rho A^{-2}$, $\tau A^{4}$} \\ \\ \\ \\ \\ \\ \\ \\
\midrule
\multirow{9}{1.5cm}{Fastt Wiki} & \multirow{9}{1.5cm}{} & \multirow{9}{1.5cm}{$cP^{2}$, $r P^{2}$, $fJ^{-2}$, $mJ^{-3}$, $cA^{1}$} & \multirow{9}{1.5cm}{$mJ^{-2}$, $\rho A^{1}$, $\tau A^{1}$} & \multirow{9}{1.5cm}{} & \multirow{9}{1.5cm}{$r P^{1}$, $\rho P^{1}$, $\tau P^{1}$, $fJ^{1}$, $mJ^{-8}$, $cA^{1}$, $r A^{1}$, $\rho A^{1}$, $\tau A^{1}$} & \multirow{9}{1.5cm}{$cP^{2}$, $r P^{2}$, $\rho P^{2}$, $\tau P^{2}$, $fJ^{1/-8}$, $mJ^{-9}$, $cA^{2}$, $r A^{2}$, $\rho A^{2}$, $\tau A^{2}$} & \multirow{9}{1.5cm}{$cP^{2}$, $r P^{2}$, $\rho P^{2}$, $\tau P^{2}$, $fJ^{1/-8}$, $mJ^{-9}$, $cA^{2}$, $r A^{2}$, $\rho A^{2}$, $\tau A^{2}$} & \multirow{9}{1.5cm}{$cP^{2}$, $r P^{2}$, $\rho P^{2}$, $\tau P^{2}$, $fJ^{1/-6}$, $mJ^{-9}$, $cA^{2}$, $r A^{1}$, $\rho A^{2}$, $\tau A^{1}$} & \multirow{9}{1.5cm}{$cP^{2}$, $r P^{2}$, $\rho P^{1}$, $\tau P^{1}$, $fJ^{1/-4}$, $mJ^{-9}$, $cA^{2}$, $r A^{1}$, $\rho A^{2}$, $\tau A^{1}$} & \multirow{9}{1.5cm}{$cP^{2}$, $r P^{2}$, $\rho P^{2}$, $\tau P^{2}$, $fJ^{1/-8}$, $mJ^{-9}$, $cA^{2}$, $r A^{2}$, $\rho A^{2}$, $\tau A^{2}$} & \multirow{9}{1.5cm}{$cP^{2/-4}$, $r P^{2/-4}$, $\rho P^{6}$, $\tau P^{6}$, $fJ^{-8}$, $mJ^{-8}$, $cA^{2/-4}$, $r A^{6}$, $\rho A^{2/-4}$, $\tau A^{6}$} \\ \\ \\ \\ \\ \\ \\ \\ \\
\midrule
\multirow{6}{1.5cm}{Fastt Crawl} & \multirow{6}{1.5cm}{} & \multirow{6}{1.5cm}{} & \multirow{6}{1.5cm}{} & \multirow{6}{1.5cm}{} & \multirow{6}{1.5cm}{$cP^{1}$, $r P^{1}$, $\rho P^{1}$, $\tau P^{1}$, $fJ^{1}$, $mJ^{-9}$, $cA^{1}$, $r A^{1}$, $\rho A^{1}$, $\tau A^{1}$} & \multirow{6}{1.5cm}{$cP^{1}$, $r P^{1}$, $\rho P^{1}$, $\tau P^{1}$, $fJ^{1}$, $mJ^{-9}$, $cA^{1}$, $r A^{1}$, $\rho A^{1}$, $\tau A^{1}$} & \multirow{6}{1.5cm}{$cP^{1}$, $r P^{1}$, $\rho P^{1}$, $fJ^{1}$, $mJ^{-8}$, $cA^{1}$, $r A^{1}$, $\rho A^{1}$, $\tau A^{1}$} & \multirow{6}{1.5cm}{$cP^{1}$, $r P^{1}$, $mJ^{-4}$, $cA^{1}$, $\rho A^{1}$} & \multirow{6}{1.5cm}{$cP^{1}$, $r P^{1}$, $mJ^{-4}$, $cA^{1}$, $\rho A^{1}$} & \multirow{6}{1.5cm}{$cP^{2}$, $r P^{2}$, $\rho P^{2}$, $\tau P^{2}$, $fJ^{-8}$, $mJ^{-8}$, $cA^{2}$, $r A^{2}$, $\rho A^{2}$, $\tau A^{2}$} & \multirow{6}{1.5cm}{$\rho P^{2}$, $\tau P^{2}$, $fJ^{-4}$, $mJ^{-4}$, $r A^{2}$, $\tau A^{2}$} \\ \\ \\ \\ \\ \\
\midrule
\multirow{6}{1.5cm}{Fastt Crawl M} & \multirow{6}{1.5cm}{} & \multirow{6}{1.5cm}{$cP^{-1}$, $r P^{-1}$, $\tau P^{3}$, $mJ^{-1}$} & \multirow{6}{1.5cm}{} & \multirow{6}{1.5cm}{$r P^{-1}$, $\tau P^{1}$} & \multirow{6}{1.5cm}{$cP^{1}$, $r P^{1}$, $\tau P^{1}$, $fJ^{1}$, $mJ^{-8}$, $cA^{1}$, $r A^{1}$, $\rho A^{1}$, $\tau A^{1}$} & \multirow{6}{1.5cm}{$cP^{1}$, $r P^{1}$, $\rho P^{1}$, $\tau P^{1}$, $fJ^{1}$, $mJ^{-9}$, $cA^{1}$, $r A^{1}$, $\rho A^{1}$, $\tau A^{1}$} & \multirow{6}{1.5cm}{$cP^{1}$, $r P^{1}$, $\rho P^{1}$, $\tau P^{1}$, $fJ^{1}$, $mJ^{-8}$, $cA^{1}$, $r A^{1}$, $\rho A^{1}$} & \multirow{6}{1.5cm}{$cP^{2}$, $r P^{2}$, $\rho P^{1}$, $fJ^{-2}$, $mJ^{-7}$, $cA^{1}$, $r A^{1}$, $\rho A^{1}$, $\tau A^{1}$} & \multirow{6}{1.5cm}{$cP^{1}$, $r P^{1}$, $mJ^{-4}$, $cA^{1}$, $\rho A^{1}$} & \multirow{6}{1.5cm}{$cP^{2}$, $r P^{2}$, $\rho P^{2}$, $\tau P^{2}$, $fJ^{-8}$, $mJ^{-8}$, $cA^{2}$, $r A^{2}$, $\rho A^{2}$, $\tau A^{2}$} & \multirow{6}{1.5cm}{$cP^{2}$, $r P^{2}$, $\rho P^{2}$, $\tau P^{2}$, $fJ^{-8}$, $mJ^{-8}$, $cA^{2}$, $r A^{2}$, $\rho A^{2}$, $\tau A^{2}$} \\ \\ \\ \\ \\ \\
\bottomrule
    \end{tabular}
    \caption{Similarity metrics for each embedding and existing dataset that have significantly better/worse Spearman's correlation than the number of methods given by the positive/negative superscripts. Significance measured by bias-corrected and accelerated (BCa)
bootstrap confidence intervals with $\alpha = 0.001$/$0.008$ (word/sentence embeddings), i.e. $\alpha = 0.05$ with Bonferroni correction.}
    \label{tab:significance_existingDatasets_methods3} 
\end{sidewaystable*}

\begin{sidewaystable*}
    \centering
    \small
    \begin{tabular}{l l l l l l l l l l l l}
    \toprule
         Dataset & Hlia. & MM-av & MM-c & MM-p & Mayo & Sim & Sim-m & Rel & Rel-m & SimLex & SimVerb\\ 
Subset Size & 36/36 & 29/29 & 29/29 & 29/29 & 81/101 & 352/566 & 340/449 & 347/587 & 339/458 & 964/988 & 909/1000\\
\midrule
\multirow{4}{1.5cm}{ELMoPubMed} & \multirow{4}{1.5cm}{} & \multirow{4}{1.5cm}{} & \multirow{4}{1.5cm}{} & \multirow{4}{1.5cm}{} & \multirow{4}{1.5cm}{} & \multirow{4}{1.5cm}{$c^{-1}$, $r^{-1}$, $\rho^{2}$} & \multirow{4}{1.5cm}{$r^{-1}$, $\rho^{1}$} & \multirow{4}{1.5cm}{} & \multirow{4}{1.5cm}{} & \multirow{4}{1.5cm}{} & \multirow{4}{1.5cm}{$c^{-2}$, $r^{-2}$, $\rho^{2}$, $\tau^{2}$} \\ \\ \\ \\
\midrule
\multirow{2}{1.5cm}{Flair} & \multirow{2}{1.5cm}{} & \multirow{2}{1.5cm}{} & \multirow{2}{1.5cm}{} & \multirow{2}{1.5cm}{} & \multirow{2}{1.5cm}{} & \multirow{2}{1.5cm}{} & \multirow{2}{1.5cm}{} & \multirow{2}{1.5cm}{} & \multirow{2}{1.5cm}{} & \multirow{2}{1.5cm}{} & \multirow{2}{1.5cm}{$\rho^{1}$, $\tau^{-1}$} \\ \\
\midrule
\multirow{4}{1.5cm}{SciBERT} & \multirow{4}{1.5cm}{} & \multirow{4}{1.5cm}{} & \multirow{4}{1.5cm}{} & \multirow{4}{1.5cm}{} & \multirow{4}{1.5cm}{} & \multirow{4}{1.5cm}{$c^{-2}$, $r^{-2}$, $\rho^{2}$, $\tau^{2}$} & \multirow{4}{1.5cm}{$c^{-2}$, $r^{-2}$, $\rho^{2}$, $\tau^{2}$} & \multirow{4}{1.5cm}{} & \multirow{4}{1.5cm}{$c^{-1}$, $\tau^{1}$} & \multirow{4}{1.5cm}{$c^{-2}$, $r^{-2}$, $\rho^{2}$, $\tau^{2}$} & \multirow{4}{1.5cm}{$c^{-2}$, $r^{-2}$, $\rho^{2}$, $\tau^{2}$} \\ \\ \\ \\
\midrule
\multirow{4}{1.5cm}{BERT} & \multirow{4}{1.5cm}{} & \multirow{4}{1.5cm}{} & \multirow{4}{1.5cm}{} & \multirow{4}{1.5cm}{} & \multirow{4}{1.5cm}{} & \multirow{4}{1.5cm}{} & \multirow{4}{1.5cm}{} & \multirow{4}{1.5cm}{} & \multirow{4}{1.5cm}{} & \multirow{4}{1.5cm}{$\rho^{-1}$, $\tau^{1}$} & \multirow{4}{1.5cm}{$c^{-2}$, $r^{-2}$, $\rho^{2}$, $\tau^{2}$} \\ \\ \\ \\
\midrule
\multirow{3}{1.5cm}{ELMo} & \multirow{3}{1.5cm}{} & \multirow{3}{1.5cm}{} & \multirow{3}{1.5cm}{$c^{1}$, $r^{1}$, $\rho^{-2}$} & \multirow{3}{1.5cm}{} & \multirow{3}{1.5cm}{} & \multirow{3}{1.5cm}{} & \multirow{3}{1.5cm}{} & \multirow{3}{1.5cm}{} & \multirow{3}{1.5cm}{} & \multirow{3}{1.5cm}{} & \multirow{3}{1.5cm}{} \\ \\ \\
\midrule
\multirow{2}{1.5cm}{GPT} & \multirow{2}{1.5cm}{} & \multirow{2}{1.5cm}{} & \multirow{2}{1.5cm}{} & \multirow{2}{1.5cm}{} & \multirow{2}{1.5cm}{} & \multirow{2}{1.5cm}{} & \multirow{2}{1.5cm}{} & \multirow{2}{1.5cm}{} & \multirow{2}{1.5cm}{} & \multirow{2}{1.5cm}{$\rho^{-1}$, $\tau^{1}$} & \multirow{2}{1.5cm}{} \\ \\
\bottomrule
    \end{tabular}
    \caption{Similarity metrics for each embedding and existing dataset that have significantly better/worse Spearman's correlation than the number of methods given by the positive/negative superscripts. Significance measured by bias-corrected and accelerated (BCa)
bootstrap confidence intervals with $\alpha = 0.001$/$0.008$ (word/sentence embeddings), i.e. $\alpha = 0.05$ with Bonferroni correction.}
    \label{tab:significance_existingDatasets_methods4} 
\end{sidewaystable*}

\clearpage
\section{Evaluation on New Large-Scale Datasets}
Table~\ref{tab:significance_newDatasets_subset_negR} reports accuracy for each embedding (with thresholds optimising accuracy for each embedding and dataset) on the new datasets created with random negative sampling. It can be observed that all accuracy scores are very high, illustrating that similar concept pairs in these datasets are easy to sepearte from dissimilar ones.
Tables~\ref{tab:newDatasets_auc_negL} and~\ref{tab:newDatasets_auc_negR} give the area under the ROC curve (AUC) for datasets with, respectively, Levenshtein and random negative sampling. The low AUC for the hard datasets in the first table indicate the high difficulty of these datasets, for which none of the embeddings is able to distinguish similar and dissimilar concept pairs.

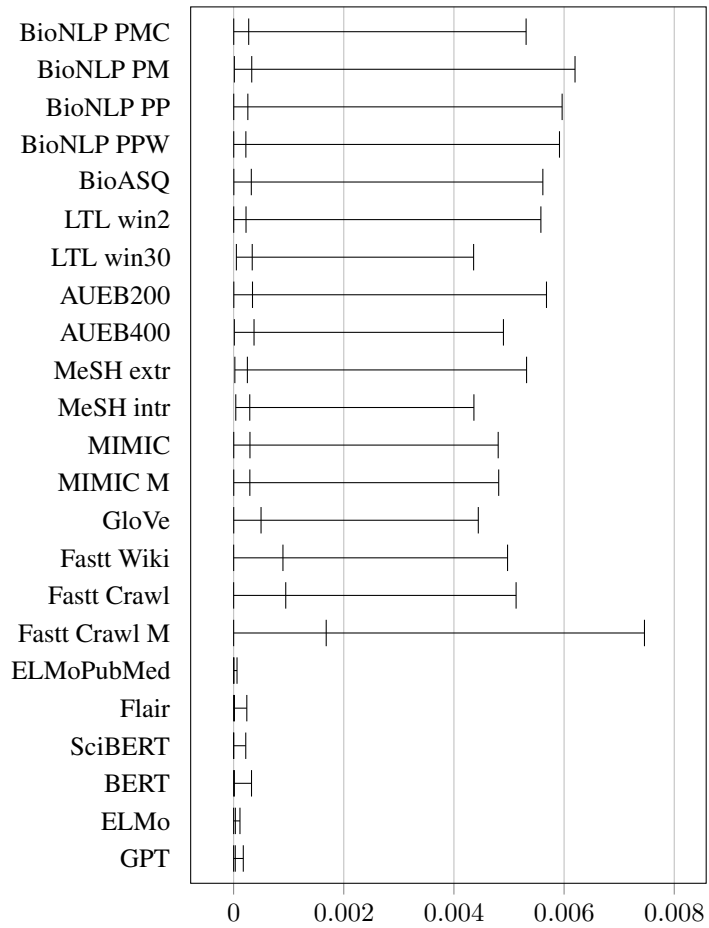
\begin{figure}[h]
    \centering
\begin{tikzpicture}
\begin{axis}[symbolic y coords={
                GPT,
                ELMo,
                BERT,
                SciBERT,
                Flair,
                ELMoPubMed,
                Fastt Crawl M,
                Fastt Crawl,
                Fastt Wiki,
                GloVe,
                MIMIC M, 
                MIMIC,
                MeSH intr,
                MeSH extr,
                AUEB400,
                AUEB200,
                LTL win30,
                LTL win2,
                BioASQ,
                BioNLP PPW,
                BioNLP PP,
                BioNLP PM,
                BioNLP PMC
                },
                ytick={
                GPT,
                ELMo,
                BERT,
                SciBERT,
                Flair,
                ELMoPubMed,
                Fastt Crawl M,
                Fastt Crawl,
                Fastt Wiki,
                GloVe,
                MIMIC M, 
                MIMIC,
                MeSH intr,
                MeSH extr,
                AUEB400,
                AUEB200,
                LTL win30,
                LTL win2,
                BioASQ,
                BioNLP PPW,
                BioNLP PP,
                BioNLP PM,
                BioNLP PMC
                },
                y=0.5cm, enlarge y limits ={true, value=0.03}, enlarge x limits=true, ytick=data, xtick align=outside, ytick align=outside, tick pos=left, ytick style = {draw=none}, xmajorgrids = true,
  x tick label style=
{/pgf/number format/fixed, 
/pgf/number format/precision=3,
scaled x ticks = false}
  ]

\addplot[white,only marks] coordinates{(0.0078,SciBERT)(0.0078,BioASQ)(0.0078,Fastt Crawl M)(0.0078,MeSH extr)(0.0078,GloVe)(0.0078,Fastt Wiki)(0.0078,Fastt Crawl)(0.0078,MeSH intr)(0.0078,MIMIC)(0.0078,MIMIC M)(0.0078,BioNLP PMC)(0.0078,BioNLP PM)(0.0078,BioNLP PP)(0.0078,BioNLP PPW)(0.0078,LTL win2)(0.0078,LTL win30)(0.0078,AUEB200)(0.0078,AUEB400)(0.0078,BERT)(0.0078,ELMo)(0.0078,ELMoPubMed)(0.0078,Flair)(0.0078,GPT)};
\addplot[mark=|, mark size = 5pt] coordinates {(1.715082095834194e-06,BioNLP PMC) (0.00027359484320453874,BioNLP PMC) (0.005310502592955429,BioNLP PMC)};
\addplot[mark=|, mark size = 5pt] coordinates {(1.3646372705159932e-05,BioNLP PM) (0.0003287398828538874,BioNLP PM) (0.00619869633982916,BioNLP PM)};
\addplot[mark=|, mark size = 5pt] coordinates {(1.7740582147490053e-06,BioNLP PP) (0.0002580782955524797,BioNLP PP) (0.005964507316401937,BioNLP PP)};
\addplot[mark=|, mark size = 5pt] coordinates {(1.756533950076043e-06,BioNLP PPW) (0.00022218250798960166,BioNLP PPW) (0.005916583136432756,BioNLP PPW)};
\addplot[mark=|, mark size = 5pt] coordinates {(3.516983711887264e-06,BioASQ) (0.0003210770183556626,BioASQ) (0.005614420181633465,BioASQ)};
\addplot[mark=|, mark size = 5pt] coordinates {(1.5965618303791064e-06,LTL win2) (0.0002249727694348401,LTL win2) (0.005578733793000121,LTL win2)};
\addplot[mark=|, mark size = 5pt] coordinates {(4.998384578534882e-05,LTL win30) (0.0003378090152433336,LTL win30) (0.0043573875196063734,LTL win30)};
\addplot[mark=|, mark size = 5pt] coordinates {(3.1764319163583304e-06,AUEB200) (0.0003427015594545678,AUEB200) (0.005680509277771287,AUEB200)};
\addplot[mark=|, mark size = 5pt] coordinates {(1.3146081225766597e-05,AUEB400) (0.00037059830206440927,AUEB400) (0.00489897450188237,AUEB400)};
\addplot[mark=|, mark size = 5pt] coordinates {(2.209662768476252e-05,MeSH extr) (0.00024962143150220967,MeSH extr) (0.005318941530484101,MeSH extr)};
\addplot[mark=|, mark size = 5pt] coordinates {(3.899485425638729e-05,MeSH intr) (0.000292899558618182,MeSH intr) (0.0043628906670613084,MeSH intr)};
\addplot[mark=|, mark size = 5pt] coordinates {(1.8088426342151093e-06,MIMIC) (0.00029655622894243436,MIMIC) (0.004803671522968709,MIMIC)};
\addplot[mark=|, mark size = 5pt] coordinates {(1.8088426342151093e-06,MIMIC M) (0.0002944722892833993,MIMIC M) (0.004812833145838603,MIMIC M)};
\addplot[mark=|, mark size = 5pt] coordinates {(1.6123062241233245e-06,GloVe) (0.0004977490674139353,GloVe) (0.004442422995389776,GloVe)};
\addplot[mark=|, mark size = 5pt] coordinates {(6.049402271157022e-07,Fastt Wiki) (0.0008955058183144657,Fastt Wiki) (0.004973718750953878,Fastt Wiki)};
\addplot[mark=|, mark size = 5pt] coordinates {(1.0362434609352464e-07,Fastt Crawl) (0.000946232412055377,Fastt Crawl) (0.005128683665198116,Fastt Crawl)};
\addplot[mark=|, mark size = 5pt] coordinates {(1.173655768326758e-07,Fastt Crawl M) (0.0016808951270624335,Fastt Crawl M) (0.0074596115299695774,Fastt Crawl M)};
\addplot[mark=|, mark size = 5pt] coordinates {(0.0,ELMoPubMed) (4.950969762628764e-06,ELMoPubMed) (6.165796396683682e-05,ELMoPubMed)};
\addplot[mark=|, mark size = 5pt] coordinates {(0.0,Flair) (1.4846458230904776e-05,Flair) (0.00023944361680219018,Flair)};
\addplot[mark=|, mark size = 5pt] coordinates {(0.0,SciBERT) (8.809309961394332e-07,SciBERT) (0.00022025680941770205,SciBERT)};
\addplot[mark=|, mark size = 5pt] coordinates {(0.0,BERT) (1.359138124417108e-05,BERT) (0.00032482728447442576,BERT)};
\addplot[mark=|, mark size = 5pt] coordinates {(0.0,ELMo) (3.1093792213775756e-05,ELMo) (0.00011502075166061201,ELMo)};
\addplot[mark=|, mark size = 5pt] coordinates {(0.0,GPT) (2.9710575919805347e-05,GPT) (0.00017542395761792456,GPT)};
\end{axis}
\end{tikzpicture}
    \caption{Min/median/max variance between the accuracy of the different similarity metrics applied to an embedding observed for one of our new datasets (both with random and Levenshtein negative sampling).}
    \label{fig:variance_newdatasets}
\end{figure}

Figure~\ref{fig:variance_newdatasets} illustrates the variance of accuracy obtained with the different similarity metrics for each dataset.
Median variances are very low.
We observe that the contextual embeddings are barely affected by the similarity metric.
Tables~\ref{tab:significance_snomedDatasets_methods1}-\ref{tab:significance_snomedDatasets_methods5} detail for each embedding, which similarity metrics perform \emph{significantly} worse or better than others for each dataset with Levenshtein negative sampling.
All analyses are performed on subsets of each dataset that have no OOV concepts for any embedding.

\begin{table}[h]
    \centering
\scriptsize	
     \begin{tabular}{l l l l l l l l l l l}
\multirow{2}{*}{Dataset}& \textbf{\dataset{FSN-syn.}} & \textbf{\dataset{FSN-syn.}} & \textbf{\dataset{syn-syn.}} & \textbf{\dataset{syn-syn.}} & \textbf{\dataset{poss.-equ.}} & \textbf{\dataset{poss.-equ.}} & \textbf{\dataset{repl.-by}} & \textbf{\dataset{repl.-by}} & \textbf{\dataset{same-as}} & \textbf{\dataset{same-as}} \\
& \textbf{easy} & \textbf{hard} & \textbf{easy} & \textbf{hard} & \textbf{easy} & \textbf{hard} & \textbf{easy} & \textbf{hard} & \textbf{easy} & \textbf{hard} \\ 
Subset Size & 58.0\% & 51.0\% & 55.3\% & 48.5\% & 69.5\% & 65.3\% & 40.0\% & 56.3\% & 65.5\% & 66.0\% \\
\midrule
BioNLP PMC & $98.4^{6/-14}$ & $95.1^{9/-13}$ & $98.2^{6/-14}$ & $93.9^{9/-13}$ & $97.5^{4}$ & $91.9^{12/-10}$ & $98.1^{3}$ & $93.6^{8/-6}$ & $97.9^{3/-10}$ & $95.9^{9/-8}$ \\
BioNLP PM & $98.8^{8/-8}$ & $96.1^{13/-8}$ & $98.6^{12/-7}$ & $95.1^{13/-8}$ & $98.2^{4}$ & $93.4^{16/-6}$ & $98.5^{3}$ & $94.8^{9}$ & $99.0^{5}$ & $96.7^{13/-4}$ \\
BioNLP PP & $98.7^{8/-9}$ & $95.8^{11/-11}$ & $98.5^{8/-9}$ & $94.7^{11/-10}$ & $98.1^{4}$ & $92.9^{15/-7}$ & $98.5^{3}$ & $94.4^{9}$ & $98.6^{4/-4}$ & $96.6^{11/-5}$ \\
BioNLP PPW & $98.7^{8/-9}$ & $95.5^{10/-12}$ & $98.5^{8/-10}$ & $94.4^{10/-12}$ & $98.2^{4}$ & $92.4^{13/-8}$ & $98.5^{3}$ & $94.1^{9}$ & $98.8^{4}$ & $96.3^{9/-7}$ \\
BioASQ & $98.9^{12/-7}$ & $96.4^{17/-3}$ & $98.7^{13/-6}$ & $95.4^{17/-3}$ & $98.5^{4}$ & $95.5^{18/-3}$ & $98.5^{3}$ & $94.8^{10}$ & $98.9^{4}$ & $97.0^{15/-2}$ \\
LTL win2 & $98.9^{13/-7}$ & $95.9^{12/-10}$ & $98.7^{13/-6}$ & $94.8^{11/-10}$ & $98.5^{4}$ & $92.4^{13/-8}$ & $98.1^{3}$ & $94.2^{9}$ & $99.1^{5}$ & $96.2^{9/-7}$ \\
LTL win30 & $98.6^{7/-10}$ & $96.4^{17/-3}$ & $98.5^{8/-10}$ & $95.4^{17/-3}$ & $98.2^{4}$ & $94.2^{17/-5}$ & $98.8^{4}$ & $94.9^{10}$ & $98.9^{5}$ & $97.2^{17/-1}$ \\
AUEB200 & $99.0^{16/-2}$ & $96.5^{17/-3}$ & $98.8^{16/-2}$ & $95.4^{17/-3}$ & $98.9^{5}$ & $95.6^{18/-3}$ & $98.5^{3}$ & $94.9^{10}$ & $99.3^{6}$ & $97.1^{16/-2}$ \\
AUEB400 & $99.1^{16/-2}$ & $\mathbf{96.7^{20}}$ & $98.9^{17/-2}$ & $95.6^{20/-1}$ & $98.9^{5}$ & $95.8^{20/-1}$ & $98.8^{4}$ & $95.0^{10}$ & $99.4^{6}$ & $97.4^{17}$ \\
MeSH extr & $99.2^{19}$ & $\mathbf{96.7^{20}}$ & $99.0^{19/-1}$ & $\mathbf{95.7^{21}}$ & $98.9^{5}$ & $96.0^{20}$ & $98.8^{4}$ & $95.1^{10}$ & $99.6^{7}$ & $97.5^{19}$ \\
MeSH intr & $99.0^{16/-2}$ & $\mathbf{96.7^{20}}$ & $98.8^{14/-5}$ & $95.7^{20}$ & $98.3^{4}$ & $\mathbf{96.2^{21}}$ & $98.8^{4}$ & $94.9^{10}$ & $99.5^{6}$ & $\mathbf{97.7^{20}}$ \\
MIMIC & $99.1^{16/-1}$ & $96.3^{14/-6}$ & $98.9^{17/-1}$ & $95.2^{14/-6}$ & $98.1^{4}$ & $91.2^{10/-11}$ & $98.8^{4}$ & $94.1^{9}$ & $99.6^{7}$ & $96.2^{9/-6}$ \\
MIMIC M & $99.1^{16/-1}$ & $96.3^{14/-6}$ & $98.9^{17/-1}$ & $95.2^{14/-6}$ & $98.1^{4}$ & $91.1^{10/-11}$ & $98.8^{4}$ & $94.1^{9}$ & $99.6^{7}$ & $96.2^{9/-6}$ \\
\midrule
GloVe & $98.8^{9/-7}$ & $94.7^{8/-14}$ & $98.4^{8/-10}$ & $93.0^{8/-14}$ & $98.6^{5}$ & $89.2^{8/-14}$ & $96.9^{3}$ & $92.4^{7/-13}$ & $99.0^{4}$ & $95.2^{8/-13}$ \\
Fastt Wiki & $86.4^{-22}$ & $74.9^{-22}$ & $86.6^{-22}$ & $72.7^{-22}$ & $79.4^{-21}$ & $58.1^{-22}$ & $76.2^{-20}$ & $64.1^{-22}$ & $79.1^{-21}$ & $66.9^{-22}$ \\
Fastt Crawl & $89.3^{2/-20}$ & $82.3^{2/-20}$ & $89.5^{2/-20}$ & $80.0^{2/-20}$ & $83.9^{2/-20}$ & $62.3^{2/-20}$ & $84.2^{-19}$ & $70.3^{2/-20}$ & $86.3^{2/-20}$ & $73.4^{2/-20}$ \\
Fastt Crawl M & $87.0^{1/-21}$ & $77.6^{1/-21}$ & $87.3^{1/-21}$ & $75.3^{1/-21}$ & $80.4^{-21}$ & $59.3^{1/-21}$ & $77.7^{-20}$ & $65.8^{1/-21}$ & $81.7^{-21}$ & $68.7^{1/-21}$ \\
\midrule
ELMoPubMed & $\mathbf{99.3^{21}}$ & $96.1^{13/-6}$ & $\mathbf{99.2^{22}}$ & $95.2^{13/-6}$ & $99.0^{5}$ & $90.2^{9/-13}$ & $98.5^{3}$ & $94.5^{9}$ & $99.6^{7}$ & $95.8^{8/-8}$ \\
Flair & $96.8^{3/-19}$ & $87.9^{3/-19}$ & $96.6^{3/-19}$ & $85.3^{3/-19}$ & $94.1^{3/-18}$ & $74.8^{3/-19}$ & $93.8^{3}$ & $83.8^{3/-19}$ & $96.8^{3/-17}$ & $85.6^{3/-19}$ \\
SciBERT & $98.6^{6/-9}$ & $93.8^{7/-15}$ & $98.4^{7/-10}$ & $92.5^{7/-15}$ & $97.8^{4}$ & $87.1^{7/-15}$ & $97.7^{3}$ & $91.0^{7/-14}$ & $98.8^{4}$ & $92.8^{7/-15}$ \\
BERT & $97.5^{4/-18}$ & $89.9^{4/-18}$ & $97.1^{4/-18}$ & $87.2^{4/-18}$ & $96.3^{3/-5}$ & $81.5^{4/-17}$ & $95.8^{3}$ & $87.2^{4/-16}$ & $97.8^{3/-7}$ & $89.1^{4/-18}$ \\
ELMo & $98.0^{5/-17}$ & $91.7^{5/-17}$ & $97.7^{5/-17}$ & $89.0^{5/-17}$ & $97.8^{4}$ & $83.8^{6/-16}$ & $93.5^{2/-6}$ & $88.4^{4/-16}$ & $98.8^{4}$ & $90.4^{5/-16}$ \\
GPT & $98.4^{6/-13}$ & $92.2^{6/-16}$ & $98.1^{6/-15}$ & $89.9^{6/-16}$ & $97.4^{4}$ & $81.7^{4/-17}$ & $95.4^{3}$ & $86.9^{4/-16}$ & $98.6^{4}$ & $91.3^{5/-16}$ \\
    \end{tabular}   
    \caption{Accuracy of each embedding ($fJ$ for word embeddings, $\tau$ for contextual embeddings) on datasets created with random negative sampling. An embedding has significantly better/worse accuracy than the number of embeddings given by the positive/negative superscripts. Significance measured by McNemar's test with $\alpha = 0.0002$, i.e. $\alpha = 0.05$ with Bonferroni correction.}
    \label{tab:significance_newDatasets_subset_negR} 
\end{table}

\begin{table}[h]
    \centering
\scriptsize	
     \begin{tabular}{l l l l l l l l l l l}
\multirow{2}{*}{Dataset}& \textbf{\dataset{FSN-syn.}} & \textbf{\dataset{FSN-syn.}} & \textbf{\dataset{syn-syn.}} & \textbf{\dataset{syn-syn.}} & \textbf{\dataset{poss.-equ.}} & \textbf{\dataset{poss.-equ.}} & \textbf{\dataset{repl.-by}} & \textbf{\dataset{repl.-by}} & \textbf{\dataset{same-as}} & \textbf{\dataset{same-as}} \\
& \textbf{easy} & \textbf{hard} & \textbf{easy} & \textbf{hard} & \textbf{easy} & \textbf{hard} & \textbf{easy} & \textbf{hard} & \textbf{easy} & \textbf{hard} \\ 
Subset Size & 65.2\% & 58.6\% & 63.0\% & 56.5\% & 72.3\% & 69.0\% & 49.4\% & 62.9\% & 69.7\% & 71.0\% \\
\midrule
BioNLP PMC & $0.81$ & $0.55$ & $0.80$ & $0.50$ & $0.82$ & $0.48$ & $0.66$ & $0.56$ & $0.86$ & $0.60$ \\
BioNLP PM & $0.84$ & $0.56$ & $0.83$ & $0.52$ & $0.83$ & $0.50$ & $0.72$ & $0.60$ & $0.87$ & $0.61$ \\
BioNLP PP & $0.83$ & $0.55$ & $0.82$ & $0.50$ & $0.83$ & $0.49$ & $0.68$ & $0.57$ & $0.87$ & $0.60$ \\
BioNLP PPW & $0.83$ & $0.54$ & $0.82$ & $0.50$ & $0.83$ & $0.48$ & $0.67$ & $0.56$ & $0.86$ & $0.59$ \\
BioASQ & $0.84$ & $0.56$ & $0.83$ & $0.52$ & $0.87$ & $0.61$ & $0.73$ & $0.61$ & $0.87$ & $0.65$ \\
LTL win2 & $0.84$ & $0.52$ & $0.83$ & $0.48$ & $0.83$ & $0.43$ & $0.73$ & $0.54$ & $0.87$ & $0.55$ \\
LTL win30 & $0.80$ & $0.59$ & $0.80$ & $0.55$ & $0.84$ & $0.53$ & $0.72$ & $0.61$ & $0.88$ & $0.66$ \\
AUEB200 & $\mathbf{0.86}$ & $0.56$ & $0.84$ & $0.51$ & $0.87$ & $0.58$ & $0.74$ & $0.58$ & $0.89$ & $0.62$ \\
AUEB400 & $0.85$ & $0.56$ & $0.84$ & $0.52$ & $\mathbf{0.88}$ & $0.58$ & $\mathbf{0.75}$ & $0.58$ & $\mathbf{0.90}$ & $0.63$ \\
MeSH extr & $\mathbf{0.86}$ & $0.58$ & $\mathbf{0.85}$ & $0.53$ & $\mathbf{0.88}$ & $0.60$ & $0.74$ & $0.60$ & $\mathbf{0.90}$ & $0.65$ \\
MeSH intr & $0.84$ & $\mathbf{0.61}$ & $0.83$ & $\mathbf{0.56}$ & $\mathbf{0.88}$ & $\mathbf{0.63}$ & $0.74$ & $\mathbf{0.63}$ & $\mathbf{0.90}$ & $\mathbf{0.70}$ \\
MIMIC & $0.84$ & $0.55$ & $0.83$ & $0.51$ & $0.82$ & $0.43$ & $0.71$ & $0.55$ & $0.88$ & $0.58$ \\
MIMIC M & $0.84$ & $0.55$ & $0.83$ & $0.51$ & $0.82$ & $0.43$ & $0.71$ & $0.55$ & $0.88$ & $0.58$ \\
\midrule
GloVe & $0.78$ & $0.51$ & $0.78$ & $0.46$ & $0.83$ & $0.41$ & $0.68$ & $0.51$ & $0.86$ & $0.54$ \\
Fastt Wiki & $0.64$ & $0.41$ & $0.65$ & $0.37$ & $0.55$ & $0.27$ & $0.48$ & $0.33$ & $0.60$ & $0.35$ \\
Fastt Crawl & $0.70$ & $0.42$ & $0.70$ & $0.39$ & $0.62$ & $0.28$ & $0.54$ & $0.36$ & $0.68$ & $0.38$ \\
Fastt Crawl M & $0.67$ & $0.41$ & $0.67$ & $0.38$ & $0.56$ & $0.27$ & $0.50$ & $0.33$ & $0.63$ & $0.36$ \\
\midrule
ELMoPubMed & $0.83$ & $0.41$ & $0.83$ & $0.38$ & $0.83$ & $0.40$ & $0.71$ & $0.48$ & $0.87$ & $0.48$ \\
Flair & $0.76$ & $0.33$ & $0.76$ & $0.30$ & $0.75$ & $0.29$ & $0.61$ & $0.35$ & $0.78$ & $0.36$ \\
SciBERT & $0.64$ & $0.41$ & $0.67$ & $0.39$ & $0.78$ & $0.36$ & $0.64$ & $0.43$ & $0.77$ & $0.46$ \\
BERT & $0.73$ & $0.39$ & $0.73$ & $0.34$ & $0.81$ & $0.36$ & $0.69$ & $0.44$ & $0.81$ & $0.43$ \\
ELMo & $0.77$ & $0.40$ & $0.76$ & $0.35$ & $0.82$ & $0.41$ & $0.68$ & $0.45$ & $0.83$ & $0.46$ \\
GPT & $0.71$ & $0.41$ & $0.71$ & $0.38$ & $0.83$ & $0.35$ & $0.69$ & $0.44$ & $0.85$ & $0.47$ \\
    \end{tabular}     
    \caption{AUC of each embedding ($fJ$ for word embeddings, $\tau$ for contextual embeddings) for datasets created with Levenshtein negative sampling.}
    \label{tab:newDatasets_auc_negL} 
\end{table}

\begin{table}[h]
    \centering
\scriptsize	
     \begin{tabular}{l l l l l l l l l l l}
\multirow{2}{*}{Dataset}& \textbf{\dataset{FSN-syn.}} & \textbf{\dataset{FSN-syn.}} & \textbf{\dataset{syn-syn.}} & \textbf{\dataset{syn-syn.}} & \textbf{\dataset{poss.-equ.}} & \textbf{\dataset{poss.-equ.}} & \textbf{\dataset{repl.-by}} & \textbf{\dataset{repl.-by}} & \textbf{\dataset{same-as}} & \textbf{\dataset{same-as}} \\
& \textbf{easy} & \textbf{hard} & \textbf{easy} & \textbf{hard} & \textbf{easy} & \textbf{hard} & \textbf{easy} & \textbf{hard} & \textbf{easy} & \textbf{hard} \\ 
Subset Size & 58.0\% & 51.0\% & 55.3\% & 48.5\% & 69.5\% & 65.3\% & 40.0\% & 56.3\% & 65.5\% & 66.0\% \\
\midrule
BioNLP PMC & $0.99$ & $0.98$ & $0.99$ & $0.98$ & $1.0$ & $0.97$ & $0.99$ & $0.98$ & $0.99$ & $0.99$ \\
BioNLP PM & $1.0$ & $0.99$ & $1.0$ & $0.98$ & $1.0$ & $0.98$ & $0.99$ & $0.98$ & $1.0$ & $0.99$ \\
BioNLP PP & $1.0$ & $0.99$ & $1.0$ & $0.98$ & $1.0$ & $0.98$ & $0.99$ & $0.98$ & $1.0$ & $0.99$ \\
BioNLP PPW & $1.0$ & $0.98$ & $1.0$ & $0.98$ & $1.0$ & $0.97$ & $0.99$ & $0.98$ & $1.0$ & $0.99$ \\
BioASQ & $1.0$ & $0.99$ & $1.0$ & $0.99$ & $1.0$ & $0.99$ & $0.99$ & $0.98$ & $1.0$ & $0.99$ \\
LTL win2 & $1.0$ & $0.99$ & $1.0$ & $0.98$ & $0.99$ & $0.97$ & $0.99$ & $0.98$ & $1.0$ & $0.99$ \\
LTL win30 & $1.0$ & $0.99$ & $1.0$ & $0.98$ & $1.0$ & $0.98$ & $0.99$ & $0.98$ & $1.0$ & $0.99$ \\
AUEB200 & $1.0$ & $0.99$ & $1.0$ & $0.98$ & $1.0$ & $0.99$ & $0.99$ & $0.98$ & $1.0$ & $0.99$ \\
AUEB400 & $1.0$ & $0.99$ & $1.0$ & $0.99$ & $1.0$ & $0.99$ & $0.99$ & $0.98$ & $1.0$ & $1.0$ \\
MeSH extr & $1.0$ & $0.99$ & $1.0$ & $0.99$ & $1.0$ & $0.99$ & $0.99$ & $0.98$ & $1.0$ & $1.0$ \\
MeSH intr & $1.0$ & $0.99$ & $1.0$ & $0.99$ & $1.0$ & $0.99$ & $0.99$ & $0.98$ & $1.0$ & $1.0$ \\
MIMIC & $1.0$ & $0.99$ & $1.0$ & $0.98$ & $1.0$ & $0.97$ & $1.0$ & $0.98$ & $1.0$ & $0.99$ \\
MIMIC M & $1.0$ & $0.99$ & $1.0$ & $0.98$ & $1.0$ & $0.97$ & $1.0$ & $0.98$ & $1.0$ & $0.99$ \\
\midrule
GloVe & $1.0$ & $0.97$ & $0.99$ & $0.96$ & $1.0$ & $0.95$ & $0.99$ & $0.97$ & $1.0$ & $0.99$ \\
Fastt Wiki & $0.92$ & $0.81$ & $0.92$ & $0.78$ & $0.85$ & $0.61$ & $0.83$ & $0.69$ & $0.86$ & $0.71$ \\
Fastt Crawl & $0.95$ & $0.89$ & $0.95$ & $0.86$ & $0.91$ & $0.66$ & $0.91$ & $0.77$ & $0.94$ & $0.8$ \\
Fastt Crawl M & $0.92$ & $0.83$ & $0.93$ & $0.8$ & $0.86$ & $0.62$ & $0.85$ & $0.69$ & $0.89$ & $0.73$ \\
\midrule
ELMoPubMed & $1.0$ & $0.99$ & $1.0$ & $0.99$ & $1.0$ & $0.96$ & $0.99$ & $0.98$ & $1.0$ & $0.99$ \\
Flair & $0.99$ & $0.94$ & $0.99$ & $0.92$ & $0.98$ & $0.82$ & $0.98$ & $0.91$ & $0.99$ & $0.92$ \\
SciBERT & $1.0$ & $0.98$ & $1.0$ & $0.97$ & $0.99$ & $0.94$ & $0.99$ & $0.96$ & $1.0$ & $0.98$ \\
BERT & $0.99$ & $0.95$ & $0.99$ & $0.93$ & $0.99$ & $0.89$ & $0.98$ & $0.94$ & $0.99$ & $0.95$ \\
ELMo & $1.0$ & $0.96$ & $0.99$ & $0.94$ & $0.99$ & $0.91$ & $0.97$ & $0.94$ & $1.0$ & $0.96$ \\
GPT & $0.99$ & $0.95$ & $0.99$ & $0.93$ & $0.99$ & $0.86$ & $0.98$ & $0.9$ & $1.0$ & $0.95$ \\

    \end{tabular}     
    \caption{AUC of each embedding ($fJ$ for word embeddings, $\tau$ for contextual embeddings) for datasets created with random negative sampling.}
    \label{tab:newDatasets_auc_negR} 
\end{table}

\begin{sidewaystable*}
    \centering
    \small
    \begin{tabular}{l l l l l l l l l l l }
    \toprule
         Dataset & \dataset{FSN-syn.} & \dataset{FSN-syn.} & \dataset{syn-syn.} & \dataset{syn-syn.}& \dataset{poss.-equ.} & \dataset{poss.-equ.} & \dataset{repl.-by} & \dataset{repl.-by} & \dataset{same-as} & \dataset{same-as} \\
& easy & hard & easy & hard& easy & hard & easy & hard & easy & hard \\ 
         \midrule
\multirow{10}{1.7cm}{BioNLP PMC} & \multirow{10}{1.7cm}{$cP^{-7}$, $pP^{-7}$, $sP^{-7}$, $kP^{3/-6}$, $fJ^{9}$, $mJ^{4/-5}$, $cA^{5/-1}$, $pA^{5/-1}$, $sA^{5/-1}$, $kA^{5/-1}$} & \multirow{10}{1.7cm}{$cP^{-7}$, $pP^{-7}$, $sP^{-7}$, $kP^{3/-6}$, $fJ^{9}$, $mJ^{4/-1}$, $cA^{4/-2}$, $pA^{6/-1}$, $sA^{4/-2}$, $kA^{4/-1}$} & \multirow{10}{1.7cm}{$cP^{-7}$, $pP^{-7}$, $sP^{-7}$, $kP^{3/-6}$, $fJ^{9}$, $mJ^{4/-5}$, $cA^{5/-1}$, $pA^{5/-1}$, $sA^{5/-1}$, $kA^{5/-1}$} & \multirow{10}{1.7cm}{$cP^{-6}$, $pP^{-6}$, $sP^{-6}$, $kP^{-6}$, $fJ^{8}$, $mJ^{8}$, $cA^{4/-2}$, $pA^{4/-2}$, $sA^{4/-2}$, $kA^{4/-2}$} & \multirow{10}{1.7cm}{$fJ^{1}$, $mJ^{-5}$, $cA^{1}$, $pA^{1}$, $sA^{1}$, $kA^{1}$} & \multirow{10}{1.7cm}{$cP^{-5}$, $pP^{-5}$, $sP^{-5}$, $kP^{1/-5}$, $fJ^{5}$, $mJ^{-6}$, $cA^{5}$, $pA^{5}$, $sA^{5}$, $kA^{5}$} & \multirow{10}{1.7cm}{} & \multirow{10}{1.7cm}{$cP^{-1}$, $pP^{1/-3}$, $sP^{-5}$, $kP^{-4}$, $fJ^{2}$, $mJ^{-6}$, $cA^{4}$, $pA^{3}$, $sA^{4}$, $kA^{5}$} & \multirow{10}{1.7cm}{$cP^{-5}$, $pP^{-5}$, $sP^{-5}$, $kP^{-5}$, $fJ^{5}$, $mJ^{-5}$, $cA^{5}$, $pA^{5}$, $sA^{5}$, $kA^{5}$} & \multirow{10}{1.7cm}{$cP^{-6}$, $pP^{-6}$, $sP^{-7}$, $kP^{1/-6}$, $fJ^{4}$, $mJ^{4}$, $cA^{4}$, $pA^{4}$, $sA^{4}$, $kA^{4}$} \\ \\ \\ \\ \\ \\ \\ \\ \\ \\
\midrule
\multirow{10}{1.7cm}{BioNLP PM} & \multirow{10}{1.7cm}{$cP^{-7}$, $pP^{-7}$, $sP^{-7}$, $kP^{3/-6}$, $fJ^{9}$, $mJ^{4/-5}$, $cA^{5/-1}$, $pA^{5/-1}$, $sA^{5/-1}$, $kA^{5/-1}$} & \multirow{10}{1.7cm}{$cP^{-6}$, $pP^{-6}$, $sP^{-6}$, $kP^{-6}$, $fJ^{9}$, $mJ^{8/-1}$, $cA^{4/-2}$, $pA^{4/-2}$, $sA^{4/-2}$, $kA^{4/-2}$} & \multirow{10}{1.7cm}{$cP^{-7}$, $pP^{-7}$, $sP^{-7}$, $kP^{3/-6}$, $fJ^{9}$, $mJ^{4/-5}$, $cA^{6/-1}$, $pA^{5/-1}$, $sA^{5/-1}$, $kA^{5/-2}$} & \multirow{10}{1.7cm}{$cP^{-6}$, $pP^{-6}$, $sP^{-6}$, $kP^{-6}$, $fJ^{9}$, $mJ^{8/-1}$, $cA^{4/-2}$, $pA^{4/-2}$, $sA^{4/-2}$, $kA^{4/-2}$} & \multirow{10}{1.7cm}{$cP^{-5}$, $pP^{-5}$, $sP^{-5}$, $fJ^{4}$, $mJ^{-5}$, $cA^{4}$, $pA^{4}$, $sA^{4}$, $kA^{4}$} & \multirow{10}{1.7cm}{$cP^{-5}$, $pP^{-5}$, $sP^{-5}$, $kP^{-5}$, $fJ^{5/-2}$, $mJ^{-5}$, $cA^{6}$, $pA^{5}$, $sA^{6}$, $kA^{5}$} & \multirow{10}{1.7cm}{$fJ^{1}$, $mJ^{-1}$} & \multirow{10}{1.7cm}{$cP^{-5}$, $pP^{-5}$, $sP^{-5}$, $kP^{-5}$, $fJ^{5}$, $mJ^{-5}$, $cA^{5}$, $pA^{5}$, $sA^{5}$, $kA^{5}$} & \multirow{10}{1.7cm}{$cP^{-6}$, $pP^{-7}$, $sP^{-5}$, $kP^{1/-5}$, $fJ^{5}$, $mJ^{2/-5}$, $cA^{5}$, $pA^{5}$, $sA^{5}$, $kA^{5}$} & \multirow{10}{1.7cm}{$cP^{-7}$, $pP^{-7}$, $sP^{-7}$, $kP^{3/-6}$, $fJ^{4}$, $mJ^{4}$, $cA^{4}$, $pA^{4}$, $sA^{4}$, $kA^{4}$} \\ \\ \\ \\ \\ \\ \\ \\ \\ \\
\midrule
\multirow{10}{1.7cm}{BioNLP PP} & \multirow{10}{1.7cm}{$cP^{-7}$, $pP^{-7}$, $sP^{-7}$, $kP^{3/-6}$, $fJ^{9}$, $mJ^{4/-5}$, $cA^{5/-1}$, $pA^{5/-1}$, $sA^{5/-1}$, $kA^{5/-1}$} & \multirow{10}{1.7cm}{$cP^{-6}$, $pP^{-6}$, $sP^{-6}$, $kP^{-6}$, $fJ^{8/-1}$, $mJ^{9}$, $cA^{4/-2}$, $pA^{4/-2}$, $sA^{4/-2}$, $kA^{4/-2}$} & \multirow{10}{1.7cm}{$cP^{-7}$, $pP^{-7}$, $sP^{-7}$, $kP^{3/-6}$, $fJ^{9}$, $mJ^{4/-5}$, $cA^{5/-1}$, $pA^{5/-1}$, $sA^{5/-1}$, $kA^{5/-1}$} & \multirow{10}{1.7cm}{$cP^{-6}$, $pP^{-6}$, $sP^{-6}$, $kP^{-6}$, $fJ^{8}$, $mJ^{8}$, $cA^{4/-2}$, $pA^{4/-2}$, $sA^{4/-2}$, $kA^{4/-2}$} & \multirow{10}{1.7cm}{$sP^{-1}$, $fJ^{2}$, $mJ^{-4}$, $cA^{1}$, $sA^{1}$, $kA^{1}$} & \multirow{10}{1.7cm}{$cP^{-5}$, $pP^{-5}$, $sP^{-5}$, $kP^{-5}$, $fJ^{5}$, $mJ^{-5}$, $cA^{5}$, $pA^{5}$, $sA^{5}$, $kA^{5}$} & \multirow{10}{1.7cm}{} & \multirow{10}{1.7cm}{$cP^{-5}$, $pP^{-5}$, $sP^{-5}$, $kP^{-5}$, $fJ^{5}$, $mJ^{-5}$, $cA^{5}$, $pA^{5}$, $sA^{5}$, $kA^{5}$} & \multirow{10}{1.7cm}{$cP^{-5}$, $pP^{-6}$, $sP^{-5}$, $kP^{-5}$, $fJ^{5}$, $mJ^{1/-2}$, $cA^{4}$, $pA^{5}$, $sA^{4}$, $kA^{4}$} & \multirow{10}{1.7cm}{$cP^{-6}$, $pP^{-6}$, $sP^{-6}$, $kP^{-6}$, $fJ^{4}$, $mJ^{4}$, $cA^{4}$, $pA^{4}$, $sA^{4}$, $kA^{4}$} \\ \\ \\ \\ \\ \\ \\ \\ \\ \\
\midrule
\multirow{10}{1.7cm}{BioNLP PPW} & \multirow{10}{1.7cm}{$cP^{-7}$, $pP^{-7}$, $sP^{-7}$, $kP^{3/-6}$, $fJ^{9}$, $mJ^{4/-5}$, $cA^{5/-1}$, $pA^{5/-1}$, $sA^{5/-1}$, $kA^{5/-1}$} & \multirow{10}{1.7cm}{$cP^{-6}$, $pP^{-6}$, $sP^{-6}$, $kP^{-6}$, $fJ^{9}$, $mJ^{8/-1}$, $cA^{4/-2}$, $pA^{4/-2}$, $sA^{4/-2}$, $kA^{4/-2}$} & \multirow{10}{1.7cm}{$cP^{-7}$, $pP^{-7}$, $sP^{-7}$, $kP^{3/-6}$, $fJ^{9}$, $mJ^{4/-5}$, $cA^{5/-1}$, $pA^{5/-1}$, $sA^{5/-1}$, $kA^{5/-1}$} & \multirow{10}{1.7cm}{$cP^{-6}$, $pP^{-6}$, $sP^{-6}$, $kP^{-6}$, $fJ^{8}$, $mJ^{8}$, $cA^{4/-2}$, $pA^{4/-2}$, $sA^{4/-2}$, $kA^{4/-2}$} & \multirow{10}{1.7cm}{$fJ^{1}$, $mJ^{-1}$} & \multirow{10}{1.7cm}{$cP^{-5}$, $pP^{-5}$, $sP^{-5}$, $kP^{-5}$, $fJ^{5}$, $mJ^{-5}$, $cA^{5}$, $pA^{5}$, $sA^{5}$, $kA^{5}$} & \multirow{10}{1.7cm}{} & \multirow{10}{1.7cm}{$cP^{-4}$, $pP^{-4}$, $sP^{-5}$, $kP^{-5}$, $fJ^{5}$, $mJ^{-5}$, $cA^{5}$, $pA^{5}$, $sA^{5}$, $kA^{3}$} & \multirow{10}{1.7cm}{$cP^{-5}$, $pP^{-5}$, $sP^{-5}$, $kP^{-5}$, $fJ^{5}$, $mJ^{-3}$, $cA^{5}$, $pA^{4}$, $sA^{5}$, $kA^{4}$} & \multirow{10}{1.7cm}{$cP^{-6}$, $pP^{-6}$, $sP^{-6}$, $kP^{-6}$, $fJ^{6}$, $mJ^{4}$, $cA^{4/-1}$, $pA^{4}$, $sA^{4/-1}$, $kA^{4}$} \\ \\ \\ \\ \\ \\ \\ \\ \\ \\

\bottomrule
    \end{tabular}
\caption{Similarity metrics for each embedding and dataset (with Levenshtein negative sampling) that have significantly better/worse accuracy than the number of methods given by the positive/negative superscripts. Significance measured by McNemar's test with $\alpha = 0.001$/$0.008$ (word/sentence embeddings), i.e. $\alpha = 0.05$ with Bonferroni correction.}
    \label{tab:significance_snomedDatasets_methods1} 
\end{sidewaystable*}

\begin{sidewaystable*}
    \centering
    \small
    \begin{tabular}{l l l l l l l l l l l }
    \toprule
         Dataset & \dataset{FSN-syn.} & \dataset{FSN-syn.} & \dataset{syn-syn.} & \dataset{syn-syn.}& \dataset{poss.-equ.} & \dataset{poss.-equ.} & \dataset{repl.-by} & \dataset{repl.-by} & \dataset{same-as} & \dataset{same-as} \\
& easy & hard & easy & hard& easy & hard & easy & hard & easy & hard \\ 
         \midrule
\multirow{10}{1.7cm}{BioASQ} & \multirow{10}{1.7cm}{$cP^{-7}$, $pP^{-7}$, $sP^{-7}$, $kP^{3/-6}$, $fJ^{9}$, $mJ^{4/-5}$, $cA^{5/-1}$, $pA^{5/-1}$, $sA^{5/-1}$, $kA^{5/-1}$} & \multirow{10}{1.7cm}{$cP^{-6}$, $pP^{-6}$, $sP^{-6}$, $kP^{-6}$, $fJ^{4/-1}$, $mJ^{4}$, $cA^{4/-2}$, $pA^{5}$, $sA^{4/-1}$, $kA^{7}$} & \multirow{10}{1.7cm}{$cP^{-7}$, $pP^{-7}$, $sP^{-7}$, $kP^{3/-6}$, $fJ^{9}$, $mJ^{4/-5}$, $cA^{5/-1}$, $pA^{5/-1}$, $sA^{5/-1}$, $kA^{5/-1}$} & \multirow{10}{1.7cm}{$cP^{-6}$, $pP^{-6}$, $sP^{-6}$, $kP^{-6}$, $fJ^{9}$, $mJ^{8/-1}$, $cA^{4/-2}$, $pA^{4/-2}$, $sA^{4/-2}$, $kA^{4/-2}$} & \multirow{10}{1.7cm}{} & \multirow{10}{1.7cm}{$cP^{1/-6}$, $pP^{1/-6}$, $sP^{1/-6}$, $kP^{-9}$, $fJ^{5/-4}$, $mJ^{4/-5}$, $cA^{6}$, $pA^{6}$, $sA^{6}$, $kA^{6}$} & \multirow{10}{1.7cm}{} & \multirow{10}{1.7cm}{$cP^{-5}$, $pP^{-5}$, $sP^{-5}$, $kP^{-5}$, $fJ^{5}$, $mJ^{-5}$, $cA^{5}$, $pA^{5}$, $sA^{5}$, $kA^{5}$} & \multirow{10}{1.7cm}{$cP^{-5}$, $pP^{-5}$, $sP^{-5}$, $kP^{-5}$, $fJ^{5}$, $mJ^{-5}$, $cA^{5}$, $pA^{5}$, $sA^{5}$, $kA^{5}$} & \multirow{10}{1.7cm}{$cP^{-6}$, $pP^{-6}$, $sP^{-6}$, $kP^{-6}$, $fJ^{4}$, $mJ^{4}$, $cA^{4}$, $pA^{4}$, $sA^{4}$, $kA^{4}$} \\ \\ \\ \\ \\ \\ \\ \\ \\ \\
\midrule
\multirow{10}{1.7cm}{LTL win2} & \multirow{10}{1.7cm}{$cP^{-7}$, $pP^{-7}$, $sP^{-7}$, $kP^{3/-6}$, $fJ^{9}$, $mJ^{4/-5}$, $cA^{5/-1}$, $pA^{5/-1}$, $sA^{5/-1}$, $kA^{5/-1}$} & \multirow{10}{1.7cm}{$cP^{-6}$, $pP^{-6}$, $sP^{-6}$, $kP^{-6}$, $fJ^{4}$, $mJ^{4}$, $cA^{4}$, $pA^{4}$, $sA^{4}$, $kA^{4}$} & \multirow{10}{1.7cm}{$cP^{-7}$, $pP^{-7}$, $sP^{-7}$, $kP^{3/-6}$, $fJ^{9}$, $mJ^{4/-5}$, $cA^{5/-1}$, $pA^{5/-1}$, $sA^{5/-1}$, $kA^{5/-1}$} & \multirow{10}{1.7cm}{$cP^{-7}$, $pP^{-7}$, $sP^{-7}$, $kP^{3/-6}$, $fJ^{8/-1}$, $mJ^{9}$, $cA^{4/-2}$, $pA^{4/-2}$, $sA^{4/-2}$, $kA^{4/-2}$} & \multirow{10}{1.7cm}{$sP^{-1}$, $fJ^{6}$, $mJ^{-1}$, $cA^{-1}$, $pA^{-1}$, $sA^{-1}$, $kA^{-1}$} & \multirow{10}{1.7cm}{$cP^{-5}$, $pP^{-5}$, $sP^{-6}$, $kP^{2/-5}$, $fJ^{5}$, $mJ^{-6}$, $cA^{5}$, $pA^{5}$, $sA^{5}$, $kA^{5}$} & \multirow{10}{1.7cm}{} & \multirow{10}{1.7cm}{$cP^{-1}$, $pP^{-1}$, $sP^{-1}$, $kP^{1/-1}$, $fJ^{5}$, $mJ^{-6}$, $cA^{1}$, $pA^{1}$, $sA^{1}$, $kA^{1}$} & \multirow{10}{1.7cm}{$cP^{-5}$, $pP^{-5}$, $sP^{-5}$, $kP^{-3}$, $fJ^{5}$, $mJ^{-1}$, $cA^{3}$, $pA^{4}$, $sA^{3}$, $kA^{4}$} & \multirow{10}{1.7cm}{$cP^{-5}$, $pP^{-5}$, $sP^{-5}$, $kP^{-6}$, $fJ^{4}$, $mJ^{1/-4}$, $cA^{5}$, $pA^{5}$, $sA^{5}$, $kA^{5}$} \\ \\ \\ \\ \\ \\ \\ \\ \\ \\
\midrule
\multirow{10}{1.7cm}{LTL win30} & \multirow{10}{1.7cm}{$cP^{-7}$, $pP^{-7}$, $sP^{-7}$, $kP^{3/-6}$, $fJ^{9}$, $mJ^{4/-5}$, $cA^{5/-1}$, $pA^{5/-1}$, $sA^{5/-1}$, $kA^{5/-1}$} & \multirow{10}{1.7cm}{$cP^{-7}$, $pP^{-7}$, $sP^{-7}$, $kP^{3/-6}$, $fJ^{9}$, $mJ^{8/-1}$, $cA^{4/-2}$, $pA^{4/-2}$, $sA^{4/-2}$, $kA^{4/-2}$} & \multirow{10}{1.7cm}{$cP^{-7}$, $pP^{-7}$, $sP^{-7}$, $kP^{3/-6}$, $fJ^{9}$, $mJ^{4/-5}$, $cA^{5/-1}$, $pA^{5/-1}$, $sA^{5/-1}$, $kA^{5/-1}$} & \multirow{10}{1.7cm}{$cP^{-7}$, $pP^{-7}$, $sP^{-7}$, $kP^{3/-6}$, $fJ^{9}$, $mJ^{8/-1}$, $cA^{4/-2}$, $pA^{4/-2}$, $sA^{4/-2}$, $kA^{4/-2}$} & \multirow{10}{1.7cm}{$cP^{-1}$, $pP^{-1}$, $sP^{-1}$, $kP^{-1}$, $fJ^{9}$, $mJ^{-1}$, $cA^{-1}$, $pA^{-1}$, $sA^{-1}$, $kA^{-1}$} & \multirow{10}{1.7cm}{$cP^{-5}$, $pP^{-5}$, $sP^{-5}$, $kP^{-5}$, $fJ^{9}$, $mJ^{-5}$, $cA^{5/-1}$, $pA^{5/-1}$, $sA^{5/-1}$, $kA^{5/-1}$} & \multirow{10}{1.7cm}{} & \multirow{10}{1.7cm}{$cP^{-5}$, $pP^{-5}$, $sP^{-5}$, $kP^{-1}$, $fJ^{9}$, $mJ^{-1}$, $cA^{3/-1}$, $pA^{3/-1}$, $sA^{3/-1}$, $kA^{3/-1}$} & \multirow{10}{1.7cm}{$cP^{-5}$, $pP^{-5}$, $sP^{-5}$, $kP^{-5}$, $fJ^{9}$, $mJ^{-4}$, $cA^{5/-1}$, $pA^{4/-1}$, $sA^{5/-1}$, $kA^{5/-1}$} & \multirow{10}{1.7cm}{$cP^{-7}$, $pP^{-7}$, $sP^{2/-6}$, $kP^{-6}$, $fJ^{8}$, $mJ^{6}$, $cA^{4/-2}$, $pA^{4/-1}$, $sA^{4/-2}$, $kA^{4/-1}$} \\ \\ \\ \\ \\ \\ \\ \\ \\ \\
\midrule
\multirow{10}{1.7cm}{AUEB200} & \multirow{10}{1.7cm}{$cP^{-7}$, $pP^{-7}$, $sP^{-7}$, $kP^{3/-6}$, $fJ^{9}$, $mJ^{4/-5}$, $cA^{5/-1}$, $pA^{5/-1}$, $sA^{5/-1}$, $kA^{5/-1}$} & \multirow{10}{1.7cm}{$cP^{-3}$, $pP^{-3}$, $sP^{-3}$, $kP^{-3}$, $fJ^{9}$, $mJ^{8/-1}$, $cA^{-3}$, $pA^{-3}$, $sA^{-3}$, $kA^{7/-2}$} & \multirow{10}{1.7cm}{$cP^{-6}$, $pP^{-6}$, $sP^{-7}$, $kP^{1/-6}$, $fJ^{9}$, $mJ^{4/-5}$, $cA^{5/-1}$, $pA^{5/-1}$, $sA^{5/-1}$, $kA^{5/-1}$} & \multirow{10}{1.7cm}{$cP^{-6}$, $pP^{-6}$, $sP^{-6}$, $kP^{-6}$, $fJ^{9}$, $mJ^{8/-1}$, $cA^{4/-2}$, $pA^{4/-2}$, $sA^{4/-2}$, $kA^{4/-2}$} & \multirow{10}{1.7cm}{$fJ^{5}$, $mJ^{-1}$, $cA^{-1}$, $pA^{-1}$, $sA^{-1}$, $kA^{-1}$} & \multirow{10}{1.7cm}{$cP^{3/-5}$, $pP^{3/-5}$, $sP^{2/-7}$, $kP^{-8}$, $fJ^{5/-4}$, $mJ^{-8}$, $cA^{6}$, $pA^{6/-1}$, $sA^{7}$, $kA^{6}$} & \multirow{10}{1.7cm}{} & \multirow{10}{1.7cm}{$cP^{1}$, $pP^{1}$, $sP^{1}$, $kP^{1}$, $fJ^{1}$, $mJ^{-9}$, $cA^{1}$, $pA^{1}$, $sA^{1}$, $kA^{1}$} & \multirow{10}{1.7cm}{$cP^{-6}$, $pP^{-5}$, $sP^{-5}$, $kP^{-5}$, $fJ^{5}$, $mJ^{1/-5}$, $cA^{5}$, $pA^{5}$, $sA^{5}$, $kA^{5}$} & \multirow{10}{1.7cm}{$cP^{-6}$, $pP^{-6}$, $sP^{-6}$, $kP^{-6}$, $fJ^{5}$, $mJ^{4/-5}$, $cA^{5}$, $pA^{5}$, $sA^{5}$, $kA^{5}$} \\ \\ \\ \\ \\ \\ \\ \\ \\ \\
\bottomrule
    \end{tabular}
\caption{Similarity metrics for each embedding and dataset (with Levenshtein negative sampling) that have significantly better/worse accuracy than the number of methods given by the positive/negative superscripts. Significance measured by McNemar's test with $\alpha = 0.001$/$0.008$ (word/sentence embeddings), i.e. $\alpha = 0.05$ with Bonferroni correction.}
    \label{tab:significance_snomedDatasets_methods2} 
\end{sidewaystable*}

\begin{sidewaystable*}
    \centering
    \small
    \begin{tabular}{l l l l l l l l l l l }
    \toprule
         Dataset & \dataset{FSN-syn.} & \dataset{FSN-syn.} & \dataset{syn-syn.} & \dataset{syn-syn.}& \dataset{poss.-equ.} & \dataset{poss.-equ.} & \dataset{repl.-by} & \dataset{repl.-by} & \dataset{same-as} & \dataset{same-as} \\
& easy & hard & easy & hard& easy & hard & easy & hard & easy & hard \\ 
         \midrule
\multirow{10}{1.7cm}{AUEB400} & \multirow{10}{1.7cm}{$cP^{-6}$, $pP^{-6}$, $sP^{-6}$, $kP^{-6}$, $fJ^{9}$, $mJ^{4/-5}$, $cA^{5/-1}$, $pA^{5/-1}$, $sA^{5/-1}$, $kA^{5/-1}$} & \multirow{10}{1.7cm}{$cP^{-5}$, $pP^{-5}$, $sP^{-5}$, $kP^{-5}$, $fJ^{9}$, $mJ^{8/-1}$, $cA^{4/-4}$, $pA^{6/-2}$, $sA^{-4}$, $kA^{6/-2}$} & \multirow{10}{1.7cm}{$cP^{-6}$, $pP^{-6}$, $sP^{-6}$, $kP^{-6}$, $fJ^{9}$, $mJ^{4/-5}$, $cA^{5/-1}$, $pA^{5/-1}$, $sA^{5/-1}$, $kA^{5/-1}$} & \multirow{10}{1.7cm}{$cP^{-6}$, $pP^{-6}$, $sP^{-6}$, $kP^{-6}$, $fJ^{9}$, $mJ^{8/-1}$, $cA^{4/-2}$, $pA^{4/-2}$, $sA^{4/-2}$, $kA^{4/-2}$} & \multirow{10}{1.7cm}{$fJ^{1}$, $mJ^{-5}$, $cA^{1}$, $pA^{1}$, $sA^{1}$, $kA^{1}$} & \multirow{10}{1.7cm}{$cP^{1/-5}$, $pP^{1/-5}$, $sP^{1/-5}$, $kP^{-8}$, $fJ^{5/-4}$, $mJ^{-5}$, $cA^{6}$, $pA^{6}$, $sA^{6}$, $kA^{6}$} & \multirow{10}{1.7cm}{} & \multirow{10}{1.7cm}{$fJ^{1}$, $mJ^{-5}$, $cA^{1}$, $pA^{1}$, $sA^{1}$, $kA^{1}$} & \multirow{10}{1.7cm}{$cP^{-5}$, $pP^{-5}$, $sP^{-5}$, $kP^{-5}$, $fJ^{5}$, $mJ^{-5}$, $cA^{5}$, $pA^{5}$, $sA^{5}$, $kA^{5}$} & \multirow{10}{1.7cm}{$cP^{-6}$, $pP^{-6}$, $sP^{-6}$, $kP^{-6}$, $fJ^{4}$, $mJ^{4/-4}$, $cA^{5}$, $pA^{5}$, $sA^{5}$, $kA^{5}$} \\ \\ \\ \\ \\ \\ \\ \\ \\ \\
\midrule
\multirow{10}{1.7cm}{MeSH extr} & \multirow{10}{1.7cm}{$cP^{-7}$, $pP^{-7}$, $sP^{-7}$, $kP^{3/-6}$, $fJ^{9}$, $mJ^{4/-5}$, $cA^{5/-1}$, $pA^{5/-1}$, $sA^{5/-1}$, $kA^{5/-1}$} & \multirow{10}{1.7cm}{$cP^{-5}$, $pP^{-5}$, $sP^{-5}$, $kP^{-5}$, $fJ^{9}$, $mJ^{8/-1}$, $cA^{7/-2}$, $pA^{5/-3}$, $sA^{-5}$, $kA^{5/-3}$} & \multirow{10}{1.7cm}{$cP^{-7}$, $pP^{-7}$, $sP^{-7}$, $kP^{3/-6}$, $fJ^{9}$, $mJ^{4/-5}$, $cA^{5/-1}$, $pA^{5/-1}$, $sA^{5/-1}$, $kA^{5/-1}$} & \multirow{10}{1.7cm}{$cP^{-6}$, $pP^{-6}$, $sP^{-6}$, $kP^{-6}$, $fJ^{9}$, $mJ^{8/-1}$, $cA^{4/-2}$, $pA^{4/-2}$, $sA^{4/-2}$, $kA^{4/-2}$} & \multirow{10}{1.7cm}{$fJ^{5}$, $mJ^{-1}$, $cA^{-1}$, $pA^{-1}$, $sA^{-1}$, $kA^{-1}$} & \multirow{10}{1.7cm}{$cP^{1/-6}$, $pP^{1/-6}$, $sP^{1/-6}$, $kP^{-9}$, $fJ^{5}$, $mJ^{4/-5}$, $cA^{5}$, $pA^{5}$, $sA^{5}$, $kA^{5}$} & \multirow{10}{1.7cm}{} & \multirow{10}{1.7cm}{$cP^{-1}$, $fJ^{6}$, $mJ^{-1}$, $cA^{-1}$, $pA^{-1}$, $sA^{-1}$, $kA^{-1}$} & \multirow{10}{1.7cm}{$cP^{-5}$, $pP^{-5}$, $sP^{-5}$, $kP^{-4}$, $fJ^{5}$, $mJ^{-2}$, $cA^{4}$, $pA^{4}$, $sA^{3}$, $kA^{5}$} & \multirow{10}{1.7cm}{$cP^{-6}$, $pP^{-6}$, $sP^{-6}$, $kP^{-6}$, $fJ^{4}$, $mJ^{4}$, $cA^{4}$, $pA^{4}$, $sA^{4}$, $kA^{4}$} \\ \\ \\ \\ \\ \\ \\ \\ \\ \\
\midrule
\multirow{10}{1.7cm}{MeSH intr} & \multirow{10}{1.7cm}{$cP^{-7}$, $pP^{-7}$, $sP^{-7}$, $kP^{3/-6}$, $fJ^{9}$, $mJ^{4/-5}$, $cA^{5/-1}$, $pA^{5/-1}$, $sA^{5/-1}$, $kA^{5/-1}$} & \multirow{10}{1.7cm}{$cP^{-6}$, $pP^{-6}$, $sP^{-6}$, $kP^{-6}$, $fJ^{9}$, $mJ^{8/-1}$, $cA^{4/-2}$, $pA^{4/-2}$, $sA^{4/-2}$, $kA^{4/-2}$} & \multirow{10}{1.7cm}{$cP^{-7}$, $pP^{-7}$, $sP^{-7}$, $kP^{3/-6}$, $fJ^{9}$, $mJ^{4/-5}$, $cA^{5/-1}$, $pA^{5/-1}$, $sA^{5/-1}$, $kA^{5/-1}$} & \multirow{10}{1.7cm}{$cP^{-6}$, $pP^{-6}$, $sP^{-6}$, $kP^{-6}$, $fJ^{9}$, $mJ^{8/-1}$, $cA^{4/-2}$, $pA^{4/-2}$, $sA^{4/-2}$, $kA^{4/-2}$} & \multirow{10}{1.7cm}{$sP^{-1}$, $fJ^{6}$, $mJ^{-1}$, $cA^{-1}$, $pA^{-1}$, $sA^{-1}$, $kA^{-1}$} & \multirow{10}{1.7cm}{$cP^{1/-6}$, $pP^{1/-6}$, $sP^{1/-6}$, $kP^{-9}$, $fJ^{9}$, $mJ^{4/-5}$, $cA^{5/-1}$, $pA^{5/-1}$, $sA^{5/-1}$, $kA^{5/-1}$} & \multirow{10}{1.7cm}{} & \multirow{10}{1.7cm}{$cP^{-1}$, $pP^{-1}$, $sP^{-1}$, $kP^{-1}$, $fJ^{9}$, $mJ^{-1}$, $cA^{-1}$, $pA^{-1}$, $sA^{-1}$, $kA^{-1}$} & \multirow{10}{1.7cm}{$cP^{-6}$, $pP^{-7}$, $sP^{-5}$, $kP^{1/-1}$, $fJ^{9}$, $mJ^{2/-1}$, $cA^{3/-1}$, $pA^{3/-1}$, $sA^{3/-1}$, $kA^{3/-1}$} & \multirow{10}{1.7cm}{$cP^{-6}$, $pP^{-6}$, $sP^{-6}$, $kP^{-6}$, $fJ^{9}$, $mJ^{8/-1}$, $cA^{4/-2}$, $pA^{4/-2}$, $sA^{4/-2}$, $kA^{4/-2}$} \\ \\ \\ \\ \\ \\ \\ \\ \\ \\
\midrule
\multirow{10}{1.7cm}{MIMIC} & \multirow{10}{1.7cm}{$cP^{-7}$, $pP^{-7}$, $sP^{-7}$, $kP^{3/-6}$, $fJ^{9}$, $mJ^{4/-5}$, $cA^{7/-1}$, $pA^{5/-3}$, $sA^{7/-1}$, $kA^{5/-3}$} & \multirow{10}{1.7cm}{$cP^{-6}$, $pP^{-6}$, $sP^{-6}$, $kP^{-6}$, $fJ^{9}$, $mJ^{4/-5}$, $cA^{5/-1}$, $pA^{5/-1}$, $sA^{5/-1}$, $kA^{5/-1}$} & \multirow{10}{1.7cm}{$cP^{-7}$, $pP^{-7}$, $sP^{-7}$, $kP^{3/-6}$, $fJ^{9}$, $mJ^{4/-5}$, $cA^{7/-1}$, $pA^{5/-3}$, $sA^{7/-1}$, $kA^{5/-3}$} & \multirow{10}{1.7cm}{$cP^{-6}$, $pP^{-6}$, $sP^{-6}$, $kP^{-6}$, $fJ^{8}$, $mJ^{8}$, $cA^{4/-2}$, $pA^{4/-2}$, $sA^{4/-2}$, $kA^{4/-2}$} & \multirow{10}{1.7cm}{$cP^{-1}$, $pP^{-1}$, $kP^{3}$, $fJ^{5}$, $mJ^{-6}$, $cA^{1/-1}$, $pA^{1/-1}$, $sA^{1/-1}$, $kA^{1/-1}$} & \multirow{10}{1.7cm}{$cP^{1/-5}$, $pP^{1/-6}$, $sP^{1/-6}$, $kP^{3/-5}$, $fJ^{5}$, $mJ^{-9}$, $cA^{5}$, $pA^{5}$, $sA^{5}$, $kA^{5}$} & \multirow{10}{1.7cm}{$fJ^{1}$, $kA^{-1}$} & \multirow{10}{1.7cm}{$cP^{-1}$, $pP^{-1}$, $sP^{-1}$, $kP^{-1}$, $fJ^{9}$, $mJ^{-5}$, $cA^{1/-1}$, $pA^{1/-1}$, $sA^{1/-1}$, $kA^{1/-1}$} & \multirow{10}{1.7cm}{$cP^{-4}$, $pP^{-4}$, $sP^{-5}$, $kP^{-2}$, $fJ^{9}$, $mJ^{-1}$, $cA^{3/-1}$, $pA^{4/-1}$, $sA^{1/-1}$, $kA^{3/-1}$} & \multirow{10}{1.7cm}{$cP^{-7}$, $pP^{-7}$, $sP^{-7}$, $kP^{3/-6}$, $fJ^{9}$, $mJ^{4/-1}$, $cA^{5/-1}$, $pA^{4/-3}$, $sA^{5/-1}$, $kA^{4/-1}$} \\ \\ \\ \\ \\ \\ \\ \\ \\ \\
\bottomrule
    \end{tabular}
\caption{Similarity metrics for each embedding and dataset (with Levenshtein negative sampling) that have significantly better/worse accuracy than the number of methods given by the positive/negative superscripts. Significance measured by McNemar's test with $\alpha = 0.001$/$0.008$ (word/sentence embeddings), i.e. $\alpha = 0.05$ with Bonferroni correction.}
    \label{tab:significance_snomedDatasets_methods3} 
\end{sidewaystable*}

\begin{sidewaystable*}
    \centering
    \small
    \begin{tabular}{l l l l l l l l l l l }
    \toprule
         Dataset & \dataset{FSN-syn.} & \dataset{FSN-syn.} & \dataset{syn-syn.} & \dataset{syn-syn.}& \dataset{poss.-equ.} & \dataset{poss.-equ.} & \dataset{repl.-by} & \dataset{repl.-by} & \dataset{same-as} & \dataset{same-as} \\
& easy & hard & easy & hard& easy & hard & easy & hard & easy & hard \\ 
         \midrule
\multirow{10}{1.7cm}{MIMIC M} & \multirow{10}{1.7cm}{$cP^{-7}$, $pP^{-7}$, $sP^{-7}$, $kP^{3/-6}$, $fJ^{9}$, $mJ^{4/-5}$, $cA^{7/-1}$, $pA^{5/-3}$, $sA^{7/-1}$, $kA^{5/-3}$} & \multirow{10}{1.7cm}{$cP^{-6}$, $pP^{-6}$, $sP^{-6}$, $kP^{-6}$, $fJ^{9}$, $mJ^{4/-5}$, $cA^{5/-1}$, $pA^{5/-1}$, $sA^{5/-1}$, $kA^{5/-1}$} & \multirow{10}{1.7cm}{$cP^{-7}$, $pP^{-7}$, $sP^{-7}$, $kP^{3/-6}$, $fJ^{9}$, $mJ^{4/-5}$, $cA^{7/-1}$, $pA^{5/-3}$, $sA^{7/-1}$, $kA^{5/-3}$} & \multirow{10}{1.7cm}{$cP^{-6}$, $pP^{-6}$, $sP^{-6}$, $kP^{-6}$, $fJ^{8}$, $mJ^{8}$, $cA^{4/-2}$, $pA^{4/-2}$, $sA^{4/-2}$, $kA^{4/-2}$} & \multirow{10}{1.7cm}{$cP^{-1}$, $pP^{-1}$, $kP^{3}$, $fJ^{5}$, $mJ^{-6}$, $cA^{1/-1}$, $pA^{1/-1}$, $sA^{1/-1}$, $kA^{1/-1}$} & \multirow{10}{1.7cm}{$cP^{1/-5}$, $pP^{1/-6}$, $sP^{1/-6}$, $kP^{3/-5}$, $fJ^{5}$, $mJ^{-9}$, $cA^{5}$, $pA^{5}$, $sA^{5}$, $kA^{5}$} & \multirow{10}{1.7cm}{$fJ^{1}$, $kA^{-1}$} & \multirow{10}{1.7cm}{$cP^{-1}$, $pP^{-1}$, $sP^{-1}$, $kP^{-1}$, $fJ^{9}$, $mJ^{-5}$, $cA^{1/-1}$, $pA^{1/-1}$, $sA^{1/-1}$, $kA^{1/-1}$} & \multirow{10}{1.7cm}{$cP^{-5}$, $pP^{-3}$, $sP^{-5}$, $kP^{-1}$, $fJ^{9}$, $mJ^{-1}$, $cA^{3/-1}$, $pA^{3/-1}$, $sA^{2/-1}$, $kA^{2/-1}$} & \multirow{10}{1.7cm}{$cP^{-7}$, $pP^{-7}$, $sP^{-7}$, $kP^{3/-6}$, $fJ^{9}$, $mJ^{4/-1}$, $cA^{5/-1}$, $pA^{4/-3}$, $sA^{5/-1}$, $kA^{4/-1}$} \\ \\ \\ \\ \\ \\ \\ \\ \\ \\
\midrule
\multirow{10}{1.7cm}{GloVe} & \multirow{10}{1.7cm}{$cP^{-7}$, $pP^{-7}$, $sP^{-7}$, $kP^{3/-6}$, $fJ^{9}$, $mJ^{4/-5}$, $cA^{5/-3}$, $pA^{7/-1}$, $sA^{5/-3}$, $kA^{7/-1}$} & \multirow{10}{1.7cm}{$cP^{-7}$, $pP^{-7}$, $sP^{-7}$, $kP^{3/-6}$, $fJ^{4/-3}$, $mJ^{7}$, $cA^{4/-3}$, $pA^{7}$, $sA^{4/-3}$, $kA^{7}$} & \multirow{10}{1.7cm}{$cP^{-8}$, $pP^{-8}$, $sP^{2/-7}$, $kP^{3/-6}$, $fJ^{9}$, $mJ^{4/-5}$, $cA^{5/-3}$, $pA^{7/-1}$, $sA^{5/-3}$, $kA^{7/-1}$} & \multirow{10}{1.7cm}{$cP^{-7}$, $pP^{-7}$, $sP^{-7}$, $kP^{3/-6}$, $fJ^{8}$, $mJ^{8}$, $cA^{4/-2}$, $pA^{4/-2}$, $sA^{4/-2}$, $kA^{4/-2}$} & \multirow{10}{1.7cm}{$cP^{-4}$, $pP^{-3}$, $sP^{-1}$, $kP^{1}$, $fJ^{6}$, $mJ^{-3}$, $cA^{-1}$, $pA^{3}$, $sA^{-2}$, $kA^{4}$} & \multirow{10}{1.7cm}{$cP^{-8}$, $pP^{-8}$, $sP^{4/-2}$, $kP^{5/-2}$, $fJ^{5/-2}$, $mJ^{2/-4}$, $cA^{2/-5}$, $pA^{8}$, $sA^{2/-5}$, $kA^{8}$} & \multirow{10}{1.7cm}{} & \multirow{10}{1.7cm}{$sP^{1}$, $kP^{-2}$, $fJ^{1}$, $mJ^{-6}$, $cA^{1}$, $pA^{2}$, $sA^{1}$, $kA^{2}$} & \multirow{10}{1.7cm}{$cP^{-7}$, $pP^{-7}$, $sP^{-6}$, $kP^{3/-5}$, $fJ^{7}$, $mJ^{2/-2}$, $cA^{4/-1}$, $pA^{4}$, $sA^{4/-1}$, $kA^{5}$} & \multirow{10}{1.7cm}{$cP^{-8}$, $pP^{-8}$, $sP^{2/-6}$, $kP^{2/-4}$, $fJ^{6/-2}$, $mJ^{4}$, $cA^{3/-3}$, $pA^{7}$, $sA^{3/-3}$, $kA^{7}$} \\ \\ \\ \\ \\ \\ \\ \\ \\ \\
\midrule
\multirow{10}{1.7cm}{Fastt Wiki} & \multirow{10}{1.7cm}{$cP^{1/-6}$, $pP^{1/-6}$, $sP^{-9}$, $kP^{1/-6}$, $fJ^{4/-5}$, $mJ^{5/-4}$, $cA^{6/-2}$, $pA^{8/-1}$, $sA^{6/-2}$, $kA^{9}$} & \multirow{10}{1.7cm}{$cP^{-7}$, $pP^{-7}$, $sP^{-7}$, $kP^{3/-6}$, $fJ^{9}$, $mJ^{8/-1}$, $cA^{4/-2}$, $pA^{4/-2}$, $sA^{4/-2}$, $kA^{4/-2}$} & \multirow{10}{1.7cm}{$cP^{1/-6}$, $pP^{1/-6}$, $sP^{-9}$, $kP^{1/-6}$, $fJ^{4/-5}$, $mJ^{5/-4}$, $cA^{6/-2}$, $pA^{8}$, $sA^{6/-2}$, $kA^{8}$} & \multirow{10}{1.7cm}{$cP^{-7}$, $pP^{-7}$, $sP^{-7}$, $kP^{3/-6}$, $fJ^{9}$, $mJ^{8/-1}$, $cA^{4/-2}$, $pA^{4/-2}$, $sA^{4/-2}$, $kA^{4/-2}$} & \multirow{10}{1.7cm}{$kP^{4}$, $fJ^{-1}$, $mJ^{-1}$, $cA^{-1}$, $sA^{-1}$} & \multirow{10}{1.7cm}{$cP^{2}$, $pP^{2}$, $sP^{2}$, $kP^{2}$, $fJ^{2}$, $mJ^{2}$, $cA^{2}$, $pA^{-8}$, $sA^{2}$, $kA^{-8}$} & \multirow{10}{1.7cm}{$cP^{4}$, $pP^{4}$, $sP^{4}$, $kP^{6}$, $fJ^{-4}$, $mJ^{-4}$, $cA^{-4}$, $pA^{-1}$, $sA^{-4}$, $kA^{-1}$} & \multirow{10}{1.7cm}{$cP^{7}$, $pP^{7}$, $sP^{7}$, $kP^{-3}$, $fJ^{-3}$, $mJ^{-3}$, $cA^{-3}$, $pA^{-3}$, $sA^{-3}$, $kA^{-3}$} & \multirow{10}{1.7cm}{$cP^{1/-1}$, $pP^{1/-1}$, $sP^{1/-1}$, $kP^{7}$, $fJ^{-6}$, $mJ^{-3}$, $cA^{-3}$, $pA^{4}$, $sA^{-3}$, $kA^{4}$} & \multirow{10}{1.7cm}{} \\ \\ \\ \\ \\ \\ \\ \\ \\ \\
\midrule
\multirow{10}{1.7cm}{Fastt Crawl} & \multirow{10}{1.7cm}{$cP^{-7}$, $pP^{-7}$, $sP^{-7}$, $kP^{3/-6}$, $fJ^{5/-2}$, $mJ^{4/-5}$, $cA^{5/-2}$, $pA^{8}$, $sA^{5/-2}$, $kA^{8}$} & \multirow{10}{1.7cm}{$cP^{-7}$, $pP^{-7}$, $sP^{-7}$, $kP^{3/-6}$, $fJ^{9}$, $mJ^{8/-1}$, $cA^{4/-2}$, $pA^{4/-2}$, $sA^{4/-2}$, $kA^{4/-2}$} & \multirow{10}{1.7cm}{$cP^{-7}$, $pP^{-7}$, $sP^{-7}$, $kP^{3/-6}$, $fJ^{5/-4}$, $mJ^{4/-5}$, $cA^{6/-2}$, $pA^{8}$, $sA^{6/-2}$, $kA^{8}$} & \multirow{10}{1.7cm}{$cP^{-7}$, $pP^{-7}$, $sP^{-7}$, $kP^{3/-6}$, $fJ^{9}$, $mJ^{8/-1}$, $cA^{4/-2}$, $pA^{4/-2}$, $sA^{4/-2}$, $kA^{4/-2}$} & \multirow{10}{1.7cm}{$cP^{3}$, $pP^{3}$, $sP^{1}$, $kP^{3}$, $fJ^{-6}$, $cA^{-5}$, $pA^{3}$, $sA^{-5}$, $kA^{3}$} & \multirow{10}{1.7cm}{} & \multirow{10}{1.7cm}{$cP^{1}$, $pP^{1}$, $sP^{1}$, $kP^{4}$, $fJ^{-6}$, $mJ^{-1}$, $cA^{-1}$, $pA^{1}$, $sA^{-1}$, $kA^{1}$} & \multirow{10}{1.7cm}{$cP^{6}$, $pP^{6}$, $fJ^{-2}$, $mJ^{-2}$, $cA^{-2}$, $pA^{-2}$, $sA^{-2}$, $kA^{-2}$} & \multirow{10}{1.7cm}{$cP^{3}$, $pP^{3}$, $sP^{1/-1}$, $kP^{4}$, $fJ^{-7}$, $mJ^{1/-1}$, $cA^{-5}$, $pA^{3}$, $sA^{-5}$, $kA^{4}$} & \multirow{10}{1.7cm}{$kP^{3}$, $fJ^{-2}$, $mJ^{3}$, $cA^{-2}$, $sA^{-2}$} \\ \\ \\ \\ \\ \\ \\ \\ \\ \\
\bottomrule
    \end{tabular}
\caption{Similarity metrics for each embedding and dataset (with Levenshtein negative sampling) that have significantly better/worse accuracy than the number of methods given by the positive/negative superscripts. Significance measured by McNemar's test with $\alpha = 0.001$/$0.008$ (word/sentence embeddings), i.e. $\alpha = 0.05$ with Bonferroni correction.}
    \label{tab:significance_snomedDatasets_methods4} 
\end{sidewaystable*}

\begin{sidewaystable*}
    \centering
    \small
    \begin{tabular}{l l l l l l l l l l l }
    \toprule
         Dataset & \dataset{FSN-syn.} & \dataset{FSN-syn.} & \dataset{syn-syn.} & \dataset{syn-syn.}& \dataset{poss.-equ.} & \dataset{poss.-equ.} & \dataset{repl.-by} & \dataset{repl.-by} & \dataset{same-as} & \dataset{same-as} \\
& easy & hard & easy & hard& easy & hard & easy & hard & easy & hard \\ 
         \midrule
\multirow{10}{1.7cm}{Fastt Crawl M} & \multirow{10}{1.7cm}{$cP^{-7}$, $pP^{-7}$, $sP^{-7}$, $kP^{3/-6}$, $fJ^{4/-5}$, $mJ^{5/-4}$, $cA^{6/-2}$, $pA^{8}$, $sA^{6/-2}$, $kA^{8}$} & \multirow{10}{1.7cm}{$cP^{-7}$, $pP^{-7}$, $sP^{-7}$, $kP^{3/-6}$, $fJ^{9}$, $mJ^{8/-1}$, $cA^{4/-2}$, $pA^{4/-2}$, $sA^{4/-2}$, $kA^{4/-2}$} & \multirow{10}{1.7cm}{$cP^{-7}$, $pP^{-7}$, $sP^{-7}$, $kP^{3/-6}$, $fJ^{4/-5}$, $mJ^{5/-4}$, $cA^{6/-2}$, $pA^{8/-1}$, $sA^{6/-2}$, $kA^{9}$} & \multirow{10}{1.7cm}{$cP^{-7}$, $pP^{-7}$, $sP^{-7}$, $kP^{3/-6}$, $fJ^{9}$, $mJ^{8/-1}$, $cA^{4/-2}$, $pA^{4/-2}$, $sA^{4/-2}$, $kA^{4/-2}$} & \multirow{10}{1.7cm}{$cP^{6}$, $pP^{6}$, $sP^{4}$, $kP^{6}$, $fJ^{-4}$, $mJ^{-4}$, $cA^{-4}$, $pA^{-3}$, $sA^{-4}$, $kA^{-3}$} & \multirow{10}{1.7cm}{} & \multirow{10}{1.7cm}{$cP^{4}$, $pP^{2}$, $sP^{2}$, $kP^{6}$, $fJ^{-8}$, $mJ^{-4}$, $cA^{1/-2}$, $pA^{1/-1}$, $sA^{1/-2}$, $kA^{1/-1}$} & \multirow{10}{1.7cm}{$cP^{6}$, $pP^{6}$, $fJ^{-2}$, $mJ^{-2}$, $cA^{-2}$, $pA^{-2}$, $sA^{-2}$, $kA^{-2}$} & \multirow{10}{1.7cm}{$cP^{4}$, $pP^{4}$, $sP^{4/-1}$, $kP^{7}$, $fJ^{-6}$, $mJ^{-5}$, $cA^{-6}$, $pA^{4/-1}$, $sA^{-6}$, $kA^{3/-1}$} & \multirow{10}{1.7cm}{$kP^{6}$, $fJ^{-1}$, $mJ^{-1}$, $cA^{-1}$, $pA^{-1}$, $sA^{-1}$, $kA^{-1}$} \\ \\ \\ \\ \\ \\ \\ \\ \\ \\
\midrule
\multirow{4}{1.7cm}{ElmoPubmed} & \multirow{4}{1.7cm}{$cosine^{-2}$, $pears^{-2}$, $spear^{2}$, $kendall^{2}$} & \multirow{4}{1.7cm}{} & \multirow{4}{1.7cm}{$cosine^{-2}$, $pears^{-2}$, $spear^{2}$, $kendall^{2}$} & \multirow{4}{1.7cm}{} & \multirow{4}{1.7cm}{} & \multirow{4}{1.7cm}{} & \multirow{4}{1.7cm}{} & \multirow{4}{1.7cm}{} & \multirow{4}{1.7cm}{$cosine^{-1}$, $pears^{-1}$, $spear^{2}$} & \multirow{4}{1.7cm}{} \\ \\ \\ \\
\midrule
\multirow{4}{1.7cm}{Flair} & \multirow{4}{1.7cm}{$cosine^{-2}$, $pears^{-2}$, $spear^{2/-1}$, $kendall^{3}$} & \multirow{4}{1.7cm}{} & \multirow{4}{1.7cm}{$cosine^{-2}$, $pears^{-2}$, $spear^{2/-1}$, $kendall^{3}$} & \multirow{4}{1.7cm}{} & \multirow{4}{1.7cm}{} & \multirow{4}{1.7cm}{} & \multirow{4}{1.7cm}{} & \multirow{4}{1.7cm}{} & \multirow{4}{1.7cm}{$cosine^{-2}$, $pears^{-2}$, $spear^{2/-1}$, $kendall^{3}$} & \multirow{4}{1.7cm}{} \\ \\ \\ \\
\midrule
\multirow{4}{1.7cm}{Scibert} & \multirow{4}{1.7cm}{} & \multirow{4}{1.7cm}{} & \multirow{4}{1.7cm}{} & \multirow{4}{1.7cm}{} & \multirow{4}{1.7cm}{} & \multirow{4}{1.7cm}{$cosine^{-1}$, $pears^{-1}$, $kendall^{2}$} & \multirow{4}{1.7cm}{} & \multirow{4}{1.7cm}{} & \multirow{4}{1.7cm}{} & \multirow{4}{1.7cm}{} \\ \\ \\ \\ 
\midrule
\multirow{4}{1.7cm}{Bert} & \multirow{4}{1.7cm}{$cosine^{-2}$, $pears^{-2}$, $spear^{2}$, $kendall^{2}$} & \multirow{4}{1.7cm}{} & \multirow{4}{1.7cm}{$cosine^{-2}$, $pears^{-2}$, $spear^{2}$, $kendall^{2}$} & \multirow{4}{1.7cm}{} & \multirow{4}{1.7cm}{} & \multirow{4}{1.7cm}{} & \multirow{4}{1.7cm}{} & \multirow{4}{1.7cm}{} & \multirow{4}{1.7cm}{} & \multirow{4}{1.7cm}{} \\ \\ \\ \\ 
\midrule
\multirow{4}{1.7cm}{ElmoOrig} & \multirow{4}{1.7cm}{$cosine^{-2}$, $pears^{-2}$, $spear^{3}$, $kendall^{2/-1}$} & \multirow{4}{1.7cm}{} & \multirow{4}{1.7cm}{$cosine^{-2}$, $pears^{-2}$, $spear^{3}$, $kendall^{2/-1}$} & \multirow{4}{1.7cm}{} & \multirow{4}{1.7cm}{} & \multirow{4}{1.7cm}{} & \multirow{4}{1.7cm}{} & \multirow{4}{1.7cm}{} & \multirow{4}{1.7cm}{$cosine^{-1}$, $pears^{-1}$, $spear^{2}$} & \multirow{4}{1.7cm}{} \\ \\ \\ \\ 
\midrule
\multirow{4}{1.7cm}{GPT} & \multirow{4}{1.7cm}{$cosine^{-2}$, $pears^{-2}$, $spear^{2/-1}$, $kendall^{3}$} & \multirow{4}{1.7cm}{} & \multirow{4}{1.7cm}{$cosine^{-2}$, $pears^{-2}$, $spear^{2/-1}$, $kendall^{3}$} & \multirow{4}{1.7cm}{} & \multirow{4}{1.7cm}{} & \multirow{4}{1.7cm}{} & \multirow{4}{1.7cm}{} & \multirow{4}{1.7cm}{} & \multirow{4}{1.7cm}{} & \multirow{4}{1.7cm}{} \\ \\ \\ \\ 
\bottomrule
    \end{tabular}
\caption{Similarity metrics for each embedding and dataset (with Levenshtein negative sampling) that have significantly better/worse accuracy than the number of methods given by the positive/negative superscripts. Significance measured by McNemar's test with $\alpha = 0.001$/$0.008$ (word/sentence embeddings), i.e. $\alpha = 0.05$ with Bonferroni correction.}
    \label{tab:significance_snomedDatasets_methods5} 
\end{sidewaystable*}

\clearpage
\section{Evaluation on Category Separation}
To gain a deeper insight into the separation of categories, we also analysed the \emph{average similarity} of concepts in semantically close categories (TP,DP) versus those in semantically distant categories (TP,Org), as illustrated in Table~\ref{tab:categorisation_evaluation_withSimScores}. We observe that all embeddings result in a higher similarity for the former compared to the latter, as desired.
As a comparison, we also compute the average similarity scores between all concept pairs of a category (last three columns in Table~\ref{tab:categorisation_evaluation_withSimScores}). We expect these in-category similarity scores to be higher than any of the cross-category similarity scores, which is indeed the case.

\begin{table*}[th]
    \centering
    \small
    \begin{tabular}{l l l l l l l l}
           & metric (best/worst) & $O$ (best/worst)  & sim(DP,TP) & sim(DP,Org)  &  sim(TP,TP)  & sim(DP,DP)  & sim(Org,Org) \\ 
         \midrule
LTL win2  & $fJ / pair\_\tau$ & $8.6\% / 20.1\%$  & $0.31/0.16$ & $0.22/0.13$ & $0.42/0.23$ & $0.37/0.19$ & $0.37/0.23$\\
AUEB200 & $fJ / mJ$ & $8.6\%/15.4\%$ & $0.28/0.41$ & $0.19/0.33$ & $0.40/0.45$ & $0.34/0.45$ & $0.35/0.21$\\
MeSH intr & $avg\_cos / pair\_\rho$ & $13.9\% / 17.1\%$ & $0.57/0.36$ & $0.44/0.44$ & $0.65/0.43$ & $0.63/0.38$ & $0.54/0.40$\\
MeSH extr & $avg\_cos / mJ$ & $12.2\% / 19.0\%$ & $0.51/0.41$ & $0.37/0.35$ & $0.59/0.44$ & $0.57/0.45$ & $0.49/0.38$\\
\midrule
ELMo & $r / \tau$ &$14.4\%/18.8\%$ & $0.46/0.23$ & $0.35/0.17$ & $0.53/0.30$ & $0.51/0.27$ & $0.44/0.25$\\
ELMoPubMed & $r / \tau$  &$5.6\% /13.4\% $  & $0.39/0.19$ & $0.26/0.13$ & $0.49/0.27$ & $0.43/0.22$ & $0.43/0.26$ \\
Flair & $r / \tau$ &$17.5\%/21.9\%$ & $0.68/0.48$ & $0.59/0.52$ &
$0.75/0.57$ & $0.70/0.55$ & $0.69/0.56$\\
BERT & $\rho / cos$ &$10.9\%/11.9\%$ & $0.55/0.84$ & $0.42/0.78$ & $0.61/0.86$ & $0.57/0.85$ & $0.42/0.79$ \\
SciBERT & $\rho / r$ &$21.1\% /24.4\%$& $0.41/0.67$ & $0.33/0.62$ & $0.52/0.73$ & $0.45/0.70$ & $0.46/0.68$\\
GPT & $\rho / r$ &$33.2\%/35.4\%$ & $0.76/0.75$ & $0.66/0.67$ & $0.77/0.76$ & $0.77/0.76$ & $0.67/0.69$\\
    \end{tabular}
    \caption{Relative overlap error ($O$) and average similarity scores between concepts in the same and different categories. For each embedding the best/worst performing similarity metric is reported.}
    \label{tab:categorisation_evaluation_withSimScores} 
\end{table*}

\end{document}